\documentclass[main]{article}

\usepackage[normalem]{ulem} 
\newcommand{\rev}[1]{#1}       

\PassOptionsToPackage{numbers,sort&compress}{natbib}
\usepackage[final]{neurips_2026}
\usepackage{amsthm}
\usepackage{subcaption}
\usepackage{wrapfig}
\usepackage[numbers,sort&compress]{natbib}

\usepackage[utf8]{inputenc} 
\usepackage[T1]{fontenc}    
\usepackage{hyperref}       
\usepackage{url}            
\usepackage{booktabs}       
\usepackage{amsfonts}       
\usepackage{nicefrac}       
\usepackage{microtype}      
\usepackage{xcolor}         
\usepackage{multirow}
\usepackage{graphicx}
\usepackage{amsmath}
\usepackage{amsthm}
\usepackage{algorithm}
\usepackage{algpseudocode}
\usepackage{amssymb}
\usepackage{stmaryrd}

\title{Energy-Guided Transport for Projection-Free Physics-Informed Flow Matching}

\author{%
  Giuliano Galadini \qquad Aurelio Uncini \qquad Danilo Comminiello \\
  Department of Information Engineering, Electronics and Telecommunications \\
  Sapienza University of Rome \\
  Rome, Italy \\
  \texttt{\{giuliano.galadini, aurelio.uncini, danilo.comminiello\}@uniroma1.it} \\
} 

\begin{document}

\maketitle

\begin{abstract}
Deep generative models, particularly Flow Matching frameworks, have emerged as powerful tools for synthesizing complex physical systems. However, standard architectures remain purely statistically driven and agnostic to the underlying governing equations, often violating fundamental physical constraints during inference. Existing zero-shot approaches designed to enforce these constraints typically rely on rigid algebraic projections. While effective for sparse boundary conditions, applying such discrete geometric corrections to dense, full-domain PDEs may disrupt the probability path, \rev{potentially deviating from the learned manifold and introducing substantial computational overhead}. To address these challenges, we introduce Energy-Guided Transport, a novel training-free framework that formulates non-linear physical constraints as a smooth potential energy landscape. Rather than interrupting the ODE integration, our method applies a continuous dissipative force directly to the intermediate generative states, steering them toward the physical manifold. By evaluating this force implicitly via reverse-mode automatic differentiation, our framework preserves the natural dynamics of the flow while maintaining an efficient $\mathcal{O}(N)$ algorithmic complexity. We extensively evaluate our framework across various 1D and 2D physical systems, including the chaotic Navier-Stokes equations. Our method achieves stable generation without structural divergence, significantly improves the stability-accuracy trade-off compared to state-of-the-art methods, and substantially accelerates inference with respect to exact projection strategies. Code available at: \url{https://github.com/ispamm/Projection-free-energy-guided-flow-matching}.
\end{abstract}

\section{Introduction}
\label{sec:intro}

\begin{figure*}[h!]
     \centering
     \begin{subfigure}[b]{0.45\textwidth}
         \centering
         \includegraphics[width=\textwidth]{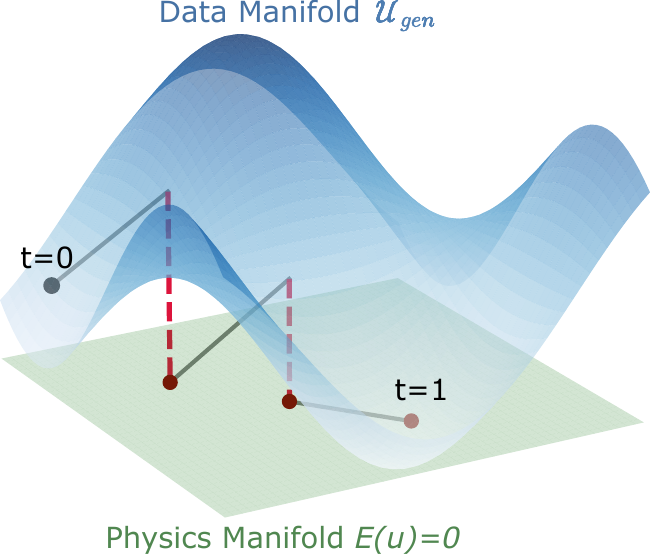}
         \caption{Explicit Hard Projections}
         \label{fig:pcfm_concept}
     \end{subfigure}
     \hfill
     \begin{subfigure}[b]{0.45\textwidth}
         \centering
         \includegraphics[width=\textwidth]{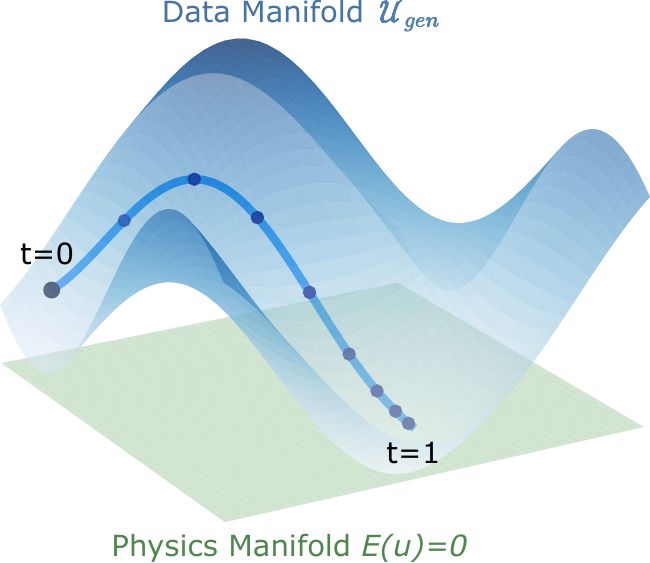}
         \caption{EGT-FFM (Ours)}
         \label{fig:ours_concept}
     \end{subfigure}
     \caption{Conceptual comparison of generative probability flows. (a) \rev{Algebraic projections interrupt continuous integration, potentially displacing trajectories from the learned data manifold.} (b) Our dissipative guidance smoothly steers the trajectory along the data manifold towards the physical equilibrium.}
     \label{fig:teaser}
\end{figure*}

The accurate simulation of complex spatiotemporal dynamics has traditionally relied on computationally expensive numerical solvers. To bypass the curse of dimensionality inherent in high-dimensional PDEs, deep neural networks were initially introduced as direct physics solvers~\cite{han2018solving, sirignano2018dgm}. Building on this foundation, deep generative models, ranging from Score-Based Diffusion \cite{ho2020denoising, DBLP:conf/iclr/0011SKKEP21} to Flow Matching frameworks \cite{DBLP:conf/iclr/LipmanCBNL23, DBLP:journals/jmlr/AlbergoBV25}, have recently emerged as scalable data-driven surrogates for modeling physical systems \cite{karniadakis2021physics, kochkov2021machine, brandstettermessage, brunton2020machine}. These models have achieved breakthrough successes in global weather forecasting and fluid dynamics \cite{DBLP:journals/nature/BiXZCG023, lam2023learning, pathak2022fourcastnet, kashinath2021physics, DBLP:conf/kdd/WangKMAY20, cai2021physics}.

By learning the intrinsic data manifold from high-fidelity solvers, these generative architectures offer unprecedented scalability, producing complex, high-dimensional physical states in a fraction of the time required by traditional simulators~\cite{kochkov2021machine, DBLP:journals/nature/BiXZCG023, lam2023learning}. However, standard generative models remain purely statistical; they synthesize visually convincing fields but lack an inherent understanding of the governing physics. This purely statistical nature often leads to severe violations of conservation laws, boundary conditions, and internal dynamic consistencies \cite{DBLP:conf/icml/HansenMAGM23}. Ultimately, violations of these macroscopic constraints severely degrade out-of-distribution generalization \cite{mouli2024using} and the reliability of foundation models for PDEs \cite{herde2024poseidon}.

To bridge this gap, historical techniques attempted injecting physical priors into the generative process. Early approaches primarily focused on data-driven inverse problems and inpainting \cite{batzolis2021conditional, song2021maximum, deng2021continuous}. In these settings, models rely heavily on dense interior observations (e.g., sparse sensor data) to anchor the generative trajectory, reducing the physics problem to a conditioned interpolation task. While highly effective for data assimilation, these observation-dependent methods fail to act as rigorous, independent PDE solvers. Consequently, the field recently shifted toward Zero-Shot Physics Guidance, a much stricter paradigm. In this regime, a pre-trained base model is guided exclusively by the analytical PDE and its formal Boundary/Initial Conditions (BC/IC) during inference, without access to interior ground-truth data or computationally prohibitive retraining.

Despite the theoretical elegance of zero-shot inference, modern guidance architectures suffer from critical structural and computational limitations. \rev{Existing approaches often face trade-offs between numerical stability and computational scalability: gradient-based approximations can become unstable in non-linear regimes, while explicit algebraic projections (e.g., PCFM~\cite{utkarshphysics}), though effective for sparse boundary conditions, incur significant computational overhead and may perturb the generative trajectory when applied to dense, full-domain PDE constraints.}

To address these limitations, we propose Energy-Guided Transport (EGT), \rev{a continuous guidance framework for physics-informed Functional Flow Matching.} Rather than relying on unstable approximations or discrete geometric projections that intermittently halt the generative process, we formulate the physical PDE residual as a smooth potential energy landscape. Central to our approach are the Dissipation properties of Flow Matching, which proves that modulating the generative vector field with the gradient of this physical energy applies a Lyapunov-like dissipative force. As conceptualized in Figure \ref{fig:teaser}, this allows our guidance to natively harmonize with the probability flow: it smoothly counteracts physical violations along the trajectory, steering samples toward the true physical manifold without the structural discontinuities or $\mathcal{O}(N^2)$ memory bottlenecks of exact algebraic projections. 

To empirically validate our theoretical claims, we conduct an exhaustive experimental campaign across four distinct physical systems of increasing non-linearity and dimensionality: the 1D Heat equation, 1D Reaction-Diffusion (RD), 1D Burgers' equation, and the chaotic 2D Navier-Stokes equations. Most importantly, for each physical system, we rigorously evaluate our framework under two distinct regimes: partial physics guidance (enforcing only boundary and initial conditions) and a highly constrained full-PDE stress test. Furthermore, we provide extensive ablation studies investigating the hyperparameter dynamics of the dissipative force and the structural impact of generative priors, comparing standard White Noise against spatially correlated Gaussian Random Fields (Smooth Kernels).

Our primary contributions are summarized as follows:
\begin{itemize}
    \item \textbf{Theoretical Formulation of Dissipative Guidance:} we establish the dissipation properties of Flow Matching, demonstrating via a Lagrangian perspective how a continuous dissipative force ensures the minimization of physical energy functionals. By steering the vector field natively rather than interrupting the ODE to compute discrete projections, our framework seamlessly preserves the foundational continuity of Flow Matching.
    \item \textbf{Analysis of Projection-Based Guidance:} we provide a rigorous empirical analysis exposing the structural limitations of explicit projection methods. We demonstrate that enforcing rigid algebraic constraints can disrupt the generative flow, highlighting the trade-off between exact boundary matching and internal physical coherence (\rev{the waterbed effect}).
    \item \textbf{Computational Scalability:} by completely avoiding explicit Jacobian materialization and employing VJP-based dissipative guidance coupled with gradient norm clipping, our method circumvents the $\mathcal{O}(N^2)$ memory bottlenecks, enabling scalable full-PDE guidance on 2D fluid dynamics.
    \item \rev{\textbf{Favorable Speed--Stability--Accuracy Trade-Off:}} through a strict zero-shot evaluation protocol across four benchmark datasets, our method is capable of maintaining consistent numerical stability ($100\%$ success rate) while delivering \rev{competitive distributional fidelity and low physical residuals at up to two orders of magnitude faster inference speeds than explicit projection methods.}
\end{itemize}

\section{Related Work}
\label{sec:related_work}

\textbf{Generative Models for Physical Systems.} While early advancements in modeling complex physical dynamics relied heavily on Score-Based Diffusion Models~\cite{DBLP:conf/iclr/0011SKKEP21}, recent attention has shifted toward Flow Matching (FM) frameworks~\cite{DBLP:conf/iclr/LipmanCBNL23, DBLP:journals/jmlr/AlbergoBV25, liuflow}, drawing deep connections with computational optimal transport~\cite{peyre2019computational, DBLP:journals/tmlr/0001FMHZRWB24,DBLP:conf/icml/PooladianBDALC23}. By formulating generation as an Ordinary Differential Equation (ODE)~\cite{chen2018neural} and learning a continuous vector field, FM provides faster convergence and structurally smoother trajectories, seamlessly adapting even to complex geometries and Riemannian manifolds~\cite{DBLP:conf/iclr/ChenL24, mathieu2020riemannian, gemici2016normalizing, DBLP:conf/nips/BortoliMHTTD22}. In the context of spatiotemporal physics, Functional Flow Matching (FFM) combines continuous normalizing flows with Neural Operators (e.g., Fourier Neural Operators~\cite{DBLP:conf/iclr/LiKALBSA21}, DeepONet~\cite{lu2021learning}, and general operator frameworks~\cite{kovachki2023neural, li2024physics}). Recently, the extension of these score-based and flow paradigms to functional spaces has gained significant traction~\cite{lim2025score, DBLP:conf/aistats/KerriganMS24}. To overcome this purely statistical nature, the literature has historically turned to physics-informed training.

\textbf{Physics-Informed Training.} The historical paradigm for injecting physics into neural networks incorporates analytical residuals directly into the training objective (e.g., PINNs~\cite{DBLP:journals/jcphy/RaissiPK19, raissi2020hidden, DBLP:journals/jcphy/JinCLK21}). While highly successful for specific forward and inverse regression tasks, applying this paradigm to generative modeling fundamentally limits generalizability. Enforcing physics during training structurally couples the generative prior to specific boundary and initial conditions. Consequently, adapting the model to novel constraints requires computationally prohibitive retraining, rendering these approaches inflexible. Furthermore, directly optimizing physical residuals alongside data objectives during network training often suffers from severe gradient flow pathologies, where competing loss terms prevent convergence~\cite{krishnapriyan2021characterizing, wang2021understanding}. This limitation has motivated the recent paradigm shift toward Zero-Shot guidance.

\textbf{Zero-Shot Physics Guidance.} Zero-shot inference methods attempt to correct generative trajectories using analytical PDE constraints without altering the pre-trained weights. Drawing inspiration from classifier-free~\cite{ho2021classifier} and universal guidance~\cite{DBLP:conf/iclr/BansalCSSGGG24} in vision tasks, these approaches generally fall into three categories:

\begin{itemize}
    \item \textit{Optimization-Based Guidance:} Methods such as DiffusionPDE~\cite{DBLP:conf/nips/HuangYWP24} and D-Flow~\cite{DBLP:conf/icml/Ben-HamuPGKSL24} utilize gradients of physical residuals to shift the generative trajectory. For instance, D-Flow optimizes the initial noise to align the final state with the constraints. However, because diffusion backbones are heavily corrupted by noise, computing physical gradients directly on intermediate states is unstable. Consequently, these methods often rely on a ``predict-and-correct" mechanism using denoised terminal predictions (e.g., via Tweedie's formula), which can lead to overshooting in highly non-linear dynamics.

    \item \textit{Projection-Based Guidance:} To avoid gradient-based instability, frameworks like Extrapolation, Correction, Interpolation (ECI)~\cite{chenggradient} and Physics-Constrained Flow Matching (PCFM)~\cite{utkarshphysics} apply explicit geometric corrections. ECI is a gradient-free approach that interleaves extrapolation, orthogonal projection, and interpolation to strictly enforce linear constraints. PCFM extends this concept to arbitrary nonlinear constraints using Gauss-Newton updates. Notably, while PCFM is highly memory-efficient ($\mathcal{O}(n)$ complexity) for sparse constraints by solving a small $m \times m$ Schur complement system, applying its Jacobian-based projections to dense, full-domain PDE residuals incurs prohibitive computational costs. Furthermore, rigid geometric projections can sometimes force the intermediate states off the learned data manifold.

    \item \textit{Proximal-Based Guidance:} To reconcile strict physical consistency with the pre-trained prior, recent frameworks like ProFlow~\cite{yu2026ProFlow} reframe the correction step as a proximal optimization problem. ProFlow alternates between a terminal optimization step, which anchors the prediction to the physically consistent set via proximal minimization, and an interpolation step that restores compatibility with the learned flow probability path. This implicitly balances trust in the generative prior against physical and data fidelity, avoiding both the intractability of full Jacobian inversions and the trajectory disruption caused by pure algebraic projections.
\end{itemize}

\rev{Unlike these paradigms, our Energy-Guided FFM avoids both unrolled terminal-state estimations and discrete intermediate projections.} By applying a continuous dissipative force directly to the intermediate states, we natively harmonize physical consistency with the probability flow while maintaining strict algorithmic scalability.

\section{Methodology}

\subsection{Problem Formulation}
\label{sec:problem_setup}

We consider physical systems governed by time-dependent Partial Differential Equations (PDEs)~\cite{evans2022partial} defined over a bounded spatial domain $\Omega \subset \mathbb{R}^d$ and a time interval $[0, T]$. Let $u: \Omega \times [0, T] \to \mathbb{R}^n$ denote the spatio-temporal solution field. A generalized physical system is strictly characterized by its internal dynamics, initial conditions (IC), and boundary conditions (BC):
\begin{equation}
    \partial_t u + \mathcal{N}(u; \theta) = 0 \text{ in } \Omega \times (0, T], \quad
    u(\cdot, 0) = u_0 \text{ in } \Omega, \quad
    \mathcal{B}(u) = 0 \text{ on } \partial\Omega \times [0, T].
    \label{eq:pde_system}
\end{equation}
where $\mathcal{N}$ is a spatial differential operator, $\theta$ denotes the physical parameters, and $\mathcal{B}$ is the boundary operator. We denote the manifold of all physically admissible functions that exactly satisfy Eq. (\ref{eq:pde_system}) as $\mathcal{U}_{true}$. 

In generative scientific machine learning, we assume access to a pre-trained base model (e.g., an unguided Flow Matching network) that approximates the empirical data distribution. However, due to finite data and neural approximations, this unguided model generates solutions from a broader, unconstrained support $\mathcal{U}_{gen}$, frequently violating the strict structural properties of $\mathcal{U}_{true}$. 

Our objective is to formulate a zero-shot inference mechanism that smoothly steers the generative trajectories \rev{toward the physically admissible manifold $\mathcal{U}_{true}$.} To achieve this, we encapsulate the targeted physical laws into a general constraint operator $h(u) = 0$. Depending on the downstream task, $h(u)$ can represent partial constraints (e.g., IC/BC enforcement) or the highly non-linear, full microscopic PDE residual. A comprehensive theoretical taxonomy of these constraint operators and their energy landscapes is provided in Appendix~\ref{app:taxonomy}.

\subsection{Functional Flow Matching Foundations}
\label{sec:ffm_theory}

Our framework builds upon Functional Flow Matching (FFM)~\cite{DBLP:conf/aistats/KerriganMS24}, which generalizes diffusion paradigms to infinite-dimensional Hilbert spaces $\mathcal{H}$ (e.g., $L^2(\Omega)$). Let $p_0(u)$ be a tractable prior distribution over functions (e.g., Gaussian random fields) and $p_1(u)$ be the empirical target distribution of valid physical states. FFM learns a neural functional vector field $v_\theta(u, t): \mathcal{H} \times [0, 1] \to \mathcal{H}$ to parameterize the probability-flow ODE:
\begin{equation}
    \frac{du}{dt} = v_\theta(u, t), \quad u(0) \sim p_0,
\end{equation}
which continuously transports probability mass such that the marginal density $p_t(u)$ interpolates between $p_0$ and $p_1$. Utilizing an Optimal Transport (OT) formulation, the conditional vector field is defined along deterministic paths as $v_t(u|u_1) = u_1 - u_0$. 

Operating directly in function space yields two structural advantages for PDE-constrained generation. First, it ensures resolution invariance: the learned field $v_\theta$ can be evaluated at arbitrary spatial resolutions during inference, mitigating numerical discretization errors when computing the physical constraint $h(u)$. Second, intermediate states $u_t$ evolve as continuous, structurally coherent fields rather than unstructured discrete noise arrays, enabling the stable evaluation of spatial differential operators at any integration timestep.

\subsection{Energy-Guided Transport}
\label{sec:continuous_guidance}

\begin{algorithm}[t]
\caption{Energy-Guided Transport for Functional Flow Matching}
\label{alg:guided_sampling}
\begin{algorithmic}[1]
\Require Pre-trained vector field $v_\theta$, Constraint $h(u)$, Energy $E(u) = \frac{1}{2}\|h(u)\|^2_2$
\Require Integration steps $N$, Max guidance $\gamma_{max}$, Clip threshold $c$
\State Sample initial state $u_0 \sim p_0(u)$ \Comment{e.g., Smooth Gaussian Kernel}
\State Set step size $\Delta t = 1 / N$
\State $u \leftarrow u_0$

\For{$i = 0$ to $N - 1$} \Comment{Continuous Guided Flow}
    \State $t \leftarrow i \cdot \Delta t$
    \State $\gamma_t \leftarrow \text{ComputeSchedule}(\gamma_{max}, t)$ \Comment{e.g., Parabolic or Sine}
    
    \State $u \leftarrow \text{RequireGradient}(u)$
    \State $v_{\text{base}} \leftarrow v_\theta(u, t)$ \Comment{Evaluate neural prior}
    \State $g \leftarrow \nabla_{u} E(u)$ \Comment{Compute implicit VJP via reverse-mode AD}
    
    \State $\tilde{g} \leftarrow g \cdot \min\left(1, \frac{c}{\|g\|_2}\right)$ \Comment{Apply gradient clipping}
    \State $u \leftarrow \text{StopGradient}(u) + \Delta t \left( v_{\text{base}} - \gamma_t \tilde{g} \right)$ \Comment{Euler ODE update}
\EndFor

\State \Return Generated physical field $u$
\end{algorithmic}
\end{algorithm}

Formulating physical dynamics through energy or Hamiltonian landscapes has shown great success in forward modeling \cite{greydanus2019hamiltonian, richter2022neural}. We extend this principle to generative guidance, leveraging implicit differentiable physics~\cite{holllearning}. To embed physical knowledge directly into the probability flow without breaking its continuous-time trajectory, we define a physical energy functional $E(u) = \frac{1}{2} \| h(u) \|^2_2$, where $h(u) \in \mathbb{R}^m$ is the discretized non-linear PDE residual. We augment the baseline neural vector field $v_\theta(u, t)$ by injecting the functional gradient of this energy, effectively interpreting it as a time-annealed, Gibbs-like log-prior:
\begin{equation}
    \bar{v}_t(u) = v_\theta(u, t) - \gamma(t) \nabla_u E(u), \quad u(0) \sim p_0
\label{eq:augmented_field}
\end{equation}
where $\gamma(t) \geq 0$ is a scheduling function acting as the inverse temperature $1/\tau(t)$. This thermodynamic cooling schedule allows the generative prior to dictate the macroscopic structure at high temperatures ($t \approx 0$) before strictly locking onto the physical manifold ($E(u)=0$) at low temperatures ($t \to 1$).

\begin{figure*}[t]
    \centering
    \includegraphics[width=\textwidth]{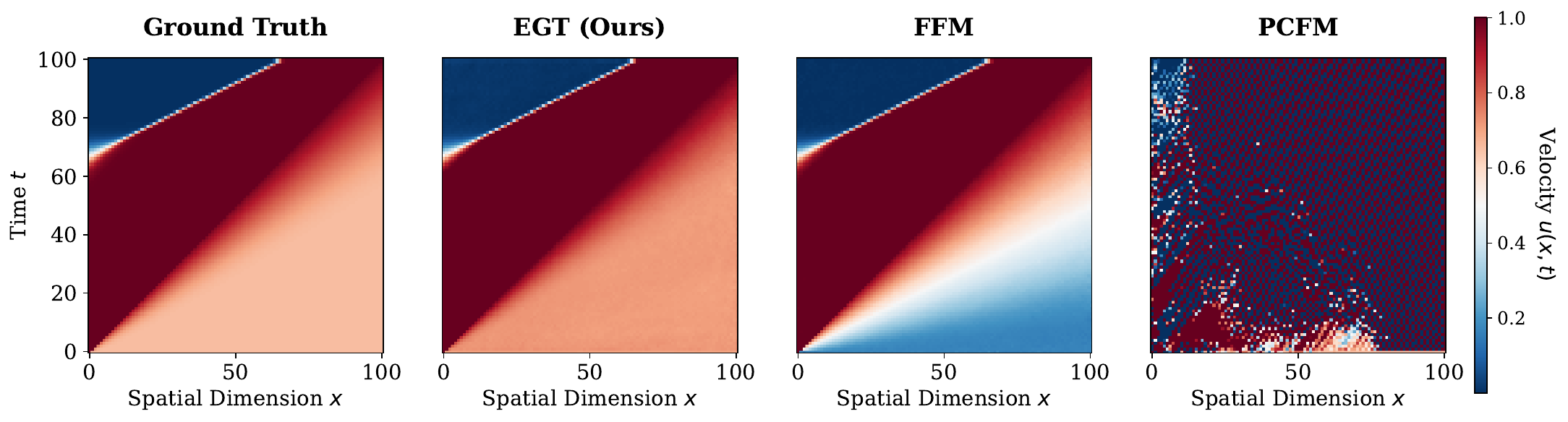}
    \caption{Qualitative evaluation on the 1D Burgers' equation (Full-PDE Stress Test). Spatiotemporal velocity fields $u(x,t)$ generated from the same initial conditions. \rev{Explicit algebraic projections (PCFM) can be sensitive to solver step sizes under non-linear shock dynamics, leading to numerical instabilities and high-frequency artifacts. In comparison, continuous Energy-Guided Transport provides smooth updates along the trajectory, accurately resolving sharp shockfronts while maintaining distributional fidelity.}}
    \label{fig:qualitative_burgers}
\end{figure*}

\textbf{Scalable Dissipative Force.} Computing the guidance field $\nabla_u E(u)$ requires differentiating through complex, non-linear spatial operators. By the chain rule, $\nabla_u E(u) = J_h(u)^T h(u)$, where $J_h \in \mathbb{R}^{m \times N}$ is the constraint Jacobian. Algebraic projection methods (e.g., PCFM~\cite{utkarshphysics}) enforce constraints by explicitly materializing and solving systems involving $J_h J_h^T$. While highly memory-efficient for sparse boundaries (where $m \ll N$), this geometric approach incurs an intractable $\mathcal{O}(N^2)$ memory bottleneck when scaled to dense, full-domain PDE residuals ($m \approx N$). Our energy-based formulation naturally circumvents this limitation: rather than materializing $J_h$, the dissipative force is evaluated implicitly via standard reverse-mode automatic differentiation (Vector-Jacobian Products). This guarantees strict $\mathcal{O}(N)$ space and time complexity, effortlessly scaling full-PDE guidance to high-dimensional fields without structural bottlenecks (see Appendix \ref{app:complexity} for a complexity analysis). \rev{Furthermore, by relying exclusively on matrix-vector multiplications via VJPs rather than inverting local linear systems, EGT naturally handles ill-conditioned or rank-deficient Jacobians without numerical instability (producing zero gradients along rank-deficient directions). Unlike approximate Gauss-Newton or preconditioned iterative solvers that still require multiple inner-loop evaluations per step to approximate a projection, EGT embeds the physical correction directly into the transport dynamics via a single VJP evaluation.}

\textbf{Theoretical Guarantees.} 
To show that Eq. (\ref{eq:augmented_field}) \rev{introduces a principled dissipative correction} without arbitrarily perturbing the underlying data manifold, we establish the following propositions governing the continuous transport of probability mass (full analytical proofs are provided in Appendix \ref{app:dissipation_proofs}).

\newtheorem*{proposition*}{Proposition}
\begin{proposition*} \textbf{(Continuous Energy Dissipation under Ideal Transport)}
\label{prop:dissipation}
Let $\mathbb{E}_{p_t}[E(u)]$ denote the expected physical energy over the marginal distribution at time $t$. The instantaneous variation of the physical error under the augmented field decomposes into the baseline variation minus a strictly non-negative dissipative term:
\begin{equation}
 \frac{d\mathbb{E}_{\text{total}}}{dt} = \mathbb{E}_{p_t}\left[ \langle \nabla_u E, v_\theta \rangle \right] - \gamma(t) \mathbb{E}_{p_t}\left[ \|\nabla_u E\|^2_2 \right]
\label{eq:dissipation}
\end{equation}
Consequently, for any $\gamma(t) \geq 0$, the continuous injection strictly dampens the physical violation relative to the baseline ODE trajectory, effectively acting as a Lyapunov-like dissipative mechanism for the physical constraint.
\end{proposition*}

While Proposition~\ref{prop:dissipation} holds identically for the generative flow, in practice, the neural field $v_\theta$ exhibits an approximation error $\delta(u,t)$ relative to the optimal transport field. By bounding this error such that $\|\delta(u, t)\| \leq \epsilon_t$, we formulate the global descent guarantees for the practical implementation:

\newtheorem*{corollary*}{Corollary}
\begin{corollary*} \textbf{(Error-Bounded Global Energy Descent)}
\label{cor:error_bounded}
The true rate of change of the expected physical energy is bounded by the ideal transport variation and an error term proportional to the network's inaccuracy. Integrating this instantaneous rate over the generation interval $t \in [0, 1]$ yields the macroscopic physical variation:
\begin{equation}
\mathbb{E}[E(u_1)] - \mathbb{E}[E(u_0)] \leq \Delta E_{\text{base}} - \int_0^1 \gamma(t) \mathbb{E}_{p_t}\left[ \|\nabla_u E\|^2_2 \right] dt + \mathcal{O}(\bar{\epsilon})
\end{equation}
\end{corollary*}
The instantaneous and cumulative forms of this bound are stated and proved as Corollaries 1 and 2 in Appendix~\ref{app:dissipation_proofs}.

\rev{\textbf{Scope of the Theoretical Guarantee.} Importantly, Proposition~\ref{prop:dissipation} and Corollary~\ref{cor:error_bounded} establish a continuous dissipative correction relative to the unguided flow under finite-step inference, rather than asymptotic convergence to $E(u_1) = 0$. Instantaneous energy decrease is guaranteed whenever the dissipative term $\gamma(t)\|\nabla_u E\|_2^2$ dominates the baseline alignment $\langle \nabla_u E, v_\theta \rangle$; if the learned flow strongly opposes the physical gradient, the residual may temporarily increase. When machine-precision constraint satisfaction is strictly required, a lightweight terminal refinement can be applied at $t=1$, motivating our Hybrid EGT variant (see Section~\ref{sec:implementation}).}

\subsection{Architectural Synergies: FFM and Smooth Priors}
\label{sec:ffm_choice}


While Proposition ~\ref{prop:dissipation} theoretically justifies continuous guidance across any probability flow, its practical success is deeply synergistic with the structural properties of Functional Flow Matching. Unlike diffusion models, where intermediate states are heavily corrupted by stochastic noise, FFM constructs a deterministic ODE path. Consequently, intermediate states $u_t$ rapidly emerge as structurally coherent fields. This coherence allows for the stable, direct evaluation of spatial differential operators (e.g., $\nabla u_t, \Delta u_t$) and the physical energy $E(u_t)$ along the generative trajectory, completely bypassing the need for terminal-state approximations.

Furthermore, this synergy can be naturally extended through the choice of the generative prior. While our dissipative guidance is highly effective under standard White Noise, utilizing a spatially correlated Gaussian Random Field (Smooth Kernel) ensures inherent spatial differentiability right from $t=0$. By providing a well-behaved energy landscape from the very first integration step, this formulation allows for the stable evaluation of physical derivatives across the entire generative trajectory, fully unlocking the potential of continuous guidance without early-stage numerical truncation.

\subsection{Practical Implementation and Sampling Algorithm}
\label{sec:implementation}

The complete generative inference process is summarized in Algorithm \ref{alg:guided_sampling}. To ensure stability across highly non-linear PDE landscapes without sacrificing the $\mathcal{O}(N)$ computational complexity, our implementation incorporates three core mechanisms: $L^2$-norm gradient clipping, dynamic guidance scheduling, and first-order Euler integration.

We formally define two variants of our framework. Our primary, natively continuous approach is denoted as EGT. Additionally, we evaluate Hybrid EGT, which incorporates an optional post-hoc refinement phase (pure energy descent at $t=1$) to strictly anchor the final sample to the physical manifold. \rev{Rather than replacing terminal refinement, pure EGT replaces expensive repeated projections along the flow, steering the trajectory into a favorable physical ``basin of attraction'' that serves as an optimal initialization for lightweight post-hoc refinement.} Unless otherwise specified, our primary visualizations highlight pure EGT. This isolates our core hypothesis: continuous guidance during the flow is fundamentally superior to discrete algebraic corrections. Comprehensive implementation details, hyperparameter configurations, and the exact formulation of the guidance schedules are provided in Appendix \ref{app:implementation_details}. 

\begin{figure*}[t]
    \centering
    \includegraphics[width=\textwidth]{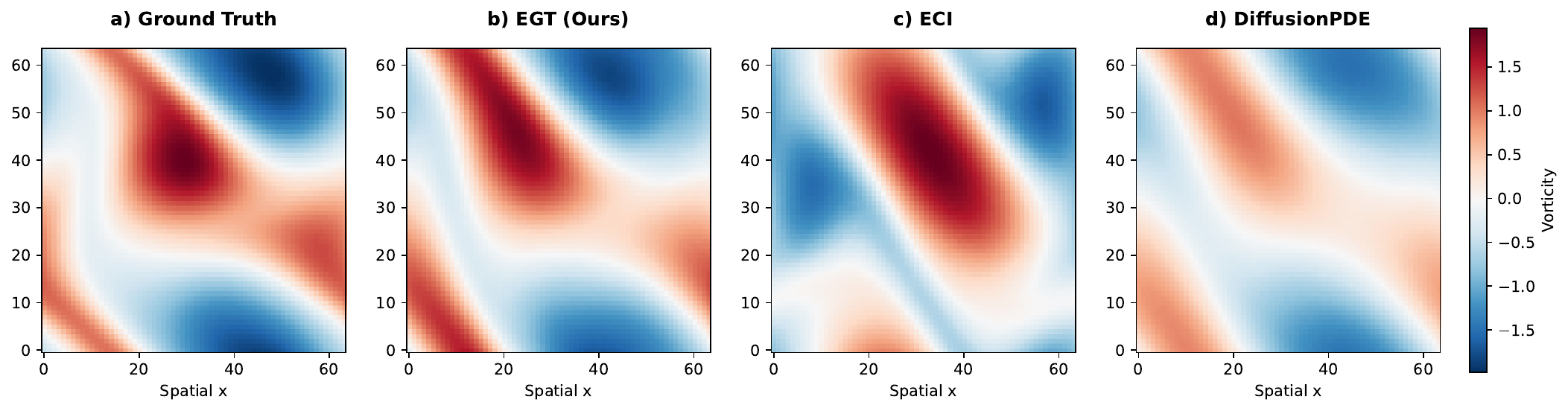}
    \caption{Qualitative evaluation on 2D Navier-Stokes vorticity fields. While all guided frameworks attempt to satisfy the underlying physical constraints, (c) ECI and (d) DiffusionPDE tend to introduce noticeable structural blurring and dampen the vorticity intensity. In contrast, our proposed (b) EGT better preserves the macroscopic structural coherence and the sharp gradients of the flow, offering a closer approximation to the (a) Ground Truth dynamics.}
    \label{fig:qualitative_ns}
\end{figure*}

\section{Experimental Results}
\label{sec:experiments}

To rigorously evaluate our Energy-Guided Transport, we benchmark against state-of-the-art physics-guided generative models: ProFlow~\cite{yu2026ProFlow}, ECI~\cite{chenggradient}, DiffusionPDE~\cite{DBLP:conf/nips/HuangYWP24}, D-Flow~\cite{DBLP:conf/icml/Ben-HamuPGKSL24}, and PCFM~\cite{utkarshphysics}. We test across four physical systems: 1D Heat, 1D Reaction-Diffusion (RD), 1D Burgers', and 2D Navier-Stokes. To isolate their capabilities as pure PDE solvers and guarantee strict fairness, we adopt the zero-shot evaluation framework established by PCFM~\cite{utkarshphysics}. Specifically, all competing methods are denied access to interior ground-truth data and share the identical pre-trained FFM backbone and initial noise realizations.
Performance is evaluated via Success Rate (SR) for numerical stability, Inference Speed, Sample MSE for distributional fidelity, and Physical Residual (Phys) MAE. Full formulations and extended metrics are deferred to Appendices \ref{app:datasets} and \ref{app:extended_results}.
Table \ref{tab:main_results} summarizes quantitative performance under a standard White Noise prior, while Table \ref{tab:smooth_kernel_results} details the corresponding performance under a spatially correlated Smooth Kernel prior. We evaluate all methods under two constraint regimes: Partial Guidance (Boundary/Initial Conditions only) and the Full-PDE Stress Test. Comprehensive visualizations are provided in our Github repository.

\textbf{Partial Physics Guidance.} Under relaxed boundary constraints, geometric projection methods (e.g., ECI, PCFM) achieve near-zero physical errors on specific equations (RD, Burgers). However, their stability and efficiency are inconsistent across domains: PCFM drops to a $0.2\%$ SR on the Heat equation and \rev{exhibits substantial latency} on 2D Navier-Stokes ($\sim 94$s per sample), while concurrently degrading the Sample MSE to $0.76$ (compared to $0.54$ for the Vanilla model). Similarly, terminal-state optimization via D-Flow struggles with high dimensionality, achieving only a $14.2\%$ SR on Navier-Stokes. While other methods like DiffusionPDE and ProFlow maintain stability, they occasionally degrade the baseline distributional fidelity (e.g., DiffusionPDE increasing the Sample MSE to $0.74$ on Navier-Stokes). In contrast, our continuous guidance maintains a $100\%$ Success Rate across all environments, completing the 2D Navier-Stokes generation in just $0.71$s while delivering competitive physical and visual fidelity.

\begin{table*}[t]
\centering
\caption{Quantitative comparison across PDE datasets (White Noise Prior). Panel A shows performance with weak constraints (BC/IC only). Panel B shows the stress test constrained by the Full PDE. For each dataset, we report Success Rate (\textbf{SR}, \%), average inference speed (\textbf{Spd}, sec/sample), Sample MSE (\textbf{MSE}), and physical residual (\textbf{Phys}). DNC (Did Not Complete) denotes trajectories that encountered numerical divergence or memory exhaustion (NaN/OOM). Best results are highlighted in \textbf{bold}, and second-best are \underline{underlined} (excluding the unguided Vanilla baseline). Values marked with $\dagger$ \rev{fall below the discrete ground-truth stencil precision limit (stencil overfitting/waterbed effect)}. See Appendix \ref{app:waterbed} for an analysis of the Waterbed Effect.}
\label{tab:main_results}
\resizebox{\textwidth}{!}{%
\begin{tabular}{l | c c c c | c c c c | c c c c | c c c c}
\toprule
\multirow{2}{*}{\textbf{Method}} & \multicolumn{4}{c|}{\textbf{Heat}} & \multicolumn{4}{c|}{\textbf{RD}} & \multicolumn{4}{c|}{\textbf{Burgers}} & \multicolumn{4}{c}{\textbf{Navier-Stokes}} \\
& SR$\uparrow$ & Spd$\downarrow$ & MSE$\downarrow$ & Phys$\downarrow$ & SR$\uparrow$ & Spd$\downarrow$ & MSE$\downarrow$ & Phys$\downarrow$ & SR$\uparrow$ & Spd$\downarrow$ & MSE$\downarrow$ & Phys$\downarrow$ & SR$\uparrow$ & Spd$\downarrow$ & MSE$\downarrow$ & Phys$\downarrow$ \\
\midrule
\multicolumn{17}{l}{\textit{Panel A: Partial Physics Guidance (Boundary \& Initial Conditions)}} \\
\midrule
Vanilla FFM  & 100 & 0.19 & 0.08 & 0.26 & 100 & 0.38 & 0.08 & 0.20 & 100 & 0.42 & 0.21 & 0.10 & 100 & 0.69 & 0.54 & 0.29 \\
\midrule
ProFlow      & \textbf{100} & 8.13 & 0.06 & 0.19 & \textbf{100} & 25.65 & 7.38e+05 & 1.16e+03 & \textbf{100} & 21.34 & 0.23 & 0.13 & \textbf{100} & \underline{0.85} & 0.55 & 0.20 \\
ECI          & \textbf{100} & 8.31 & \textbf{0.01} & \textbf{0.01} & \textbf{100} & 24.47 & 0.25 & 0.05 & \textbf{100} & 21.87 & \textbf{0.11} & 0.05 & \textbf{100} & 3.56 & 0.63 & \underline{3.75e-05} \\
DiffusionPDE & \textbf{100} & 0.95 & \textbf{0.01} & 0.03 & \textbf{100} & 2.00 & \textbf{0.02} & 0.04 & \textbf{100} & 1.17 & \underline{0.15} & 0.06 & \textbf{100} & 1.40 & \underline{0.54} & 0.12 \\
D-Flow        & \textbf{100} & 15.58 & 0.35 & 0.09 & \textbf{100} & 32.07 & 1.98e+02 & 0.74 & \textbf{100} & 39.62 & 0.31 & 0.08 & 14.2 & DNC & DNC & DNC \\
PCFM         & 0.2 & DNC & DNC & DNC & \textbf{100} & 2.15 & \underline{0.04} & \textbf{1.11e-08}$^\dagger$ & \textbf{100} & 1.29 & 0.19 & \textbf{1.19e-08}$^\dagger$ & \textbf{100} & 93.97 & 0.76 & \textbf{1.18e-10}$^\dagger$ \\
\textbf{EGT (Ours)} & \textbf{100} & \textbf{0.37} & \underline{0.02} & 0.05 & \textbf{100} & \textbf{1.13} & 0.05 & 0.07 & \textbf{100} & \textbf{0.48} & 0.21 & 0.05 & \textbf{100} & \textbf{0.71} & \textbf{0.53} & 0.01 \\
\textbf{Hybrid EGT (Ours)} & \textbf{100} & \underline{0.40} & \underline{0.02} & \underline{0.02} & \textbf{100} & \underline{1.24} & 0.05 & \underline{8.51e-03} & \textbf{100} & \underline{0.52} & 0.20 & \underline{0.01} & \textbf{100} & 0.90 & 0.55 & 1.41e-04 \\
\midrule
\multicolumn{17}{l}{\textit{Panel B: Full PDE Guidance (Stress Test)}} \\
\midrule
Vanilla FFM  & 100 & 0.14 & 0.08 & 6.24e-03 & 100 & 0.38 & 0.08 & 3.57e-03 & 100 & 0.27 & 0.20 & 5.13e-03 & 100 & 0.86 & 0.56 & 7.52e-03 \\
\midrule
ProFlow      & 0.4 & DNC & DNC & DNC & \textbf{100} & 25.05 & 5.00e+11 & 2.91e+05 & \textbf{100} & 18.68 & 0.24 & 6.27e-03 & \textbf{100} & 1.21 & \textbf{0.55} & 5.63e-03 \\
ECI          & \textbf{100} & 10.28 & \textbf{0.01} & \textbf{1.85e-03} & \textbf{100} & 24.34 & 0.25 & \underline{7.16e-05} & \textbf{100} & 18.68 & \textbf{0.12} & \underline{1.95e-04} & \textbf{100} & 4.40 & \underline{0.61} & 1.80e-03 \\
DiffusionPDE & \textbf{100} & \underline{0.69} & 18.15 & 0.99 & \textbf{100} & 1.86 & \textbf{0.02} & 6.58e-04 & \textbf{100} & 1.14 & \underline{0.15} & 9.46e-04 & \textbf{100} & 1.79 & \textbf{0.55} & 4.19e-03 \\
D-Flow        & \textbf{100} & 18.06 & 23.67 & 0.03 & \textbf{100} & 31.44 & 0.17 & 6.58e-04 & \textbf{100} & 32.70 & 0.35 & 2.91e-03 & 1.6 & DNC & DNC & DNC \\
PCFM         & 0   & DNC & DNC & DNC & \textbf{100} & 76.72 & 6.69e+16 & 1.34e+05 & \textbf{100} & 67.11 & 1.14e+17 & 4.76e+14 & 0 & DNC & DNC & DNC \\
\textbf{EGT (Ours)} & \textbf{100} & \textbf{0.24} & \underline{0.05} & 5.12e-03 & \textbf{100} & \textbf{0.89} & \underline{0.05} & 1.18e-03 & \textbf{100} & \textbf{0.46} & 0.19 & 1.92e-03 & \textbf{100} & \textbf{0.94} & \textbf{0.55} & \underline{1.63e-03} \\
\textbf{Hybrid EGT (Ours)} & \textbf{100} & \textbf{0.24} & \underline{0.05} & \underline{3.97e-03} & \textbf{100} & \underline{0.93} & \underline{0.05} & \textbf{1.95e-05} & \textbf{100} & \underline{0.54} & 0.19 & \textbf{1.49e-04} & \textbf{100} & \underline{1.02} & \textbf{0.55} & \textbf{2.89e-05} \\
\bottomrule
\end{tabular}%
}

\end{table*}

\begin{table*}[t]
\centering
\caption{Quantitative comparison across PDE datasets using the \textbf{Smooth Kernel} prior. \textit{Note$^\dagger$:} Evaluation on the 2D Navier-Stokes equations is omitted as the smooth kernel base distribution is currently not formulated for this specific environment. Values marked with $^\dagger$ \rev{fall below the discrete ground-truth stencil precision limit (stencil overfitting/waterbed effect)}; see Appendix \ref{app:waterbed}. Best results are highlighted in \textbf{bold}, and second-best are \underline{underlined} (excluding the unguided Vanilla).}
\label{tab:smooth_kernel_results}
\resizebox{\textwidth}{!}{%
\begin{tabular}{l | c c c c | c c c c | c c c c }
\toprule
\multirow{2}{*}{\textbf{Method}} & \multicolumn{4}{c|}{\textbf{Heat}} & \multicolumn{4}{c|}{\textbf{RD}} & \multicolumn{4}{c}{\textbf{Burgers}} \\
& SR$\uparrow$ & Spd$\downarrow$ & MSE$\downarrow$ & Phys$\downarrow$ & SR$\uparrow$ & Spd$\downarrow$ & MSE$\downarrow$ & Phys$\downarrow$ & SR$\uparrow$ & Spd$\downarrow$ & MSE$\downarrow$ & Phys$\downarrow$ \\
\midrule
\multicolumn{13}{l}{\textit{Panel A: Partial Physics Guidance (Boundary \& Initial Conditions)}} \\
\midrule
Vanilla FFM  & 100 & 0.19 & 0.26 & 0.43 & 100 & 0.38 & 0.09 & 0.22 & 100 & 0.41 & 0.16 & 0.13 \\
\midrule
ProFlow      & \textbf{100} & 8.16 & \underline{0.05} & 0.18 & \textbf{100} & 25.60 & 0.19 & 0.24 & \textbf{100} & 21.41 & 0.16 & 0.08 \\
ECI          & 0 & DNC & DNC & DNC & \textbf{100} & 24.43 & 0.36 & 0.09 & \textbf{100} & 21.94 & \underline{0.14} & 0.07 \\
DiffusionPDE & \textbf{100} & 0.95 & \textbf{0.04} & 0.06 & \textbf{100} & 1.98 & \textbf{9.37e-03} & 0.03 & \textbf{100} & 1.16 & \textbf{0.13} & 0.05 \\
D-Flow        & \textbf{100} & 15.64 & 24.89 & 0.97 & \textbf{100} & 31.43 & 0.09 & 0.03 & \textbf{100} & 39.29 & 10.58 & 0.37 \\
PCFM         & 1.1 & DNC & DNC & DNC & \textbf{100} & 2.11 & 0.09 & \textbf{1.11e-08}$^\dagger$ & \textbf{100} & 1.29 & 0.22 & \textbf{1.23e-08}$^\dagger$ \\
\textbf{EGT (Ours)} & \textbf{100} & \textbf{0.37} & \underline{0.05} & \underline{0.05} & \textbf{100} & \textbf{1.09} & 0.01 & 0.03 & \textbf{100} & \textbf{0.48} & \underline{0.14} & 0.02 \\
\textbf{Hybrid EGT (Ours)} & \textbf{100} & \underline{0.43} & \underline{0.05} & \textbf{0.02} & \textbf{100} & \underline{1.25} & \underline{9.96e-03} & \underline{8.48e-03} & \textbf{100} & \underline{0.52} & \underline{0.14} & \underline{4.58e-03} \\
\midrule
\multicolumn{13}{l}{\textit{Panel B: Full PDE Guidance (Stress Test)}} \\
\midrule
Vanilla FFM  & 100 & 0.20 & 0.25 & 0.17 & 100 & 0.37 & 0.09 & 3.63e-03 & 100 & 0.38 & 0.16 & 5.76e-03 \\
\midrule
ProFlow      & 0 & DNC & DNC & DNC & \textbf{100} & 25.10 & 0.19 & 4.04e-03 & \textbf{100} & 17.42 & 0.16 & 4.49e-03 \\
ECI          & 0 & DNC & DNC & DNC & \textbf{100} & 24.40 & 0.36 & 5.24e-04 & \textbf{100} & 17.08 & 0.14 & \underline{3.08e-04} \\
DiffusionPDE & \textbf{100} & 1.04 & \textbf{0.04} & 0.08 & \textbf{100} & 1.86 & \textbf{9.38e-03} & \underline{2.35e-04} & \textbf{100} & 1.74 & \textbf{0.12} & 1.46e-03 \\
D-Flow        & \textbf{100} & 16.79 & 17.62 & 0.66 & \textbf{100} & 31.31 & 1.53 & 1.60e-03 & \textbf{100} & 28.46 & 14.45 & 0.05 \\
PCFM         & 0 & DNC & DNC & DNC & \textbf{100} & 77.71 & 2.14e+17 & 5.91e+04 & \textbf{100} & 50.87 & 2.56e+16 & 5.74e+13 \\
\textbf{EGT (Ours)} & \textbf{100} & \textbf{0.52} & \textbf{0.04} & \underline{7.19e-03} & \textbf{100} & \textbf{0.89} & 9.80e-03 & 3.60e-04 & \textbf{100} & \textbf{0.95} & \underline{0.13} & 1.46e-03 \\
\textbf{Hybrid EGT (Ours)} & \textbf{100} & \underline{0.55} & \textbf{0.04} & \textbf{4.50e-03} & \textbf{100} & \underline{0.95} & \underline{9.79e-03} & \textbf{8.00e-05} & \textbf{100} & \underline{1.06} & \underline{0.13} & \textbf{1.85e-04}\\
\bottomrule
\end{tabular}%
}
\end{table*}

\textbf{Full PDE Guidance (The Stress Test).} \rev{Constraining generation to satisfy the complete spatiotemporal PDE presents a challenging regime for guidance methods. Proximal formulations like ProFlow show sensitivity in this regime, yielding a 0.4\% SR on the Heat equation and noticeable sample degradation on Reaction-Diffusion} (Sample MSE reaching $5.0\times 10^{11}$). Gradient-based optimization approaches (DiffusionPDE, D-Flow) technically avoid crashing on 1D grids but severely distort the data manifold, causing the Sample MSE on the Heat equation to spike to $18.15$ and $23.67$, respectively. For exact projection methods like PCFM, explicitly materializing the Jacobian for full-domain constraints results in Out-of-Memory errors or immediate numerical divergence ($0\%$ SR) on both Heat and Navier-Stokes systems. \rev{More broadly, comparing operating regimes across Tables~\ref{tab:main_results} and \ref{tab:smooth_kernel_results} reveals a clear structural trade-off: explicit projection methods (ECI, PCFM) excel in low-dimensional, projection-friendly regimes (e.g., 1D Heat or Burgers Sample MSE), but incur heavy computational costs ($4.40$s--$18.68$s for ECI) or numerical divergence when constraints become dense and non-linear.}
Conversely, our EGT framework, stabilized by Gradient Norm Clipping and dynamic scheduling, reliably scales to the Full-PDE regime \rev{while preserving sharp shockfronts and vortex structures without high-frequency artifacts or blurring, as visually confirmed on 1D Burgers (Figure~\ref{fig:qualitative_burgers}) and 2D Navier--Stokes (Figure~\ref{fig:qualitative_ns}).} EGT maintains a $100\%$ Success Rate globally with strictly sub-second inference speeds. On the chaotic 2D Navier-Stokes system, it safely navigates the complex physical manifold in just $0.94$s, reducing the physical residual to $1.63\times 10^{-3}$ (over $4\times$ lower than the Vanilla unguided model) while preserving the baseline distributional fidelity (Sample MSE $0.55$).

\textbf{The Smooth Kernel Prior: A Test of Robustness.}
To evaluate model stability against prior shifts, we replace standard Gaussian noise with a spatially correlated Smooth Kernel prior (Table \ref{tab:smooth_kernel_results}). Learning the transport map from correlated noise is empirically harder for the base network, leading to less accurate unguided trajectories.
\rev{Under this more challenging baseline landscape, projection- and proximal-based methods show increased sensitivity: PCFM, ECI, and ProFlow do not complete (0\% Success Rate) on the Full-PDE Heat equation. Furthermore, discrete corrections can compound early-trajectory inaccuracies, resulting in degraded sample quality} (Sample MSE reaching $2.1 \times 10^{17}$ for PCFM on Reaction-Diffusion). Among the other compared guidance techniques, only DiffusionPDE maintains consistent stability across these 1D tasks, whereas D-Flow suffers from heavily degraded sample quality (Sample MSE spiking to $17.62$ on the Heat equation).
Our continuous framework demonstrates robust corrective power under these adverse conditions. \rev{In the Heat Full-PDE benchmark, Hybrid EGT effectively guides the less accurate baseline trajectory}, reducing the physical residual by over an order of magnitude (from $0.17$ down to $4.50\text{e-}3$) and lowering the Sample MSE to $0.04$. This confirms that continuous energy guidance provides substantial resilience against both complex physical constraints and variations in the generative prior.
\subsection{Ablation Studies}
\label{sec:ablations}

\begin{wrapfigure}[14]{r}{0.46\textwidth}
\vspace{-45pt}
\centering
    \includegraphics[width=\linewidth, height=4.0cm, keepaspectratio]{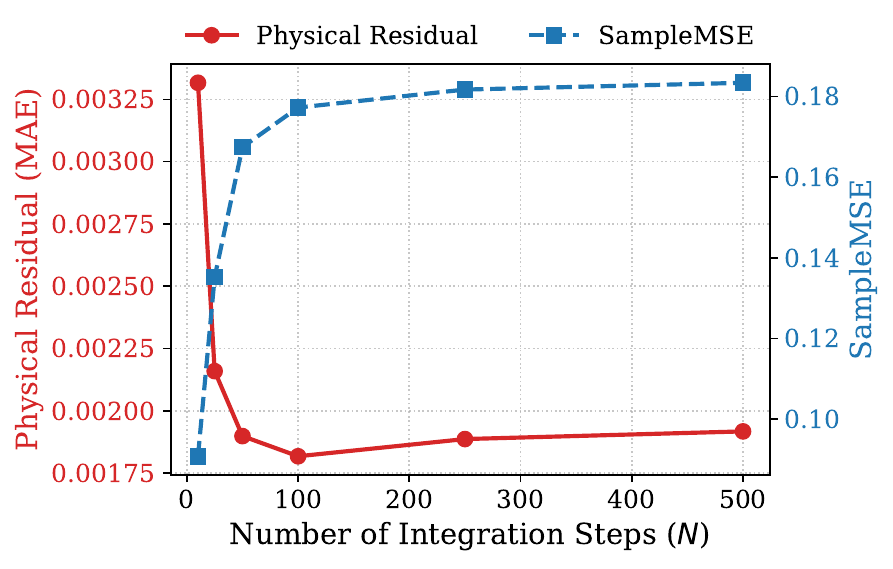}
    \vspace{-1pt}
    \caption{Impact of the number of ODE integration steps ($N$) on the 1D Burgers' equation (Full-PDE guidance). The dual-axis plot tracks Physical Residual MAE (red) and Sample MSE (blue). EGT stabilizes at low step counts ($N \approx 50$).}
    \label{fig:main_ablation}
\vspace{-10pt}
\end{wrapfigure}

To isolate the operational dynamics of our framework, we conduct targeted ablation studies. Figure \ref{fig:main_ablation} illustrates the sensitivity of Energy-Guided Transport (EGT) to the number of ODE integration steps ($N$) on the highly non-linear Full-PDE Burgers' equation, evaluated across 256 samples for $N \in \{10, 25, 50, 100, 250, 500\}$.
Notably, the analysis reveals that EGT achieves its optimal stability-accuracy trade-off at a significantly lower discretization budget (e.g., $N \approx 50$). For the primary evaluations in Section \ref{sec:experiments}, we deliberately adopted a higher, suboptimal integration budget (i.e., $N=200$) \rev{to maintain a standardized evaluation against projection-based baselines (e.g., PCFM), which generally require finer temporal discretizations for numerical stability.} This finding underscores that, in practical deployments, our continuous guidance can be substantially accelerated while concurrently improving distributional fidelity (Sample MSE). 

\rev{\textbf{Accuracy--Stability--Speed Pareto Analysis.} To further verify whether EGT improves the overall operational trade-off rather than representing a single static operating point, we perform a comprehensive Pareto analysis across guidance strengths ($\gamma$) and scheduling profiles on the 1D Burgers and 1D Heat systems (detailed in Appendix~\ref{app:ablations}). The results confirm that EGT lies on a highly favorable accuracy--stability--speed frontier: it strictly dominates Vanilla FFM and DiffusionPDE by simultaneously improving physical accuracy and distributional stability at $\sim 0.23$--$0.26$~s/sample. While ECI achieves lower absolute residuals in projection-friendly regimes, it incurs a $40$--$80\times$ higher computational cost. Importantly, EGT occupies a compact region in the runtime dimension, allowing practitioners to freely navigate the trade-off between physical residual and Sample MSE by tuning $\gamma$ or the scheduler at an essentially constant, sub-second inference time.}

\rev{\textbf{Isolating Continuous Transport vs. Terminal Refinement.} To verify that the performance of Hybrid EGT stems from our continuous transport rather than the terminal optimization itself, we apply the exact same post-hoc refinement procedure to the baseline trajectories on 1D Burgers and 2D Navier--Stokes (Table~\ref{tab:ablation_refinement} in Appendix~\ref{app:ablations}). On 2D Navier--Stokes, refining unconditional FFM samples only reduces the physical MAE from $7.50\text{e-}3$ to $3.96\text{e-}3$, whereas Pure EGT already reaches $1.60\text{e-}3$ prior to any refinement and drops to $2.90\text{e-}5$ when refined, confirming that continuous guidance steers trajectories into a markedly superior physical basin of attraction.}

\section{Conclusion}
\label{sec:conclusion}

In this work, we introduced Energy-Guided Transport (EGT), a zero-shot inference framework for physics-informed Functional Flow Matching. By formulating physical constraints as a smooth potential energy landscape, our approach maintains the continuous-time probability flow without requiring discrete geometric interruptions. {Our evaluations demonstrate that while explicit algebraic projections can face numerical instability and computational overhead in non-linear domains, continuous dissipative guidance provides a scalable and stable alternative.} We achieved up to a $130\times$ speedup over explicit projection techniques and a consistent $100\%$ success rate across chaotic environments like 2D Navier-Stokes, \rev{achieving a highly favorable balance between physical consistency, computational efficiency, and distributional fidelity.} Although evaluating the guidance field introduces a marginal per-step overhead compared to the vanilla unguided model, the framework's strict $\mathcal{O}(N)$ scalability ensures its applicability to arbitrary continuous domains.

\rev{\textbf{Limitations and Future Work.} While EGT provides a scalable alternative to explicit projections, several limitations define its operational scope. First, as a continuous dissipative mechanism rather than a hard algebraic constraint, EGT does not guarantee exact $E(u_1)=0$ convergence without optional terminal refinement, and relies on the pretrained FFM to provide a structurally reasonable transport prior. Second, excessively large guidance scales ($\gamma$) increase ODE stiffness (e.g., $\gamma > 30$ on 2D Navier--Stokes). Third, because our current energy functional minimizes pointwise $L^2$ PDE residuals, it does not explicitly constrain global macroscopic statistics such as shock gradient distributions or energy spectra (analyzed in Appendix~\ref{app:macroscopic_diagnostics}). Since EGT natively supports any differentiable objective and interacts with the domain geometry solely via $h(u)$, future work will incorporate spectral and conservation-law penalties into $E(u)$ and leverage geometry-aware operators (e.g., GNO, SFNO) to scale EGT to unstructured 3D turbulence.}

\clearpage
\bibliographystyle{unsrt}
\bibliography{references}

\begin{thebibliography}{10}

\bibitem{han2018solving}
Jiequn Han, Arnulf Jentzen, and Weinan E.
\newblock Solving high-dimensional partial differential equations using deep learning.
\newblock {\em Proceedings of the National Academy of Sciences}, 2018.

\bibitem{sirignano2018dgm}
Justin Sirignano and Konstantinos Spiliopoulos.
\newblock {DGM}: A deep learning algorithm for solving partial differential equations.
\newblock {\em Journal of Computational Physics}, 2018.

\bibitem{ho2020denoising}
Jonathan Ho, Ajay Jain, and Pieter Abbeel.
\newblock Denoising diffusion probabilistic models.
\newblock In {\em Advances in Neural Information Processing Systems}, 2020.

\bibitem{DBLP:conf/iclr/0011SKKEP21}
Yang Song, Jascha Sohl{-}Dickstein, Diederik~P. Kingma, Abhishek Kumar, Stefano Ermon, and Ben Poole.
\newblock Score-based generative modeling through stochastic differential equations.
\newblock In {\em International Conference on Learning Representations}, 2021.

\bibitem{DBLP:conf/iclr/LipmanCBNL23}
Yaron Lipman, Ricky T.~Q. Chen, Heli Ben{-}Hamu, Maximilian Nickel, and Matthew Le.
\newblock Flow matching for generative modeling.
\newblock In {\em International Conference on Learning Representations}, 2023.

\bibitem{DBLP:journals/jmlr/AlbergoBV25}
Michael~S. Albergo, Nicholas~M. Boffi, and Eric Vanden{-}Eijnden.
\newblock Stochastic interpolants: {A} unifying framework for flows and diffusions.
\newblock {\em Journal of Machine Learning Research}, 2025.

\bibitem{karniadakis2021physics}
George~Em Karniadakis, Ioannis~G Kevrekidis, Lu~Lu, Paris Perdikaris, Sifan Wang, and Liu Yang.
\newblock Physics-informed machine learning.
\newblock {\em Nature Reviews Physics}, 2021.

\bibitem{kochkov2021machine}
Dmitrii Kochkov, Jamie~A Smith, Ayya Alieva, Qing Wang, Michael~P Brenner, and Stephan Hoyer.
\newblock Machine learning--accelerated computational fluid dynamics.
\newblock {\em Proceedings of the National Academy of Sciences}, 2021.

\bibitem{brandstettermessage}
Johannes Brandstetter, Daniel~E Worrall, and Max Welling.
\newblock Message passing neural pde solvers.
\newblock In {\em International Conference on Learning Representations}, 2022.

\bibitem{brunton2020machine}
Steven~L Brunton, Bernd~R Noack, and Petros Koumoutsakos.
\newblock Machine learning for fluid mechanics.
\newblock {\em Annual Review of Fluid Mechanics}, 2020.

\bibitem{DBLP:journals/nature/BiXZCG023}
Kaifeng Bi, Lingxi Xie, Hengheng Zhang, Xin Chen, Xiaotao Gu, and Qi~Tian.
\newblock Accurate medium-range global weather forecasting with {3D} neural networks.
\newblock {\em Nature}, 2023.

\bibitem{lam2023learning}
Remi Lam, Alvaro Sanchez-Gonzalez, Matthew Willson, Peter Wirnsberger, Meire Fortunato, Ferran Alet, Suman Ravuri, Timo Ewalds, Zach Eaton-Rosen, Weihua Hu, et~al.
\newblock Learning skillful medium-range global weather forecasting.
\newblock {\em Science}, 2023.

\bibitem{pathak2022fourcastnet}
Jaideep Pathak, Shashank Subramanian, Peter Harrington, Sanjeev Raja, Ashesh Chattopadhyay, Morteza Mardani, Thorsten Kurth, David Hall, Zongyi Li, Kamyar Azizzadenesheli, et~al.
\newblock Fourcastnet: A global data-driven high-resolution weather model using adaptive fourier neural operators.
\newblock {\em arXiv preprint arXiv:2202.11214}, 2022.

\bibitem{kashinath2021physics}
C~Jiang, R~Wang, et~al.
\newblock Physics-informed machine learning: case studies for weather and climate modelling.
\newblock {\em Philosophical Transactions of the Royal Society A: Mathematical, Physical and Engineering Sciences}, 2021.

\bibitem{DBLP:conf/kdd/WangKMAY20}
Rui Wang, Karthik Kashinath, Mustafa Mustafa, Adrian Albert, and Rose Yu.
\newblock Towards physics-informed deep learning for turbulent flow prediction.
\newblock In {\em Conference on Knowledge Discovery and Data Mining}, 2020.

\bibitem{cai2021physics}
Shengze Cai, Zhiping Mao, Zhicheng Wang, Minglang Yin, and George~Em Karniadakis.
\newblock Physics-informed neural networks (pinns) for fluid mechanics: A review.
\newblock {\em Acta Mechanica Sinica}, 2021.

\bibitem{DBLP:conf/icml/HansenMAGM23}
Derek Hansen, Danielle~C. Maddix, Shima Alizadeh, Gaurav Gupta, and Michael~W. Mahoney.
\newblock Learning physical models that can respect conservation laws.
\newblock In {\em International Conference on Machine Learning}, 2023.

\bibitem{mouli2024using}
S~Chandra Mouli, Danielle~C Maddix, Shima Alizadeh, Gaurav Gupta, Andrew Stuart, Michael~W Mahoney, and Yuyang Wang.
\newblock Using uncertainty quantification to characterize and improve out-of-domain learning for {PDEs}.
\newblock In {\em International Conference on Machine Learning}, 2024.

\bibitem{herde2024poseidon}
Maximilian Herde, Bogdan Raoni{\'c}, Tobias Rohner, Roger K{\"a}ppeli, Roberto Molinaro, Emmanuel De~Bezenac, and Siddhartha Mishra.
\newblock Poseidon: Efficient foundation models for {PDEs}.
\newblock In {\em Advances in Neural Information Processing Systems}, 2024.

\bibitem{batzolis2021conditional}
Georgios Batzolis, Jan Stanczuk, Carola-Bibiane Sch{\"o}nlieb, and Christian Etmann.
\newblock Conditional image generation with score-based diffusion models.
\newblock {\em arXiv preprint arXiv:2111.13606}, 2021.

\bibitem{song2021maximum}
Yang Song, Conor Durkan, Iain Murray, and Stefano Ermon.
\newblock Maximum likelihood training of score-based diffusion models.
\newblock In {\em Advances in Neural Information Processing Systems}, 2021.

\bibitem{deng2021continuous}
Ruizhi Deng, Marcus~A Brubaker, Greg Mori, and Andreas Lehrmann.
\newblock Continuous latent process flows.
\newblock In {\em Advances in Neural Information Processing Systems}, 2021.

\bibitem{utkarshphysics}
Utkarsh Utkarsh, Pengfei Cai, Alan Edelman, Rafael Gomez-Bombarelli, and Christopher~Vincent Rackauckas.
\newblock Physics-constrained flow matching: Sampling generative models with hard constraints.
\newblock In {\em Advances in Neural Information Processing Systems}, 2025.

\bibitem{liuflow}
Xingchao Liu, Chengyue Gong, et~al.
\newblock Flow straight and fast: Learning to generate and transfer data with rectified flow.
\newblock In {\em International Conference on Learning Representations}, 2023.

\bibitem{peyre2019computational}
Gabriel Peyr{\'e} and Marco Cuturi.
\newblock {\em Computational optimal transport: With applications to data science}.
\newblock Now Foundations and Trends, 2019.

\bibitem{DBLP:journals/tmlr/0001FMHZRWB24}
Alexander Tong, Kilian Fatras, Nikolay Malkin, Guillaume Huguet, Yanlei Zhang, Jarrid Rector{-}Brooks, Guy Wolf, and Yoshua Bengio.
\newblock Improving and generalizing flow-based generative models with minibatch optimal transport.
\newblock {\em Transactions on Machine Learning Research}, 2024.

\bibitem{DBLP:conf/icml/PooladianBDALC23}
Aram{-}Alexandre Pooladian, Heli Ben{-}Hamu, Carles Domingo{-}Enrich, Brandon Amos, Yaron Lipman, and Ricky T.~Q. Chen.
\newblock Multisample flow matching: Straightening flows with minibatch couplings.
\newblock In {\em International Conference on Machine Learning}, 2023.

\bibitem{chen2018neural}
Ricky~TQ Chen, Yulia Rubanova, Jesse Bettencourt, and David~K Duvenaud.
\newblock Neural ordinary differential equations.
\newblock In {\em Advances in Neural Information Processing Systems}, 2018.

\bibitem{DBLP:conf/iclr/ChenL24}
Ricky T.~Q. Chen and Yaron Lipman.
\newblock Flow matching on general geometries.
\newblock In {\em International Conference on Learning Representations}, 2024.

\bibitem{mathieu2020riemannian}
Emile Mathieu and Maximilian Nickel.
\newblock Riemannian continuous normalizing flows.
\newblock In {\em Advances in Neural Information Processing Systems}, 2020.

\bibitem{gemici2016normalizing}
Mevlana~C Gemici, Danilo Rezende, and Shakir Mohamed.
\newblock Normalizing flows on riemannian manifolds.
\newblock {\em arXiv preprint arXiv:1611.02304}, 2016.

\bibitem{DBLP:conf/nips/BortoliMHTTD22}
Valentin~De Bortoli, Emile Mathieu, Michael~J. Hutchinson, James Thornton, Yee~Whye Teh, and Arnaud Doucet.
\newblock Riemannian score-based generative modelling.
\newblock In {\em Advances in Neural Information Processing Systems}, 2022.

\bibitem{DBLP:conf/iclr/LiKALBSA21}
Zongyi Li, Nikola~Borislavov Kovachki, Kamyar Azizzadenesheli, Burigede Liu, Kaushik Bhattacharya, Andrew~M. Stuart, and Anima Anandkumar.
\newblock Fourier neural operator for parametric partial differential equations.
\newblock In {\em International Conference on Learning Representations}, 2021.

\bibitem{lu2021learning}
Lu~Lu, Pengzhan Jin, Guofei Pang, Zhongqiang Zhang, and George~Em Karniadakis.
\newblock Learning nonlinear operators via deeponet based on the universal approximation theorem of operators.
\newblock {\em Nature Machine Intelligence}, 2021.

\bibitem{kovachki2023neural}
Nikola Kovachki, Zongyi Li, Burigede Liu, Kamyar Azizzadenesheli, Kaushik Bhattacharya, Andrew Stuart, and Anima Anandkumar.
\newblock Neural operator: Learning maps between function spaces with applications to {PDEs}.
\newblock {\em Journal of Machine Learning Research}, 2023.

\bibitem{li2024physics}
Zongyi Li, Hongkai Zheng, Nikola Kovachki, David Jin, Haoxuan Chen, Burigede Liu, Kamyar Azizzadenesheli, and Anima Anandkumar.
\newblock Physics-informed neural operator for learning partial differential equations.
\newblock {\em ACM/IMS Journal of Data Science}, 2024.

\bibitem{lim2025score}
Jae~Hyun Lim, Nikola~B Kovachki, Ricardo Baptista, Christopher Beckham, Kamyar Azizzadenesheli, Jean Kossaifi, Vikram Voleti, Jiaming Song, Karsten Kreis, Jan Kautz, et~al.
\newblock Score-based diffusion models in function space.
\newblock {\em Journal of Machine Learning Research}, 2025.

\bibitem{DBLP:conf/aistats/KerriganMS24}
Gavin Kerrigan, Giosue Migliorini, and Padhraic Smyth.
\newblock Functional flow matching.
\newblock In {\em International Conference on Artificial Intelligence and Statistics}, 2024.

\bibitem{DBLP:journals/jcphy/RaissiPK19}
Maziar Raissi, Paris Perdikaris, and George~E. Karniadakis.
\newblock Physics-informed neural networks: {A} deep learning framework for solving forward and inverse problems involving nonlinear partial differential equations.
\newblock {\em Journal of Computational Physics}, 2019.

\bibitem{raissi2020hidden}
Maziar Raissi, Alireza Yazdani, and George~Em Karniadakis.
\newblock Hidden fluid mechanics: Learning velocity and pressure fields from flow visualizations.
\newblock {\em Science}, 2020.

\bibitem{DBLP:journals/jcphy/JinCLK21}
Xiaowei Jin, Shengze Cai, Hui Li, and George~Em Karniadakis.
\newblock {NSFnets} ({Navier-Stokes} flow nets): Physics-informed neural networks for the incompressible navier-stokes equations.
\newblock {\em Journal of Computational Physics}, 2021.

\bibitem{krishnapriyan2021characterizing}
Aditi Krishnapriyan, Amir Gholami, Shandian Zhe, Robert Kirby, and Michael~W Mahoney.
\newblock Characterizing possible failure modes in physics-informed neural networks.
\newblock In {\em Advances in Neural Information Processing Systems}, 2021.

\bibitem{wang2021understanding}
Sifan Wang, Yujun Teng, and Paris Perdikaris.
\newblock Understanding and mitigating gradient flow pathologies in physics-informed neural networks.
\newblock {\em SIAM Journal on Scientific Computing}, 2021.

\bibitem{ho2021classifier}
Jonathan Ho and Tim Salimans.
\newblock Classifier-free diffusion guidance.
\newblock In {\em NeurIPS 2021 Workshop on Deep Generative Models and Downstream Applications}, 2021.

\bibitem{DBLP:conf/iclr/BansalCSSGGG24}
Arpit Bansal, Hong{-}Min Chu, Avi Schwarzschild, Soumyadip Sengupta, Micah Goldblum, Jonas Geiping, and Tom Goldstein.
\newblock Universal guidance for diffusion models.
\newblock In {\em International Conference on Learning Representations}, 2024.

\bibitem{DBLP:conf/nips/HuangYWP24}
Jiahe Huang, Guandao Yang, Zichen Wang, and Jeong~Joon Park.
\newblock Diffusionpde: Generative {PDE}-solving under partial observation.
\newblock In {\em Advances in Neural Information Processing Systems}, 2024.

\bibitem{DBLP:conf/icml/Ben-HamuPGKSL24}
Heli Ben{-}Hamu, Omri Puny, Itai Gat, Brian Karrer, Uriel Singer, and Yaron Lipman.
\newblock D-flow: Differentiating through flows for controlled generation.
\newblock In {\em International Conference on Machine Learning}, 2024.

\bibitem{chenggradient}
Chaoran Cheng, Boran Han, Danielle~C Maddix, Abdul~Fatir Ansari, Andrew Stuart, Michael~W Mahoney, and Bernie Wang.
\newblock Gradient-free generation for hard-constrained systems.
\newblock In {\em International Conference on Learning Representations}, 2025.

\bibitem{yu2026ProFlow}
Zichao Yu, Ming Li, Wenyi Zhang, Difan Zou, and Weiguo Gao.
\newblock Proflow: Zero-shot physics-consistent sampling via proximal flow guidance.
\newblock {\em arXiv preprint arXiv:2601.20227}, 2026.

\bibitem{evans2022partial}
Lawrence~C Evans.
\newblock {\em Partial Differential Equations}.
\newblock American Mathematical Society, 2022.

\bibitem{greydanus2019hamiltonian}
Samuel Greydanus, Misko Dzamba, and Jason Yosinski.
\newblock Hamiltonian neural networks.
\newblock In {\em Advances in Neural Information Processing Systems}, 2019.

\bibitem{richter2022neural}
Jack Richter-Powell, Yaron Lipman, and Ricky~TQ Chen.
\newblock Neural conservation laws: A divergence-free perspective.
\newblock In {\em Advances in Neural Information Processing Systems}, 2022.

\bibitem{holllearning}
Philipp Holl, Nils Thuerey, and Vladlen Koltun.
\newblock Learning to control pdes with differentiable physics.
\newblock In {\em International Conference on Learning Representations}, 2020.

\end{thebibliography}

\clearpage
\appendix
\section{Proofs of Energy Dissipation and Error Bounds}
\label{app:dissipation_proofs}

In this Appendix, we provide the formal proofs for the Dissipation Proposition and its associated corollaries presented in Section \ref{sec:continuous_guidance}. The derivations are formulated within the deterministic Flow Matching framework, utilizing a Lagrangian (pathwise) perspective to track the continuous probability mass transport in infinite-dimensional spaces.

\subsection{Proof of Proposition 1: Energy Dissipation (Ideal Case)}
\label{app:proof_ideal}

Let $\mathcal{H}$ be the infinite-dimensional Hilbert space representing the state space of the generated physical fields. Under the Functional Flow Matching (FFM) paradigm, the continuous temporal evolution of a generated state is governed by a deterministic ordinary differential equation (ODE) $\dot{u}_t = \bar{v}_t(u_t)$. Let $\Phi_t: \mathcal{H} \to \mathcal{H}$ denote the corresponding flow map, such that $u_t = \Phi_t(u_0)$ with the initial state drawn from the base prior $u_0 \sim p_0$.

In our continuous guidance framework, the augmented velocity field is defined as:
\begin{equation}
    \bar{v}_t(u) = v_\theta(u, t) - \gamma(t)\nabla_u E(u)
\end{equation}
where $\nabla_u E(u)$ denotes the Fréchet derivative (gradient) of the physical energy functional with respect to the field $u$. 

We assume that the neural vector field $v_\theta$ and the energy gradient $\nabla_u E(u)$ satisfy mild local Lipschitz conditions such that the ODE admits a unique global solution $u_t \in \mathcal{H}$ almost everywhere. Furthermore, we assume sufficient regularity to permit the exchange of expectations and time derivatives via the Dominated Convergence Theorem.

We define the expected physical energy at time $t$ over the marginal density $p_t(u)$. Using the Law of the Unconscious Statistician, we express this expectation by pulling it back to the base prior $p_0$ via the flow map:
\begin{equation}
    \mathbb{E}_{p_t}[E(u)] = \mathbb{E}_{p_0}[E(\Phi_t(u_0))]
\end{equation}

Applying the functional chain rule in $\mathcal{H}$, the derivative of the expected energy with respect to the generative time $t$ is:
\begin{equation}
    \frac{d}{dt}\mathbb{E}_{p_0}[E(\Phi_t(u_0))] = \mathbb{E}_{p_0}\left[ \left\langle \nabla_u E(\Phi_t(u_0)), \frac{d}{dt}\Phi_t(u_0) \right\rangle_{\mathcal{H}} \right]
\end{equation}

Substituting the ODE $\frac{d}{dt}\Phi_t(u_0) = \bar{v}_t(\Phi_t(u_0))$ and pushing the expectation forward back to the marginal distribution $p_t$, we obtain the instantaneous variation of the expected energy:
\begin{equation}
    \frac{d}{dt}\mathbb{E}_{p_t}[E(u)] = \mathbb{E}_{p_t}\left[ \langle \nabla_u E(u), \bar{v}_t(u) \rangle_{\mathcal{H}} \right]
\end{equation}

Substituting the explicit formulation of our augmented vector field into the inner product:
\begin{align}
    \frac{d}{dt}\mathbb{E}_{p_t}[E(u)] &= \mathbb{E}_{p_t}\left[ \langle \nabla_u E(u), v_\theta(u, t) - \gamma(t)\nabla_u E(u) \rangle_{\mathcal{H}} \right] \\
    &= \mathbb{E}_{p_t}\left[ \langle \nabla_u E(u), v_\theta(u, t) \rangle_{\mathcal{H}} \right] - \gamma(t) \mathbb{E}_{p_t}\left[ \| \nabla_u E(u) \|^2_{\mathcal{H}} \right]
\end{align}

Let the first term be defined as $\frac{d\mathbb{E}_{\text{base}}}{dt} \triangleq \mathbb{E}_{p_t}[ \langle \nabla_u E(u), v_\theta(u, t) \rangle_{\mathcal{H}} ]$, representing the natural energy variation induced by the unguided model. Since $\gamma(t) \geq 0$ and the expected squared norm is strictly non-negative, the second term acts as a strictly dissipative force:
\begin{equation}
    \frac{d\mathbb{E}_{\text{total}}}{dt} \leq \frac{d\mathbb{E}_{\text{base}}}{dt}
\end{equation}
This concludes the proof.

\subsection{Proof of Corollary 1: Error-Bounded Dissipation}
\label{app:proof_prop1}

In practice, the neural vector field $v_\theta(u, t)$ is an approximation of the ideal target field $v^*(u, t)$. We define the approximation error as $\delta(u, t) = v_\theta(u, t) - v^*(u, t)$. In the context of Flow Matching, the network is trained to minimize the expected mean squared error; thus, we bound the error in $L^2(p_t)$ norm: $\epsilon_t^2 \triangleq \mathbb{E}_{p_t}[\|\delta(u, t)\|^2_{\mathcal{H}}]$.

Substituting $v_\theta(u, t) = v^*(u, t) + \delta(u, t)$ into the expectation of the base variation defined in Proposition 1:
\begin{equation}
    \mathbb{E}_{p_t}\left[ \langle \nabla_u E(u), v_\theta(u, t) \rangle_{\mathcal{H}} \right] = \mathbb{E}_{p_t}\left[ \langle \nabla_u E(u), v^*(u, t) \rangle_{\mathcal{H}} \right] + \mathbb{E}_{p_t}\left[ \langle \nabla_u E(u), \delta(u, t) \rangle_{\mathcal{H}} \right]
\end{equation}

We now bound the spurious error term by applying the generalized Cauchy-Schwarz inequality for expectations:
\begin{equation}
    \mathbb{E}_{p_t}\left[ \langle \nabla_u E(u), \delta(u, t) \rangle_{\mathcal{H}} \right] \leq \sqrt{\mathbb{E}_{p_t}\left[ \|\nabla_u E(u)\|^2_{\mathcal{H}} \right]} \sqrt{\mathbb{E}_{p_t}\left[ \|\delta(u, t)\|^2_{\mathcal{H}} \right]}
\end{equation}

Defining the expected gradient energy landscape as $V(t) \triangleq \mathbb{E}_{p_t}[\|\nabla_u E(u)\|^2_{\mathcal{H}}]$, we obtain:
\begin{equation}
    \mathbb{E}_{p_t}\left[ \langle \nabla_u E(u), \delta(u, t) \rangle_{\mathcal{H}} \right] \leq \sqrt{V(t)} \epsilon_t
\end{equation}

Substituting this bound back into the full dissipation equation establishes Corollary 1:
\begin{equation}
    \frac{d\mathbb{E}_{\text{total}}}{dt} \leq \frac{d\mathbb{E}_{v^*}}{dt} - \gamma(t) V(t) + \sqrt{V(t)}\epsilon_t
\end{equation}

\subsection{Proof of Corollary 2: Global Energy Descent}
\label{app:proof_prop2}

To evaluate the macroscopic effect of the continuous guidance, we integrate the instantaneous rate of change derived in Corollary 1 over the entire generation process, from $t=0$ to $t=1$:
\begin{equation}
    \int_0^1 \frac{d}{dt}\mathbb{E}_{p_t}[E(u)] dt \leq \int_0^1 \frac{d\mathbb{E}_{v^*}}{dt} dt - \int_0^1 \gamma(t) V(t) dt + \int_0^1 \sqrt{V(t)}\epsilon_t dt
\end{equation}

By the Fundamental Theorem of Calculus, the left-hand side evaluates exactly to the total change in expected physical energy:
\begin{equation}
    \mathbb{E}_{p_1}[E(u_1)] - \mathbb{E}_{p_0}[E(u_0)] 
\end{equation}

Defining the cumulative natural energy variation under the ideal transport field as $\Delta E_{\text{base}} = \int_0^1 \frac{d\mathbb{E}_{v^*}}{dt} dt$, and the exact cumulative neural approximation error as $\mathcal{E}_{\text{err}} = \int_0^1 \sqrt{V(t)}\epsilon_t dt$, we obtain Corollary 2:
\begin{equation}
    \mathbb{E}_{p_1}[E(u_1)] - \mathbb{E}_{p_0}[E(u_0)] \leq \Delta E_{\text{base}} - \int_0^1 \gamma(t) V(t) dt + \mathcal{E}_{\text{err}}
\end{equation}

This confirms that the cumulative dissipative force must be scaled via the scheduling parameter $\gamma(t)$ to robustly mitigate the cumulative model error $\mathcal{E}_{\text{err}}$, ensuring a net reduction in the overall PDE violation. Note that if the neural approximation error is uniformly bounded by $\sup_t \epsilon_t \leq \epsilon$, the cumulative error scales as $\mathcal{O}(\epsilon)$.
while the physical energy landscape $E(u)$ may be highly non-convex for non-linear PDEs (as discussed in Appendix~\ref{app:taxonomy}), the local dissipativity guaranteed by the inner product $\langle \nabla_u E, -\nabla_u E \rangle \leq 0$ holds universally. Unlike explicit algebraic projections that diverge in non-convex topologies, our continuous gradient injection safely ensures energy descent along the probability path without requiring global convexity.

\section{Mathematical Taxonomy of Constraints}
\label{app:taxonomy}

Integrating physical knowledge into zero-shot generative models requires defining a set of constraint functions $h(u) = 0$. As referenced in Section \ref{sec:problem_setup}, we categorize these constraints based on their spatial scale, linearity, and the topological complexity of their resulting energy landscape $E(u) = \frac{1}{2}\|h(u)\|^2$. Table \ref{tab:theoretical_constraints} summarizes the mathematical properties of the constraints evaluated in this work.

\begin{enumerate}
    \item \textbf{Initial and Boundary Conditions (IC/BC):} These constraints define the admissibility and uniqueness of the solution. While essential, they only constrain the field at $t=0$ or on the spatial boundary $\partial\Omega$, leaving the internal spatio-temporal evolution entirely to the learned prior of the generative model. Mathematically, they are typically formulated as linear constraints (e.g., Dirichlet or standard Neumann conditions), though non-linear boundary physics can occur.
    
    \item \textbf{Macroscopic Constraints (Conservation Laws):} Often expressed as global integrals over the domain, such as $\int_{\Omega} u(\mathbf{x},t) d\mathbf{x} = C$, these constraints enforce the conservation of macroscopic invariants (e.g., total mass or momentum). Because the integral operator is a strictly linear functional mapping, its associated energy landscape is strictly convex. This guarantees that explicit algebraic projection methods can find a unique, global minimum efficiently. However, acting globally, they lack the spatial resolution required to correct local structural dynamics.
    
    \item \textbf{Full PDE Residual:} This represents the most rigorous microscopic constraint, requiring the governing differential equation to hold locally at every coordinate $(\mathbf{x},t)$. The topological complexity of this constraint is entirely dictated by the nature of the spatial differential operator $\mathcal{N}(u)$. For linear differential equations (e.g., the Heat equation), the resulting energy functional remains strictly convex. Conversely, for non-linear advective systems (e.g., Burgers' or Navier-Stokes equations), the residual forms a highly non-convex energy landscape. In these non-linear regimes, exact explicit projections become computationally intractable and structurally fragile, necessitating the continuous dissipative guidance proposed in our framework.
\end{enumerate}

\begin{table}[h]
    \centering
    \resizebox{\textwidth}{!}{%
    \begin{tabular}{@{}lllll@{}}
    \toprule
    \textbf{Constraint Type} & \textbf{Analytical Form $h(u) = 0$} & \textbf{Spatial Scope} & \textbf{Linearity / Landscape} & \textbf{Generative Impact} \\ \midrule
    \textbf{Initial/Boundary} & $\mathcal{B}(u) - g(\mathbf{x},t) = 0$ & $\partial\Omega \cup \{t=0\}$ & Typically Linear (Convex) & Anchors the domain; ensures uniqueness. \\ \addlinespace
    \textbf{Macroscopic} & $\int_\Omega u \, d\mathbf{x} - C = 0$ & Global ($\Omega$) & Strictly Linear (Convex) & Enforces global invariants; blind to local topology. \\ \addlinespace
    \textbf{Full PDE} & $\partial_t u + \mathcal{N}(u) = 0$ & Microscopic ($\forall \mathbf{x}, t$) & System-Dependent & Dictates true local dynamics; implicitly ensures global conservation. \\ \bottomrule
    \end{tabular}%
    }
    \vspace{0.2cm}
    \caption{Theoretical taxonomy of physical constraints in generative modeling. The structural linearity of the physical operator dictates the convexity of the resulting energy landscape. This topological property drastically impacts the computational feasibility of applying exact algebraic corrections during the inference process.}
    \label{tab:theoretical_constraints} 
\end{table}

\section{Dataset Discretization and Numerical Generation}
\label{app:datasets}

To ensure a fair and rigorous baseline comparison, all datasets used in this work follow the exact numerical generation protocols established by the Physics-Constrained Flow Matching (PCFM) framework~\cite{utkarshphysics}. The datasets are mathematically identical to those in PCFM, ensuring that performance gains stem strictly from the generative methodology.

\paragraph{1. Heat Equation (1D):} 
We consider the linear diffusion equation $\partial_t u = \alpha \partial_{xx} u$. Instead of a numerical approximation, we generate samples using the exact analytical solution:
\begin{equation}
    u(x,t) = \sin(x + \varphi) e^{-\alpha t}
\end{equation}
where the diffusion coefficient is $\alpha \sim \mathcal{U}(1.0, 5.0)$ and the phase is $\varphi \sim \mathcal{U}(0, \pi)$. The domain is discretized on a $100 \times 100$ spatiotemporal grid ($x \in [0, 2\pi], t \in [0, 1]$) with periodic boundary conditions. This provides a high-fidelity ground truth with zero discretization error.

\paragraph{2. Reaction-Diffusion (1D):} 
This system is governed by the non-linear equation $\partial_t u - \nu \partial_{xx} u - \rho u(1 - u) = 0$, with parameters set to $\rho = 0.01$ and $\nu = 0.005$. Data is generated via a semi-implicit finite difference scheme on a $128 \times 100$ grid over $x \in [0, 1]$. We construct 6,400 solutions by pairing 80 initial conditions with 80 boundary conditions.

\paragraph{3. Inviscid Burgers' Equation (1D):} 
We solve the inviscid form $\partial_t u + u \partial_x u = 0$, which is highly prone to shock formation. To handle discontinuities, we employ a Godunov finite volume method. The initial conditions are smoothed step functions $u(x,0) = (1 + \exp((x - p)/\epsilon))^{-1}$ with steepness $\epsilon=0.02$ and the center location sampled uniformly as $p \sim \mathcal{U}(0.2, 0.8)$. The grid is $101 \times 101$ over $x,t \in [0, 1]$.

\paragraph{4. Navier-Stokes Equations (2D):} 
We use the incompressible Navier-Stokes equations in vorticity form: $\partial_t \omega + (u \cdot \nabla) \omega = \nu \Delta \omega + f$. We set the viscosity to $\nu = 10^{-3}$ and use a Crank-Nicolson spectral solver on a $64 \times 64$ periodic grid, recording 50 temporal snapshots. The dataset comprises 10,000 simulations generated by sampling 100 random initial vorticities and 100 forcing phases.

\section{Generative Kernels and Prior Distributions}
\label{app:kernels}

\begin{figure*}[h]
    \centering
    \includegraphics[width=\textwidth]{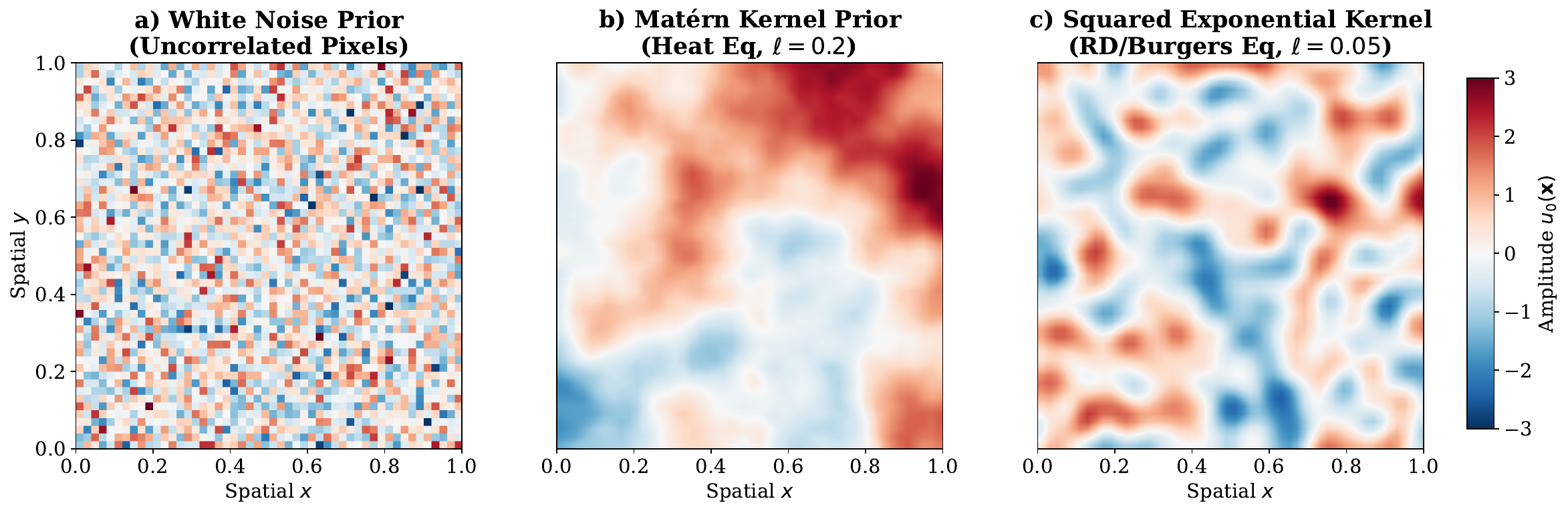}
    \caption{Qualitative 2D visual comparison of standard samples drawn from different generative functional prior distributions $p_0(u)$, visualized as spatial maps over the domain $[0,1] \times [0,1]$. Panel a) shows white noise, characterized by uncorrelated pixels that create local high-frequency content. Panels b) and c) show functional smooth kernel priors (Gaussian Random Fields), highlighting how the characteristic length scale $\ell$ dictates the typical size and smoothness of initial functional structures.}
    \label{fig:kernels_comparison}
\end{figure*}

The choice of the prior distribution $p_0(u)$ is critical in Functional Flow Matching, as it defines the starting point of the probability flow and dictates the initial geometry of the sample paths. While white noise is the standard setting for discrete diffusion models, embedding spatial smoothness directly into the functional prior is fundamental for the stable evaluation of high-order spatial physical operators. In our experiments, we compare two distinct classes of priors:

\paragraph{1. White Noise Prior:} 
This represents the uncorrelated generative setting, where the initial state is sampled as independent and identically distributed (i.i.d.) Gaussian noise across all spatial grid points, effectively mimicking $u_0 \sim \mathcal{N}(0, \mathbf{I})$. For the 2D Navier-Stokes equations, we exclusively utilize this prior due to the extreme dimensionality of the full spatio-temporal grid ($64 \times 64 \times 50$), which makes the instantiation and inversion of full dense covariance matrices computationally prohibitive for spatially correlated priors. Visualized in Figure \ref{fig:kernels_comparison}a, it presents as high-frequency noise lacking any spatial structure.

\paragraph{2. Smooth Kernel Prior (Gaussian Random Fields):} 
To investigate the impact of spatial correlation on continuous physical guidance, we employ Gaussian Random Fields (GRFs) as functional priors for the 1D equations. Unlike white noise, GRFs ensure that the initial state $u_0$ possesses a predefined degree of spatial correlation, guaranteeing spatial differentiability right from $t=0$. This property provides a well-behaved energy landscape for guiding derivatives ($\partial_x, \partial_{xx}$). 
\begin{itemize}
    \item \textbf{Heat Equation:} We utilize a Matérn covariance kernel ($\nu=3/2$) with a characteristic correlation length $\ell = 0.2$, promoting smoother and larger initial spatial features (Figure \ref{fig:kernels_comparison}b).
    \item \textbf{Reaction-Diffusion and Burgers' Equations:} We utilize a Squared Exponential kernel with a tighter correlation length $\ell = 0.05$ (Figure \ref{fig:kernels_comparison}c), allowing for the representation of sharper initial functional shapes.
\end{itemize}

Figure \ref{fig:kernels_comparison} provides a qualitative 2D visualization of samples drawn from these functional prior distributions, clearly highlighting how the characteristic correlation length dictates the smoothness and frequency spectrum of the initial generative state.

\section{Architecture and Training Details}
\label{app:architecture}

\paragraph{FNO Backbone.} All generative models in this work parameterize the underlying time-dependent vector field $v_\theta(u_t, t)$ using a Fourier Neural Operator (FNO) backbone. The FNO encoder takes as input the concatenation of the current state $u_t$, a sinusoidal positional embedding for the spatial grid, and a Fourier time embedding to capture the non-stationary nature of the probability flow. 

To maintain strict consistency with the baseline setups, we tailor the architecture capacity to the dimensionality of the problem:
\begin{itemize}
    \item \textbf{1D Systems (Heat, Reaction-Diffusion, Burgers):} The architecture consists of 4 Fourier layers preserving 32 modes across spatial and temporal dimensions, with 64 hidden channels and 256 projection channels.
    \item \textbf{2D Navier-Stokes:} Due to the high-dimensional nature of the $64 \times 64 \times 50$ spatio-temporal grid, we utilize a 3D FNO equipped with 2 Fourier layers, preserving 16 modes per dimension, with 32 hidden channels and 256 projection channels.
\end{itemize}

\paragraph{Training Regimen.} We pretrained 7 distinct Functional Flow Matching (FFM) models to comprehensively evaluate our continuous guidance against unconstrained baselines under different prior assumptions (White Noise vs. Smooth GRFs). All models were trained using the Adam optimizer ($\beta_1 = 0.9, \beta_2 = 0.999$, zero weight decay) with an initial learning rate of $3 \times 10^{-4}$. We employed a Reduce-LROnPlateau scheduler with a decay factor of 0.5, a patience of 10 validation steps, and a minimum learning rate of $1 \times 10^{-4}$. 

The batch sizes were set to 256 for the 1D environments and 24 for the 2D Navier-Stokes environment. Training was performed on clusters equipped with NVIDIA RTX 8000, A6000 and L40 GPUs. The 1D models converged after 20,000 iterations, while the Navier-Stokes model required 500,000 training steps to adequately capture the complex fluid dynamics. Table \ref{tab:hyperparameters} summarizes the specific configurations for each of the 7 pretrained models. 

\paragraph{Computational Bottleneck of Spatially Correlated Priors.} While smooth GRFs were utilized for 1D systems, the 2D Navier-Stokes environment ($64 \times 64$ spatial grid with $50$ temporal snapshots) presents a prohibitive computational bottleneck. Instantiating a spatially correlated Gaussian prior requires computing and decomposing a dense covariance matrix of size $N \times N$, where $N$ is the total number of grid points per sample. For Navier-Stokes, $N = 64 \times 64 \times 50 = 204,800$, resulting in a dense covariance matrix with over $4.19 \times 10^{10}$ elements. Even in single precision (FP32), storing this matrix for a single instance would require approximately $167$ GB of VRAM, far exceeding the capacity of standard modern GPUs, thereby restricting our evaluation to the uncorrelated White Noise prior for this dataset.

\begin{table}[h]
\centering
\caption{Architectural and optimization hyperparameters for the 7 pre-training environments.}
\label{tab:hyperparameters}
\begin{tabular}{l c c c c c c}
\toprule
\textbf{PDE System} & \textbf{Prior Type} & \textbf{Layers} & \textbf{Modes} & \textbf{Channels} & \textbf{Batch Size} & \textbf{Steps} \\
\midrule
Heat 1D    & White / Smooth & 4 & 32 & 64 & 256 & 20k \\
RD 1D      & White / Smooth & 4 & 32 & 64 & 256 & 20k \\
Burgers 1D & White / Smooth & 4 & 32 & 64 & 256 & 20k \\
Navier-Stokes 2D & White & 2 & 16 & 32 & 24 & 500k \\
\bottomrule
\end{tabular}
\end{table}

\section{Baseline Descriptions and Implementation Details}
\label{app:baselines}

To rigorously evaluate the efficacy of our proposed continuous dissipative guidance, we benchmarked it against five representative state-of-the-art methodologies for physical constraint enforcement and generative guidance. The theoretical and implementation details of each baseline are outlined below.

\paragraph{1. Physics-Constrained Flow Matching (PCFM):}
PCFM \cite{utkarshphysics} enforces nonlinear physical constraints by applying an explicit algebraic projection at each solver step. The framework relies on a first-order Taylor expansion of the physical constraint functional $h(u)$. Given the unconstrained terminal state prediction $\hat{u}_1$, PCFM finds a corrected state $u$ by solving the linearized system $h(\hat{u}_1) + J(u - \hat{u}_1) = 0$, where $J = \nabla h(\hat{u}_1)$ is the constraint Jacobian. This maps the sample onto the tangent space of the constraint manifold via the Moore-Penrose pseudo-inverse. While theoretically exact for linear equations, this projection requires computing and applying $J^{\dagger}$, leading to the severe $\mathcal{O}(N^2)$ memory bottlenecks and computational overheads discussed in Section \ref{sec:experiments} during our full-PDE stress tests.

\paragraph{2. ProFlow (Proximal Flow Guidance):}
ProFlow~\cite{yu2026ProFlow} is a recent framework (introduced in Jan 2026) specifically designed to address the computational weight and numerical instability of exact projection-based methods like PCFM. Instead of solving a strict linearized system via matrix inversion, it utilizes $K$ unrolled proximal gradient steps on the predicted terminal state $\hat{u}_1$ to softly enforce physics consistency. Most importantly, it then applies a linear interpolation step to project the refined state back onto the probability path, attempting to avoid entirely disrupting the learned prior. \textit{Note:} At the time of writing, the official source code for ProFlow was not publicly available; therefore, we provided an independent and faithful implementation based strictly on the algorithmic description and hyperparameter settings ($K=3$ proximal steps) detailed in the original manuscript.

\paragraph{3. DiffusionPDE:}
DiffusionPDE~\cite{DBLP:conf/nips/HuangYWP24} utilizes the gradient of a physical residual or observation loss to softly guide the generative process. Originally formulated for score-based diffusion models, it relies on obtaining a denoised estimate $\hat{u}_0$ and computing the guidance gradient directly on it. In the context of Flow Matching, this baseline is adapted by computing the gradient of the physical energy functional directly on the deterministic prediction of the terminal state $\hat{u}_1$. Unlike our proposed dissipative guidance, which applies a continuous local force inherently coupled with the intermediate state $u_t$, this method relies entirely on the accuracy of the one-step prediction to anchor the physical correction.

\paragraph{4. Extrapolation, Correction, and Interpolation (ECI):}
ECI~\cite{chenggradient} is a gradient-free, zero-shot framework designed for hard-constrained generation. It iteratively alternates between three stages during flow sampling: extrapolating the intermediate state to a terminal prediction, applying an exact algebraic correction (e.g., orthogonal or oblique projection) to strictly satisfy the constraint, and interpolating the corrected state back to the flow probability path. While highly effective for simple linear or non-overlapping constraints where closed-form projections exist, its reliance on analytical corrections restricts its extensibility to dense, highly non-linear constraints, such as full-domain PDE residuals.

\paragraph{5. D-Flow:}
D-Flow~\cite{DBLP:conf/icml/Ben-HamuPGKSL24} utilizes a flow-based generative prior coupled with an inner-loop optimization routine to satisfy physical constraints. Instead of modifying the intermediate sampling states, it attempts to optimize the initial noise $u_0$ by backpropagating the constraint violation loss through the entire unrolled ODE trajectory.

\section{Inference Implementation Details}
\label{app:implementation_details}

\begin{figure}[h]
    \centering
    \includegraphics[width=0.7\textwidth]{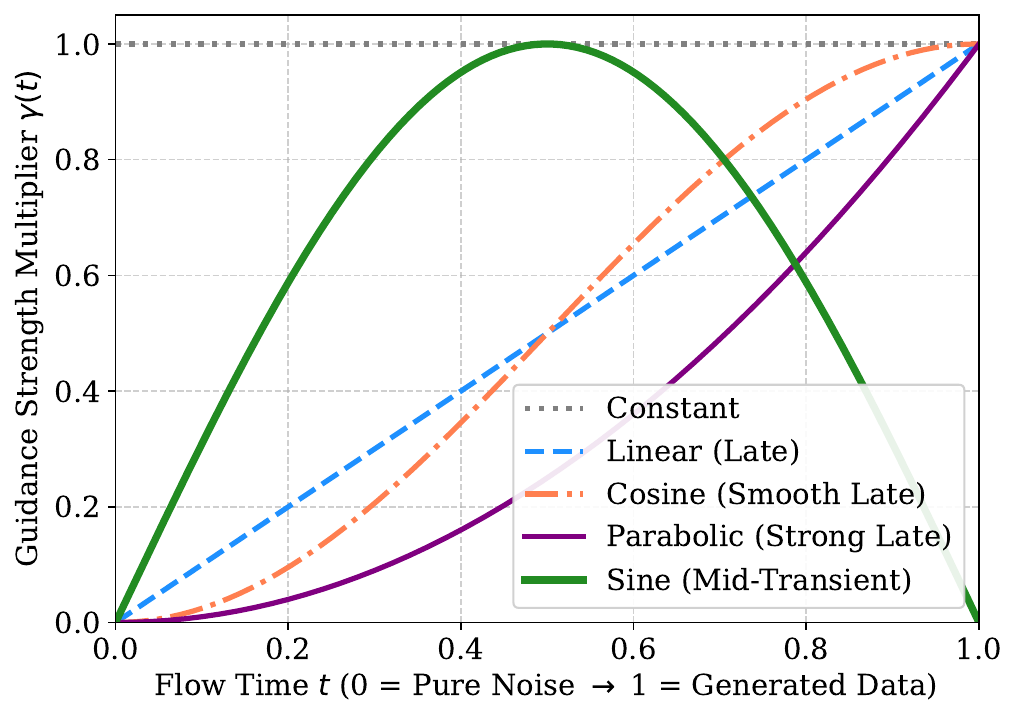}
    \caption{Comparison of guidance scheduling profiles $\gamma(t)$ as a function of the probability flow time $t \in [0, 1]$. While traditional 'Late Guidance' schedules (Linear, Parabolic, Cosine) monotonically increase to apply maximum physical correction at the terminal state, our proposed 'Transient' Sine schedule peaks at $t=0.5$ and decays to zero at $t=1$. This mitigates numerical instabilities caused by evaluating non-linear physical operators on sharp fluid structures during the final stages of the generative process.}
    \label{fig:guidance_schedulers}
\end{figure}

To maintain stability across the highly non-convex PDE landscapes discussed in Section \ref{sec:implementation}, our guided inference algorithm incorporates specific numerical mechanisms and carefully tuned hyperparameters to establish a fair evaluation against the baselines.

\subsection{Numerical Stabilization Mechanisms}

\paragraph{Gradient Clipping} The non-linear operators within the full PDE residual can produce gradients of highly variable magnitude, particularly in the presence of unrefined early-stage structures or sharp physical shocks. To prevent these gradients from dominating the neural vector field $v_\theta$ and destabilizing the ODE solver, we employ $L^2$-norm gradient clipping. For a given clip threshold $c$, the gradient $g$ is scaled as $\tilde{g} = g \cdot \min(1, c/\|g\|_2)$. Unlike global gradient normalization, this strategy bounds the maximum corrective force while preserving the natural decay of the gradient magnitude as the sample converges toward the physical manifold.

\paragraph{Transient Guidance Scheduling} At $t \approx 0$, the state $u_t$ is highly noisy, making the evaluation of spatial physical operators uninformative. Conversely, as $t \to 1$, the formation of sharp physical features can lead to numerical instability in spatial PDE gradients. To address both extremes, we employ a dynamic scheduling function $\gamma(t)$ that implements a transient guidance strategy. In our main experiments, we utilize a sine schedule ($\gamma(t) = \gamma_{max} \sin(\pi t)$, as illustrated in Figure \ref{fig:guidance_schedulers}). This schedule explicitly zeroes out the physics penalty at $t=0$ to allow the unguided neural prior to establish macroscopic structures, applies the strongest physical correction at $t=0.5$, and smoothly decays to zero as $t \to 1$ to prevent gradient explosion and preserve the generative fine-tuning. Empirically, the choice of scheduler significantly influences generative fidelity, particularly for complex PDE dynamics. A comprehensive ablation study comparing different scheduling profiles is provided in Appendix \ref{app:ablations}.

\paragraph{Efficient ODE Integration} While higher-order ODE solvers (e.g., Runge-Kutta) are commonly used in unguided flow matching, they require multiple evaluations of the vector field per time step. In an energy-guided framework, each evaluation necessitates a forward-backward pass through complex PDE operators to compute the implicit VJP ($\nabla_{u} E(u_t)$), making higher-order solvers computationally expensive. By leveraging the inherent smoothness of FFM alongside gradient clipping and transient scheduling, the augmented probability flow is sufficiently stabilized to be integrated using a first-order Euler method. This configuration ensures an $\mathcal{O}(N)$ computational complexity, maintaining efficient inference times compared to projection-based baselines.

\subsection{Inference Hyperparameters and Baseline Tuning}
\label{app:hyperparameters}

To ensure a fair zero-shot evaluation, all sampling techniques are initialized with identical noise realizations. The foundational integration parameters are summarized in Table \ref{tab:common_hyperparams}. The hyperparameters for PCFM, ECI, and D-Flow follow the configurations established by Utkarsh et al.~\cite{utkarshphysics}. ProFlow is configured according to its original manuscript~\cite{yu2026ProFlow}, and DiffusionPDE~\cite{DBLP:conf/nips/HuangYWP24} is tuned specifically to maintain numerical stability across the evaluated PDE systems.

\begin{table}[h]
\centering
\caption{Shared integration hyperparameters across all physical systems. NFE denotes the Number of Function Evaluations (forward integration steps).}
\label{tab:common_hyperparams}
\begin{tabular}{l c c c c}
\toprule
\textbf{Shared Parameter} & \textbf{Heat (1D)} & \textbf{RD (1D)} & \textbf{Burgers (1D)} & \textbf{Navier-Stokes (2D)} \\
\midrule
ODE Solver Scheme  & Forward Euler & Forward Euler & Forward Euler & Forward Euler \\
NFE (Steps)        & 100           & 200           & 200           & 100           \\
Base Flow Prior    & White / Smooth& White / Smooth& White / Smooth& White         \\
\bottomrule
\end{tabular}
\end{table}

\paragraph{Energy guided transport (Ours):} 
The dissipative energy scale $\gamma$ follows a sine schedule. For 1D systems, we set $\gamma_{max} = 75$, while for the 2D Navier-Stokes equations, we reduce the scale to $\gamma_{max} = 25$ to ensure stability. An optional post-hoc refinement phase is applied at $t=1$ to further anchor the sample to the physical manifold. By default, this involves 100 gradient descent steps with a learning rate of $\eta = 0.1$. For the ``Heat Full PDE'' experiment, we use 50 steps with $\eta = 0.01$, and for the ``Heat Partial Guidance'' experiment, we use 200 steps with $\eta = 0.1$. The gradient norm clipping threshold is fixed at $c=1.0$ across all experiments.

\paragraph{PCFM Specifics:} 
\rev{We strictly adopt the official configurations from~\cite{utkarshphysics}. Specifically, guided interpolation ($\lambda = 1.0$, step size $0.01$, $20$ refinement steps) is restricted exclusively to the linear Heat equation, whereas non-linear PDEs employ a single Gauss-Newton step per integration step without guided interpolation. To maximize stability, we retain the official implementation's default Tikhonov regularization parameter ($\epsilon = 10^{-6}$) for inverting the Schur complement and execute projections in \texttt{float64} precision.} As noted independently by~\cite{yu2026ProFlow}, explicit Jacobian-based projections can still exhibit numerical divergence when scaled to dense, non-linear full-PDE constraints.

\paragraph{Optimization-Based Baselines (DiffusionPDE \& D-Flow):} 
These baselines optimize a composite loss function $\mathcal{L}_{\text{total}} = \mathcal{L}_{\text{data}} + \lambda_{\text{phys}} \mathcal{L}_{\text{phys}}$. Following the evaluation protocol in~\cite{utkarshphysics}, we set $\lambda_{\text{phys}} = 0.01$. We observe that these methods are highly sensitive to this hyperparameter: larger values can lead to severe manifold degradation, while smaller values fail to adequately enforce the PDE constraint.
\begin{itemize}
    \item \textbf{DiffusionPDE:} Uses a guidance step size of $\eta = 0.01$. We adopt a more conservative step size than the original formulation to improve numerical stability and prevent divergence across the benchmarked PDEs.
    \item \textbf{D-Flow:} Performs $N_{iter}=20$ L-BFGS inner optimization steps with a learning rate of $\eta = 1.0$.
\end{itemize}

\paragraph{Proximal and Projection Baselines (ECI \& ProFlow):}
\begin{itemize}
    \item \textbf{ECI:} Configured with $n_{mix}=5$ mixing updates per step and a prior resampling frequency of 5 steps, \rev{strictly mirroring the benchmark settings in~\cite{utkarshphysics}.}
    \item \textbf{ProFlow:} Employs $K=3$ proximal steps with a time-dependent learning rate schedule starting from $\eta_0 = 0.01$ \rev{following~\cite{yu2026ProFlow}. When encountering numerical instabilities in our reimplementation, we escalated execution to \texttt{float64} precision and performed a best-effort stabilization search while strictly preserving the original methodology.}
\end{itemize}

\section{Computational Complexity and Efficiency Analysis}
\label{app:complexity}

A fundamental advantage of the proposed Energy-Guided Transport is its computational efficiency and structural scalability to high-dimensional domains. In this section, we formally analyze the time and memory complexity introduced by our physics-informed guidance mechanism, contrasting it with state-of-the-art baselines.

Let $N$ denote the total number of spatial degrees of freedom (e.g., $N = H \times W$ for a 2D physical field). 

\subsection{Theoretical Complexity: VJP vs. Explicit Jacobian}

The primary computational bottleneck in physics-guided generation is the interaction with the constraint operator $h(u)$. 

\paragraph{Memory Complexity: $\mathcal{O}(N)$} To compute a hard projection, methods like PCFM must explicitly materialize the Jacobian matrix $J = \nabla h(u) \in \mathbb{R}^{M \times N}$. For a full-PDE constraint on a 2D field of size $64 \times 64$, $N=4096$, resulting in a dense Jacobian with over 16 million entries per sample. Storing and inverting this matrix imposes an intractable $\mathcal{O}(N^2)$ memory footprint, severely limiting its application in 3D systems or high-resolution 2D fluids. Conversely, our framework leverages reverse-mode automatic differentiation to compute implicit Vector-Jacobian Products (VJPs). By directly computing the gradient of the scalar energy ($\nabla_u E(u_t)$), the memory requirement is strictly bounded by the storage of the intermediate forward activation tensors. Consequently, the spatial complexity of our physical guidance remains $\mathcal{O}(N)$, matching the theoretical minimum required to represent the state itself.

\paragraph{Time Complexity: $\mathcal{O}(N)$} The computational overhead per ODE integration step in our method consists of evaluating the analytical PDE residual $h(u_t)$ (forward pass) and computing the VJP (backward pass). Assuming the differential operators are discretized via standard local finite differences, the forward evaluation scales linearly, $\mathcal{O}(N)$. According to the Baur-Strassen theorem of algorithmic differentiation, the time complexity of the reverse-mode backward pass is mathematically guaranteed to be bounded by a small constant factor $c \le 5$ of the forward pass. Therefore, the total time complexity for the physical guidance step is strictly $\mathcal{O}(N)$. This linear scaling ensures that the physical guidance does not become the computational bottleneck: while the base generative backbone (e.g., Fourier Neural Operator) operates at $\mathcal{O}(N \log N)$ due to Fast Fourier Transforms, our dissipative guidance preserves the overall asymptotic complexity of the framework.

\subsection{Empirical Computational Budget (NFE and Operator Evaluations)}

In flow-based and diffusion-based generative models, the practical computational cost is typically measured by the Number of Function Evaluations (NFE) of the neural vector field $v_\theta$, combined with the number of physical operator evaluations.

\paragraph{Baseline Bottlenecks} Optimization and projection-based baselines inherently multiply the computational budget to satisfy constraints. For instance, D-Flow performs global optimization over the initial noise, requiring a forward and an adjoint-backward pass through the entire ODE for each of its 20 L-BFGS iterations. For a standard 100-step solver, this demands up to 4,000 NFE per generated sample. ECI multiplies its integration cost by repeating its mixing step $n_{mix}=5$ times per ODE step. PCFM and ProFlow compute computationally demanding proximal or projection updates at every intermediate time step (e.g., ProFlow computes $K=3$ proximal PDE steps per integration step, while PCFM evaluates explicit Jacobians alongside 20 refinement iterations per step).

\paragraph{Energy-Guided Transport} 
Unlike the aforementioned baselines, our standard guided inference requires exactly 1 NFE and 1 VJP evaluation per integration step. The continuous dynamic provided by the transient scheduler smoothly steers the generation without requiring expensive inner-loop optimizations during the flow. For a 100-step Euler integration, this strictly bounds the budget to 100 NFE and 100 PDE gradient evaluations.

\paragraph{Hybrid Energy-Guided Transport} 
When maximum physical adherence is required, our algorithm optionally incorporates a terminal refinement phase at $t=1$. Importantly, this refinement minimizes the physical energy strictly via gradient descent on the final tensor without querying the neural network. Therefore, performing $R$ refinement steps (e.g., $R=50$ or $R=200$) adds 0 NFE to the overall budget. It incurs only $R$ lightweight $\mathcal{O}(N)$ evaluations of the explicit physical operator. This design strategically uncouples the heavy generative modeling from the physical fine-tuning, resulting in rigorous physical consistency, without compromising the learned data manifold, achieved at a fraction of the computational cost of competing methodologies.

\section{The Ground Truth Floor, Stencil Overfitting, and the Waterbed Effect}
\label{app:waterbed}

To rigorously contextualize the physical residuals achieved by the generative models, we explicitly compute the evaluation metrics on the testing split of the Ground Truth (GT) simulated dataset. As shown in Table \ref{tab:gt_residuals}, the GT errors for Initial Conditions, Boundary Conditions, and Global Mass evaluate to exactly zero (up to machine precision limits). This is expected, as these macroscopic constraints are explicitly hard-coded or globally conserved during the numerical simulation process. 

Conversely, the interior PDE residual and, notably, the Global Mass error in non-linear systems (such as Burgers and Reaction-Diffusion) evaluate to strictly positive values ($> 0$). This occurs because simulating continuous physical laws on a discrete grid inherently introduces numerical truncation errors and temporal drift (e.g., $\mathcal{O}(\Delta x^2)$ for standard spatial finite difference stencils). These GT residuals serve as the theoretical numerical precision limit, or the ``floor,'' for our evaluation framework.

\paragraph{Overfitting the Evaluation Stencil (Discrete Floor Violations)} 
An interesting phenomenon arises when analyzing optimization-heavy baselines: they occasionally report global scaled PDE or mass residuals that are strictly lower than the GT floor. As observed in the broader literature on Physics-Informed Neural Networks~\cite{krishnapriyan2021characterizing, wang2021understanding}, breaching the GT floor does not necessarily indicate the discovery of a structurally superior physical solution. Rather, it is often a manifestation of stencil overfitting. 

When a generative baseline heavily relies on a specific discrete evaluation operator to guide its generation (e.g., via hard algebraic projections or continuous gradient penalties), the model treats the evaluation stencil itself as the absolute optimization objective. Most importantly, because the base generative prior is trained exclusively on the GT dataset, any sample whose discrete residual is forced significantly below the GT floor is, by definition, pushed outside the support of the training distribution. The model is essentially generating out-of-distribution samples that it has never seen, finding a local minimum where localized, high-frequency perturbations zero out the discrete arithmetic of the stencil while drifting from the true continuous physics.

\paragraph{The Waterbed Effect}
We term this topological trade-off the Waterbed Effect: artificially suppressing the scaled PDE error below the theoretical floor inevitably causes a complementary degradation in unconstrained metrics. While the generated samples generally maintain their macroscopic coherence, this subtle overfitting is empirically validated by two corresponding statistical signals in our evaluation:
\begin{enumerate}
    \item \textbf{An increase in Raw PDE Error:} The unscaled spatial derivatives become abnormally large compared to the vanilla unguided model, revealing the presence of subtle high-frequency artifacts injected across the generated field.
    \item \textbf{A degradation of Sample MSE:} Because the generated sample is pulled artificially off the true underlying data manifold to satisfy the discrete stencil, it results in a measurable worsening of the core distributional metrics.
\end{enumerate}

Our EGT framework mitigates this effect structurally. By guiding the generative process smoothly and deferring explicit physics optimizations, the framework steers the generation toward the GT floor without over-constraining the features to the specific discrete stencil. This approach preserves the continuous physical manifold, successfully minimizing the physical residual without artificially inflating the complementary distributional metrics.

\begin{table}[h]
\centering
\caption{Intrinsic physical residuals of the Ground Truth simulated datasets. These values represent the numerical precision limit (lower bound) dictated by grid discretization. To maintain strict consistency with our main evaluations, we report the global mean absolute error (MAE) computed using both the Partial and Full generative configurations. (\textit{Note: The Heat equation dataset is evaluated exclusively on Initial Conditions, hence the absence of BC error}).}
\label{tab:gt_residuals}
\resizebox{\textwidth}{!}{%
\begin{tabular}{l c c c c c c c}
\toprule
\textbf{Dataset} & \textbf{IC Error} & \textbf{BC Error} & \textbf{Mass Error} & \textbf{PDE Error (Scaled)} & \textbf{PDE Error (Raw)} & \textbf{Partial Phys (MAE)} & \textbf{Full Phys (MAE)} \\
\midrule
Heat Equation    & 0.00e0 & -        & 1.32e-7 & 3.57e-5 & 9.04e-3 & 5.44e-8 & 3.53e-5 \\
Reaction-Diff.   & 0.00e0 & 0.00e0   & 1.90e-3 & 3.07e-7 & 5.03e-3 & 8.30e-4 & 3.04e-7 \\
Burgers' Eq.     & 0.00e0 & 5.99e-9  & 4.70e-3 & 6.24e-5 & 6.24e-1 & 1.56e-3 & 6.05e-5 \\
Navier-Stokes    & 0.00e0 & 0.00e0   & 1.43e-7 & 6.98e-6 & 2.86e-2 & 1.66e-9 & 6.63e-6 \\
\bottomrule
\end{tabular}%
}
\end{table}

\section{Extended Results and Disaggregated Metrics}
\label{app:extended_results}

In this section, we provide the exhaustive evaluation metrics for all tested generative methods across our suite of PDE benchmarks. To facilitate a clear and comprehensive analysis, we consolidate the results into dedicated subsections for each physical system (Heat, Reaction-Diffusion, Burgers', and Navier-Stokes). 

Each physical system is evaluated under two distinct observation regimes:
\begin{itemize}
    \item \textbf{Partial Physics Guidance (Panel A):} The generative models are constrained solely by partial observations, typically Initial Conditions (IC), Boundary Conditions (BC), and/or global integral quantities.
    \item \textbf{Full PDE Guidance (Panel B):} The generative models are subjected to a rigorous ``stress test", where they are constrained by the full underlying differential equation across the entire spatiotemporal domain, alongside the boundary constraints.
\end{itemize}
Furthermore, we disaggregate the results based on the underlying generative prior: the standard Gaussian White Noise prior and the structured Smooth Gaussian Kernel prior.

\subsection{Evaluation Metrics Formulation}
\label{app:metrics_formulation}

To rigorously assess the performance of the generative models, we track a diverse set of metrics encompassing computational stability, distributional fidelity, and physical consistency.

\paragraph{Computational Stability and Efficiency:}
\begin{itemize}
    \item \textbf{Success Rate (SR, \%):} Represents the percentage of samples successfully generated without encountering numerical divergence (e.g., exploding gradients resulting in \texttt{NaN} values) or Out-Of-Memory (OOM) errors. Trajectories that encountered numerical divergence or memory exhaustion (NaN/OOM) are denoted as \textit{DNC} (Did Not Complete).
    \item \textbf{Inference Speed (Spd, sec/sample):} The average computational wall-clock time required to generate a single physical sample, amortized over the total batch size. \textit{Hardware Disclaimer:} To ensure a strictly fair and unbiased comparison, all competing generative methods evaluated within a specific PDE benchmark were executed on the identical hardware (e.g., NVIDIA RTX 8000, A6000, L40 GPU). Consequently, while absolute inference times may vary across different physical systems due to hardware assignments and grid dimensionalities, the relative speed comparisons within each individual experiment are strictly standardized.
\end{itemize}

\paragraph{Distributional and Observational Fidelity:}
Following established practices in generative modeling for PDEs \cite{utkarshphysics, yu2026ProFlow}, we evaluate the capability of the models to capture the true underlying data distribution. Let $U_{pred} \in \mathbb{R}^{B \times N_x \times N_t}$ and $U_{true} \in \mathbb{R}^{B \times N_x \times N_t}$ represent the batched generated and ground truth tensors.
\begin{itemize}
    \item \textbf{Mean MSE (MMSE):} Measures the mean squared error between the sample mean of the generated batch and the ground truth batch: $\text{MSE}(\mu_{pred}, \mu_{true})$. It indicates the accuracy of the estimated posterior mean.
    \item \textbf{Standard Deviation MSE (SMSE):} Measures the mean squared error between the sample standard deviations: $\text{MSE}(\sigma_{pred}, \sigma_{true})$. It quantifies how well the method captures the predictive uncertainty.
    \item \textbf{Sample-wise MSE (Sample MSE):} The standard point-wise mean squared error evaluated directly between the predicted and true spatiotemporal fields.
\end{itemize}

\paragraph{Physical Consistency (Constraint Errors):}
To evaluate adherence to the governing physical laws, we compute absolute residuals for various constraints. All metrics are reported as Mean Absolute Errors (MAE) averaged over the generated samples.
\begin{itemize}
    \item \textbf{IC Error \& BC Error:} Measure the absolute deviation of the generated solution from the prescribed Initial Conditions (at $t=0$) and boundary values (e.g., Dirichlet or Neumann fluxes).
    \item \textbf{Mass Error \& Conservation Laws (CL):} Quantify the violation of global integral invariants. \textbf{Mass} explicitly refers to the conservation of mass over time (e.g., balancing total state sum with boundary fluxes and reaction sources), while \textbf{CL} refers to additional physical invariants or local multi-step flux balances (such as Godunov flux constraints in the Burgers' equation).
    \item \textbf{Raw PDE Error:} The unscaled absolute error of the interior partial differential equation evaluated via discrete finite differences. As discussed in Appendix \ref{app:waterbed}, observing this metric is key to detect high-frequency artifacts and ``stencil overfitting".
    \item \textbf{Total Phys (MAE):} This is the global Mean Absolute Error of the specific composite guidance function, $\mathcal{H}(u)$, employed during inference. It is important to notice that the composition of this metric depends on the evaluation regime:
    \begin{itemize}
        \item In the \textbf{Partial Guidance} task, the MAE is computed exclusively over the boundary constraints (IC, BC) and the integral mass/CL residuals.
        \item In the \textbf{Full PDE Guidance} task, the MAE concatenates the boundary constraints with the internal PDE residual. In this full configuration, the PDE residual is strictly scaled by the spatial grid discretization factor (e.g., $\Delta x^2$). This matches the exact $\mathcal{H}(u)$ loss formulation structurally required by optimization-based baselines to maintain numerical stability during gradient-based guidance.
    \end{itemize}
\end{itemize}

\subsection{1D Heat Equation Results}
\label{app:results_heat}

Table \ref{tab:results_heat_master} presents the comprehensive evaluation for the 1D Heat Equation. As the simplest linear PDE in our benchmark, this environment serves as a foundational test to measure the models' capability to enforce basic spatial and temporal constraints before moving to non-linear dynamics.

\begin{table}[htbp]
\centering
\caption{Comprehensive evaluation metrics for the \textbf{1D Heat Equation}. The table compares zero-shot guidance performance across Partial (Boundary/IC) and Full-PDE tasks using both White Noise and Smooth Gaussian priors. $\downarrow$ indicates lower is better, $\uparrow$ indicates higher is better. All physical metrics are averaged over the test set. Instances with a success rate lower than $100\%$ are marked as DNC. Best results are in \textbf{bold}, second-best are \underline{underlined} (excluding Vanilla).}
\label{tab:results_heat_master}
\resizebox{\textwidth}{!}{%
\begin{tabular}{l | c | c c c c c c c }
\toprule
\textbf{Metric} & \textbf{Vanilla} & \textbf{EGT (Ours)} & \textbf{Hybrid EGT (Ours)} & \textbf{PCFM} & \textbf{ProFlow} & \textbf{DiffPDE} & \textbf{ECI} & \textbf{D-Flow} \\
\midrule
\multicolumn{9}{l}{\textbf{PART 1: PARTIAL/BOUNDARY GUIDANCE}} \\
\midrule
\multicolumn{9}{l}{\textit{\textbf{Prior: White Noise}}} \\
\midrule
Success Rate ($\uparrow$)        & 100\% & \textbf{100\%} & \textbf{100\%} & 0.2\% & \textbf{100\%} & \textbf{100\%} & \textbf{100\%} & \textbf{100\%} \\
Raw PDE Residual ($\downarrow$)  & 0.354 & \underline{0.399} & 0.483 & DNC & 0.402 & \textbf{0.318} & 0.473 & 5.676 \\
Physical Res. MAE ($\downarrow$) & 0.256 & 0.051 & \underline{0.019} & DNC & 0.189 & 0.026 & \textbf{0.013} & 0.085 \\
IC Error ($\downarrow$)          & 0.433 & 0.015 & \underline{3.1e-5} & DNC& 0.270 & 4.3e-3 & \textbf{2.2e-20} & 0.011 \\
Mass Error ($\downarrow$)        & 1.8e-3 & 3.8e-3 & 0.033 & DNC & \textbf{1.6e-3} & \underline{3.3e-3} & 0.027 & 0.152 \\
\midrule
MMSE ($\downarrow$)              & 5.2e-3 & 1.1e-3 & \textbf{9.3e-4} & DNC & 4.9e-3 & \underline{9.4e-4} & \textbf{3.3e-4} & 8.6e-3 \\
SMSE ($\downarrow$)              & 1.5e-2 & 6.1e-3 & 6.7e-3 & DNC & \underline{5.0e-3} & 6.4e-3 & \textbf{5.3e-5} & 0.177 \\
Sample MSE ($\downarrow$)         & 0.083 & 0.015 & 0.017 & DNC & 0.062 & \textbf{0.013} & \underline{0.014} & 0.348 \\
\midrule
Speed (s/sample) ($\downarrow$)  & 0.19  & \textbf{0.37} & \underline{0.46} & DNC & 8.12 & 0.94 & 8.30 & 15.58 \\
\midrule
\multicolumn{9}{l}{\textit{\textbf{Prior: Smooth Kernel}}} \\
\midrule
Success Rate ($\uparrow$)        & 100\% & \textbf{100\%} & \textbf{100\%} & 1.1\% & \textbf{100\%} & \textbf{100\%} & 0\% & \textbf{100\%} \\
Raw PDE Residual ($\downarrow$)  & 40.83 & 40.68 & 40.30 & DNC & \textbf{3.728} & 39.23 & DNC & \underline{1.3e3} \\
Physical Res. MAE ($\downarrow$) & 0.425 & \underline{0.047} & \textbf{0.018} & DNC & 0.179 & 0.062 & DNC & 0.973 \\
IC Error ($\downarrow$)          & 0.984 & \underline{0.010} & \textbf{2.7e-5} & DNC & 0.226 & \underline{0.010} & DNC & 0.092 \\
Mass Error ($\downarrow$)        & 0.053 & \textbf{0.014} & 0.032 & DNC & \underline{0.030} & 0.046 & DNC & 1.817 \\
\midrule
MMSE ($\downarrow$)              & 1.3e-2 & 5.0e-4 & \underline{3.7e-4} & DNC & 1.2e-2 & \textbf{2.6e-4} & DNC & 1.361 \\
SMSE ($\downarrow$)              & 3.4e-2 & 1.0e-2 & 1.0e-2 & DNC & \underline{7.4e-3} & \textbf{6.2e-3} & DNC & 21.66 \\
Sample MSE ($\downarrow$)         & 0.256 & \underline{0.049} & \underline{0.049} & DNC & 0.052 & \textbf{0.043} & DNC & 24.88 \\
\midrule
Speed (s/sample) ($\downarrow$)  & 0.19  & \textbf{0.37} & \underline{0.43} & DNC & 8.16 & 0.94 & DNC & 15.63 \\
\midrule
\midrule
\multicolumn{9}{l}{\textbf{PART 2: FULL-PDE GUIDANCE}} \\
\midrule
\multicolumn{9}{l}{\textit{\textbf{Prior: White Noise}}} \\
\midrule
Success Rate ($\uparrow$)        & 100\% & \textbf{100\%} & \textbf{100\%} & 0\% & 0.4\% & \textbf{100\%} & \textbf{100\%} & \textbf{100\%} \\
Raw PDE Residual ($\downarrow$)  & 0.345 & 0.530 & \underline{0.305} & DNC & DNC & \textbf{0.025} & 0.472 & 7.875 \\
Physical Res. MAE ($\downarrow$) & 6.2e-3 & 5.1e-3 & \underline{3.9e-3} & DNC & DNC & 0.985 & \textbf{1.8e-3} & 0.032 \\
IC Error ($\downarrow$)          & 0.410 & 0.174 & \underline{0.156} & DNC & DNC & 18.05 & \textbf{2.0e-20} & 28.12 \\
Mass Error ($\downarrow$)        & 1.8e-3 & \underline{0.017} & \textbf{0.012} & DNC & DNC & 0.671 & 0.026 & 0.571 \\
\midrule
MMSE ($\downarrow$)              & 4.9e-3 & \underline{4.0e-3} & \underline{4.0e-3} & DNC & DNC & 0.137 & \textbf{1.0e-4} & 0.029 \\
SMSE ($\downarrow$)              & 1.4e-2 & \underline{0.015} & 0.016 & DNC & DNC & 16.81 & \textbf{3.8e-5} & 22.11 \\
Sample MSE ($\downarrow$)         & 0.078 & \underline{0.046} & 0.049 & DNC & DNC & 18.14 & \textbf{0.014} & 23.66 \\
\midrule
Speed (s/sample) ($\downarrow$)  & 0.13  & \textbf{0.23} & \underline{0.27} & DNC & DNC & 0.68  & 10.28 & 18.06 \\
\midrule
\multicolumn{9}{l}{\textit{\textbf{Prior: Smooth Kernel}}} \\
\midrule
Success Rate ($\uparrow$)        & 100\% & \textbf{100\%} & \textbf{100\%} & 0\% & 0\% & \textbf{100\%} & 0\% & \textbf{100\%} \\
Raw PDE Residual ($\downarrow$)  & 40.48 & \underline{1.529} & \textbf{0.914} & DNC & DNC & 20.16 & DNC & 1.6e2 \\
Physical Res. MAE ($\downarrow$) & 0.165 & \underline{7.1e-3} & \textbf{4.5e-3} & DNC & DNC & 0.079 & DNC & 0.657 \\
IC Error ($\downarrow$)          & 0.941 & 0.027 & \underline{0.018} & DNC & DNC & \textbf{9.8e-3} & DNC & 0.117 \\
Mass Error ($\downarrow$)        & 0.052 & \textbf{0.066} & 0.044 & DNC & DNC & \underline{0.024} & DNC & 3.298 \\
\midrule
MMSE ($\downarrow$)              & 0.013 & \underline{1.2e-3} & \underline{1.2e-3} & DNC & DNC & \textbf{3.1e-4} & DNC & 0.620 \\
SMSE ($\downarrow$)              & 0.034 & \underline{2.3e-3} & \textbf{2.1e-3} & DNC & DNC & 6.6e-3 & DNC & 15.47 \\
Sample MSE ($\downarrow$)         & 0.248 & \textbf{0.035} & \textbf{0.035} & DNC & DNC & \underline{0.045} & DNC & 17.61 \\
\midrule
Speed (s/sample) ($\downarrow$)  & 0.19  & \textbf{0.51} & \underline{0.55} & DNC & DNC & 1.04  & DNC & 16.79 \\
\bottomrule
\end{tabular}%
}
\end{table}

\subsubsection{Discussion}

The empirical results for the 1D Heat Equation highlight the distinct operational characteristics of the evaluated methods, particularly regarding their sensitivity to the underlying generative prior and the complexity of the imposed constraints.

\paragraph{Performance under White Noise Prior} 
Given the linear nature of the Heat Equation, the uncorrelated White Noise regime favors baseline methods equipped with exact algebraic correction mechanisms. Most notably, the ECI framework achieves state-of-the-art performance across almost all metrics in both Partial and Full-PDE tasks. It successfully enforces the Initial Condition to near machine precision ($2.2 \times 10^{-20}$) while maintaining excellent distributional fidelity (Sample MSE 0.014), albeit at a higher computational cost ($\sim 10$ seconds per sample). DiffusionPDE demonstrates strong performance in the Partial task (Sample MSE 0.013), but its distributional fidelity degrades significantly under Full-PDE guidance, where the Sample MSE rises to 18.14 despite maintaining numerical stability. ProFlow operates reliably under Partial guidance (100\% SR) but struggles to complete the Full-PDE stress test (0.4\% SR). Finally, while D-Flow maintains a 100\% success rate, it incurs significant computational overhead ($\sim 18$s) and struggles to preserve the underlying data distribution, yielding elevated Sample MSE and physical errors. PCFM encounters severe numerical instabilities in this environment, resulting in Did Not Complete (DNC) in all Heat experiments.

\paragraph{Prior Shift to Smooth Gaussian Kernels}
The operational landscape shifts considerably when transitioning to the spatially correlated Smooth Kernel prior. Methods reliant on hard algebraic projections struggle to adapt to this structured kernel: ECI drops to a 0\% success rate, being unable to interpolate smoothly without diverging, and PCFM remains predominantly DNC. ProFlow continues to stably solve the Partial task but again fails to scale to the Full-PDE formulation (0\% SR). While DiffusionPDE safely completes all trajectories (100\% SR), its physical correction capability is limited in the Partial task (yielding a Raw PDE residual of 39.23, comparable to the unguided Vanilla model) and introduces distributional drift in the Full-PDE task.

\paragraph{Robustness and Efficiency of the Proposed Method}
The proposed EGT framework and its Hybrid variant demonstrate consistent robustness, maintaining a 100\% success rate across all generative priors and constraint complexities. While EGT slightly trails ECI's exact numerical matching in the simpler White Noise Partial task, it exhibits superior adaptability in the structured prior regime. Under the challenging Smooth Kernel Full-PDE task, Hybrid EGT effectively balances physical adherence and manifold preservation: it reduces the Raw PDE residual to $0.914$ (down from the Vanilla baseline of 40.48) while maintaining a highly accurate Sample MSE of $0.035$. Most importantly, this consistency is achieved with minimal computational overhead. Hybrid EGT requires only $0.55$ seconds per sample, making it highly efficient compared to optimization-heavy baselines like D-Flow ($\sim 16.7$s) or ECI's White Noise performance ($\sim 10.2$s).

\subsection{1D Burgers' Equation Results}
\label{app:results_burgers}

Table \ref{tab:results_burgers_master} presents the comprehensive evaluation for the 1D Burgers' Equation. As a canonical non-linear PDE that develops steep gradients and shock waves over time, this dataset introduces significant complexity, stress-testing the models' ability to transport the generative flow accurately while avoiding numerical divergence in highly stiff regions.

\begin{table}[htbp]
\centering
\caption{Comprehensive evaluation metrics for the \textbf{1D Burgers' Equation}. The table compares zero-shot guidance performance across Partial (Boundary/IC) and Full-PDE tasks using both White Noise and Smooth Gaussian priors. $\downarrow$ indicates lower is better, $\uparrow$ indicates higher is better. All physical metrics are averaged over the test set. Instances with a success rate lower than $100\%$ are marked as DNC. $^\dagger$ indicates values that artificially fall below the discrete ground-truth stencil precision limit (stencil overfitting/waterbed effect).}
\label{tab:results_burgers_master}
\resizebox{\textwidth}{!}{%
\begin{tabular}{l | c | c c c c c c c }
\toprule
\textbf{Metric} & \textbf{Vanilla} & \textbf{EGT (Ours)} & \textbf{Hybrid EGT (Ours)} & \textbf{PCFM} & \textbf{ProFlow} & \textbf{DiffPDE} & \textbf{ECI} & \textbf{D-Flow} \\
\midrule
\multicolumn{9}{l}{\textbf{PART 1: PARTIAL/BOUNDARY GUIDANCE}} \\
\midrule
\multicolumn{9}{l}{\textit{\textbf{Prior: White Noise}}} \\
\midrule
Success Rate ($\uparrow$)        & 100\% & \textbf{100\%} & \textbf{100\%} & \textbf{100\%} & \textbf{100\%} & \textbf{100\%} & \textbf{100\%} & \textbf{100\%} \\
Raw PDE Residual ($\downarrow$)  & 0.955 & 1.247 & 1.441 & 1.266 & \textbf{0.949} & \underline{1.128} & 1.205 & 4.370 \\
Physical Res. MAE ($\downarrow$) & 0.102 & 0.051 & \underline{0.013} & \textbf{1.2e-8}$^\dagger$ & 0.134 & 0.059 & 0.049 & 0.083 \\
IC Error ($\downarrow$)          & 0.175 & 0.167 & 0.166 & 0.187 & 0.184 & \underline{8.1e-3} & \textbf{6.0e-20} & 0.174 \\
BC Error ($\downarrow$)          & 0.064 & 7.0e-3 & \underline{2.5e-5} & \textbf{5.5e-16}$^\dagger$ & 0.098 & 6.2e-4 & 1.2e-4 & 0.160 \\
Mass Error ($\downarrow$)        & 4.9e-3 & 0.030 & 0.032 & \textbf{3.6e-8}$^\dagger$ & \underline{2.2e-3}$^\dagger$ & 0.141 & 0.138 & 0.083 \\
CL Error ($\downarrow$)          & 0.077 & 0.107 & \underline{0.083} & \textbf{0.066} & 0.172 & 0.132 & 0.140 & 0.090 \\
\midrule
MMSE ($\downarrow$)              & 0.021 & 0.039 & 0.038 & 0.028 & 0.075 & \underline{0.017} & \textbf{4.2e-3} & 0.023 \\
SMSE ($\downarrow$)              & 5.1e-3 & 5.7e-3 & 5.8e-3 & 6.2e-3 & 8.0e-3 & \underline{4.9e-3} & \textbf{1.5e-3} & 0.104 \\
Sample MSE ($\downarrow$)         & 0.205 & 0.206 & 0.205 & 0.193 & 0.235 & \underline{0.155} & \textbf{0.110} & 0.309 \\
\midrule
Speed (s/sample) ($\downarrow$)  & 0.41 & \textbf{0.48} & \underline{0.52} & 1.29 & 21.34 & 1.17 & 21.87 & 39.61 \\
\midrule
\multicolumn{9}{l}{\textit{\textbf{Prior: Smooth Kernel}}} \\
\midrule
Success Rate ($\uparrow$)        & 100\% & \textbf{100\%} & \textbf{100\%} & \textbf{100\%} & \textbf{100\%} & \textbf{100\%} & \textbf{100\%} & \textbf{100\%} \\
Raw PDE Residual ($\downarrow$)  & 1.543 & 1.684 & 1.722 & 1.856 & \textbf{1.052} & 1.733 & \underline{1.512} & 3.6e2 \\
Physical Res. MAE ($\downarrow$) & 0.127 & 0.021 & \underline{4.6e-3} & \textbf{1.2e-8}$^\dagger$ & 0.084 & 0.053 & 0.068 & 0.372 \\
IC Error ($\downarrow$)          & 0.130 & 0.131 & 0.131 & 0.206 & 0.156 & \underline{0.019} & \textbf{2.5e-18}$^\dagger$ & 14.673 \\
BC Error ($\downarrow$)          & 0.080 & 1.3e-3 & \underline{4.6e-6} & \textbf{5.5e-16}$^\dagger$ & 0.042 & 1.4e-3 & 2.9e-4 & 8.939 \\
Mass Error ($\downarrow$)        & 0.021 & 0.014 & 0.012 & \textbf{3.7e-8}$^\dagger$ & \underline{6.8e-3} & 0.106 & 0.188 & 0.532 \\
CL Error ($\downarrow$)          & 0.065 & 0.077 & 0.068 & \underline{0.066} & \textbf{0.048} & 0.111 & 0.205 & 0.109 \\
\midrule
MMSE ($\downarrow$)              & 4.4e-3 & 6.0e-3 & \textbf{5.9e-3} & 0.026 & 0.074 & \textbf{3.8e-3} & 0.057 & 0.037 \\
SMSE ($\downarrow$)              & 4.7e-3 & \textbf{4.9e-3} & \textbf{4.9e-3} & \underline{0.015} & 0.065 & \textbf{4.9e-3} & 0.060 & 9.320 \\
Sample MSE ($\downarrow$)         & 0.16 & \underline{0.137} & \underline{0.137} & 0.223 & 0.164 & \textbf{0.126} & 0.142 & 10.584 \\
\midrule
Speed (s/sample) ($\downarrow$)  & 0.41 & \underline{0.48} & 0.52 & 1.28 & 21.41 & 1.16 & 21.94 & 39.29 \\
\midrule
\midrule
\multicolumn{9}{l}{\textbf{PART 2: FULL-PDE GUIDANCE}} \\
\midrule
\multicolumn{9}{l}{\textit{\textbf{Prior: White Noise}}} \\
\midrule
Success Rate ($\uparrow$)        & 100\% & \textbf{100\%} & \textbf{100\%} & \textbf{100\%} & \textbf{100\%} & \textbf{100\%} & \textbf{100\%} & \textbf{100\%} \\
Raw PDE Residual ($\downarrow$)  & 0.955 & 1.328 & 1.529 & 4.9e+18 & \textbf{0.947} & \underline{1.131} & 1.221 & 5.116 \\
Physical Res. MAE ($\downarrow$) & 5.1e-3 & 1.9e-3 & \textbf{1.5e-4} & 4.8e+14 & 6.3e-3 & 9.5e-4 & \underline{2.0e-4} & 2.9e-3 \\
IC Error ($\downarrow$)          & 0.175 & 0.011 & \underline{1.5e-9}$^\dagger$ & 1.1e+15 & 0.184 & 7.6e-3 & \textbf{6.0e-20}$^\dagger$ & 0.170 \\
BC Error ($\downarrow$)          & 0.062 & 6.6e-3 & \textbf{1.8e-9}$^\dagger$ & 4.4e+5 & 0.101 & 6.2e-4 & \underline{8.7e-5} & 0.095 \\
Mass Error ($\downarrow$)        & 4.9e-3 & 0.147 & 0.196 & 2.9e+14 & \textbf{3.2e-3} & 0.142 & 0.143 & \underline{0.108} \\
CL Error ($\downarrow$)          & 0.074 & 0.156 & 0.200 & 3.0e+5 & 0.174 & \underline{0.134} & 0.140 & \textbf{0.096} \\
\midrule
MMSE ($\downarrow$)              & 0.021 & 0.033 & 0.033 & 1.3e+14 & 0.073 & \underline{0.018} & \textbf{5.7e-3} & 0.022 \\
SMSE ($\downarrow$)              & 4.7e-3 & 4.8e-3 & 5.2e-3 & 1.1e+17 & 6.8e-3 & \underline{4.5e-3} & \textbf{1.7e-3} & 0.140 \\
Sample MSE ($\downarrow$)         & 0.204 & 0.189 & 0.189 & 1.1e+17 & 0.239 & \underline{0.153} & \textbf{0.116} & 0.346 \\
\midrule
Speed (s/sample) ($\downarrow$)  & 0.26 & \underline{0.45} & 0.54 & 67.11 & 18.68 & 1.14 & 18.67 & 32.69 \\
\midrule
\multicolumn{9}{l}{\textit{\textbf{Prior: Smooth Kernel}}} \\
\midrule
Success Rate ($\uparrow$)        & 100\% & \textbf{100\%} & \textbf{100\%} & \textbf{100\%} & \textbf{100\%} & \textbf{100\%} & \textbf{100\%} & \textbf{100\%} \\
Raw PDE Residual ($\downarrow$)  & 1.538 & 1.819 & 1.900 & 5.9e+17 & \textbf{1.053} & 1.727 & \underline{1.513} & 4.0e2 \\
Physical Res. MAE ($\downarrow$) & 5.8e-3 & 1.5e-3 & \textbf{1.8e-4} & 5.7e+13 & 4.5e-3 & 1.5e-3 & \underline{3.1e-4} & 0.048 \\
IC Error ($\downarrow$)          & 0.127 & 0.018 & \underline{1.6e-9}$^\dagger$ & 3.3e+15 & 0.153 & 0.018 & \textbf{2.5e-18}$^\dagger$ & 16.369 \\
BC Error ($\downarrow$)          & 0.082 & 1.3e-3 & \textbf{3.4e-10}$^\dagger$ & 9.1e+4 & 0.043 & 1.4e-3 & \underline{2.9e-4} & 16.783 \\
Mass Error ($\downarrow$)        & 0.022 & 0.108 & 0.174 & 8.3e+14 & \textbf{7.2e-3} & \underline{0.107} & 0.188 & 0.369 \\
CL Error ($\downarrow$)          & 0.065 & 0.125 & 0.184 & 2.7e+5 & \textbf{0.048} & \underline{0.112} & 0.205 & 0.129 \\
\midrule
MMSE ($\downarrow$)              & 4.0e-3 & \underline{4.9e-3} & \underline{4.9e-3} & 2.8e+13 & 0.074 & \textbf{3.5e-3} & 0.057 & 0.052 \\
SMSE ($\downarrow$)              & 4.8e-3 & 5.2e-3 & \underline{5.1e-3} & 2.6e+16 & 0.066 & \textbf{5.0e-3} & 0.060 & 12.724 \\
Sample MSE ($\downarrow$)         & 0.158 & \underline{0.128} & \underline{0.128} & 2.6e+16 & 0.164 & \textbf{0.124} & 0.142 & 14.448 \\
\midrule
Speed (s/sample) ($\downarrow$)  & 0.37 & \textbf{0.95} & \underline{1.06} & 50.87 & 17.41 & 1.74 & 17.08 & 28.45 \\
\bottomrule
\end{tabular}%
}
\end{table}

\subsubsection{Discussion}

The empirical results for the 1D Burgers' Equation illustrate the varying scalability of guidance methods when subjected to non-linear physical dynamics and the formation of steep spatial gradients (shock waves).

\paragraph{Performance under White Noise Prior} 
In the uncorrelated prior regime, optimization and proximal methods demonstrate distinct trade-offs. ECI achieves the best distributional fidelity (Sample MSE 0.110 in Partial, 0.116 in Full) and enforces Initial Conditions to machine precision, though it requires higher inference times ($\sim 18-21$s). DiffusionPDE provides a highly efficient balance, lowering the Sample MSE compared to the Vanilla baseline while running in near real-time ($\sim 1.1$s). ProFlow successfully minimizes the Raw PDE Residual (0.949) more effectively than other methods, but doing so slightly increases the Sample MSE (0.239) and incurs significant computational overhead ($\sim 18-21$s). D-Flow avoids numerical divergence but consistently yields elevated physical errors and requires the longest inference times ($\sim 32-39$s). Notably, while PCFM minimizes boundary errors in the Partial task, its explicit Jacobian formulation struggles fundamentally with the non-linear shock dynamics under Full-PDE constraints, resulting in severe numerical overflow (Physical MAE $>10^{14}$).

\paragraph{Prior Shift to Smooth Gaussian Kernels}
Transitioning to the spatially correlated prior exacerbates the challenges for certain baselines while favoring others. DiffusionPDE and ECI remain highly stable, effectively lowering the Sample MSE (0.124 and 0.142 in Full-PDE, respectively) compared to the Vanilla baseline (0.158). ProFlow maintains the lowest Raw PDE Residual (1.053) but remains computationally demanding. Conversely, D-Flow exhibits high sensitivity to the correlated prior, severely degrading the visual fidelity and causing its Sample MSE to spike above 10. PCFM continues to experience massive numerical overflow in the Full-PDE task. In the Partial task, PCFM artificially drives boundary errors below the Ground Truth numerical limit (stencil overfitting, marked with $^\dagger$), which structurally disconnects the flow from the learned manifold and degrades the Sample MSE ($0.223$) relative to Vanilla ($0.160$).

\paragraph{Robustness of the Proposed Method}
Our continuous frameworks (EGT and Hybrid EGT) robustly handle both the non-linear shocks of the Burgers' equation and the complexities of the prior shift. EGT consistently improves the Sample MSE over the Vanilla baseline while maintaining sub-second inference speeds. Hybrid EGT proves highly effective under Full-PDE constraints, achieving the lowest overall Physical Residual MAE ($1.5 \times 10^{-4}$ for White Noise and $1.8 \times 10^{-4}$ for Smooth Kernel) without relying on computationally expensive unrolled trajectory optimizations. Objectively, it must be noted that under the intense terminal refinement phase of Hybrid EGT in the Full-PDE task, the local IC and BC errors are driven below the numerical floor (marked with $^\dagger$). However, unlike strict algebraic projections, this localized boundary optimization does not negatively impact the global manifold, as Hybrid EGT still achieves a highly competitive Sample MSE (e.g., 0.128 under Smooth Kernel) at a fraction of the inference time ($\sim 1.06$s) required by ECI or ProFlow.

\subsection{1D Reaction-Diffusion Equation Results}
\label{app:results_rd}

Table \ref{tab:results_rd_master} presents the comprehensive evaluation for the 1D Reaction-Diffusion (RD) Equation. By introducing a non-linear reaction source term that actively competes with spatial diffusion, this PDE dictates complex transient dynamics and localized pattern formations. Consequently, it rigorously tests the solvers' ability to balance competing physical forces across the generative trajectory without falling into numerical instability.

\begin{table}[htbp]
\centering
\caption{Comprehensive evaluation metrics for the \textbf{1D Reaction-Diffusion Equation}. The table compares zero-shot guidance performance across Partial (Boundary/IC) and Full-PDE tasks using both White Noise and Smooth Gaussian priors. $\downarrow$ indicates lower is better, $\uparrow$ indicates higher is better. All physical metrics are averaged over the test set. Instances with a success rate lower than $100\%$ are marked as DNC.. $^\dagger$ indicates values that artificially fall below the discrete ground-truth stencil precision limit (stencil overfitting/waterbed effect).}
\label{tab:results_rd_master}
\resizebox{\textwidth}{!}{%
\begin{tabular}{l | c | c c c c c c c }
\toprule
\textbf{Metric} & \textbf{Vanilla} & \textbf{EGT (Ours)} & \textbf{Hybrid EGT (Ours)} & \textbf{PCFM} & \textbf{ProFlow} & \textbf{DiffPDE} & \textbf{ECI} & \textbf{D-Flow} \\
\midrule
\multicolumn{9}{l}{\textbf{PART 1: PARTIAL/BOUNDARY GUIDANCE}} \\
\midrule
\multicolumn{9}{l}{\textit{\textbf{Prior: White Noise}}} \\
\midrule
Success Rate ($\uparrow$)        & 100\% & \textbf{100\%} & \textbf{100\%} & \textbf{100\%} & \textbf{100\%} & \textbf{100\%} & \textbf{100\%} & \textbf{100\%} \\
Raw PDE Residual ($\downarrow$)  & 0.086 & \underline{0.222} & 0.423 & 0.716 & 7.3e+3 & \textbf{0.184} & 1.185 & 4.258 \\
Physical Res. MAE ($\downarrow$) & 0.200 & 0.075 & \underline{8.5e-3} & \textbf{1.1e-8}$^\dagger$ & 1.2e+3 & 0.043 & 0.054 & 0.739 \\
IC Error ($\downarrow$)          & 0.181 & 0.024 & 5.5e-6 & \textbf{1.4e-44} & 4.5e+5 & 7.8e-3 & \underline{1.5e-20} & 2.0e+2 \\
BC Error ($\downarrow$)          & 6.9e-4 & \underline{1.7e-3} & 0.032 & 0.154 & 7.8e+6 & \textbf{7.8e-4} & 0.224 & 3.040 \\
Mass Error ($\downarrow$)        & 4.5e-3 & \underline{0.011} & 0.017 & \textbf{2.5e-8}$^\dagger$ & 2.6e+3 & 0.016 & 0.124 & 1.077 \\
CL Error ($\downarrow$)          & 0.025 & 0.052 & 0.109 & \underline{0.047} & 17.208 & \textbf{0.026} & 0.175 & 0.054 \\
\midrule
MMSE ($\downarrow$)              & 2.0e-3 & 9.1e-3 & 8.9e-3 & \underline{3.1e-3} & 3.3e+3 & \textbf{1.2e-3} & 0.041 & 0.245 \\
SMSE ($\downarrow$)              & 4.6e-3 & \textbf{5.0e-3} & 5.2e-3 & 6.7e-3 & 7.4e+5 & \underline{5.1e-3} & 0.093 & 1.9e+2 \\
Sample MSE ($\downarrow$)         & 0.083 & 0.051 & 0.050 & \underline{0.041} & 7.4e+5 & \textbf{0.023} & 0.246 & 2.0e+2 \\
\midrule
Speed (s/sample) ($\downarrow$)  & 0.38 & \textbf{1.13} & \underline{1.24} & 2.16 & 25.65 & 2.00 & 24.47 & 32.07 \\
\midrule
\multicolumn{9}{l}{\textit{\textbf{Prior: Smooth Kernel}}} \\
\midrule
Success Rate ($\uparrow$)        & 100\% & \textbf{100\%} & \underline{100\%} & 100\% & 100\% & 100\% & 100\% & 100\% \\
Raw PDE Residual ($\downarrow$)  & 1.274 & 1.697 & \underline{1.460} & 3.636 & 2.604 & \textbf{1.309} & 8.682 & 10.993 \\
Physical Res. MAE ($\downarrow$) & 0.220 & 0.028 & \underline{8.5e-3} & \textbf{1.1e-8}$^\dagger$ & 0.239 & 0.027 & 0.086 & 0.026 \\
IC Error ($\downarrow$)          & 0.182 & 1.1e-3 & 4.6e-6 & \textbf{5.7e-44} & 0.219 & 4.6e-4 & \underline{1.4e-18} & 5.2e-3 \\
BC Error ($\downarrow$)          & 0.018 & 0.117 & \underline{0.039} & 0.164 & 0.084 & \textbf{0.017} & 1.118 & 0.088 \\
Mass Error ($\downarrow$)        & 0.053 & 0.027 & \underline{0.017} & \textbf{2.5e-8}$^\dagger$ & 0.056 & 0.043 & 0.198 & 0.038 \\
CL Error ($\downarrow$)          & 0.031 & \textbf{0.036} & 0.053 & 0.100 & 0.041 & \underline{0.041} & 0.091 & 0.071 \\
\midrule
MMSE ($\downarrow$)              & 1.4e-3 & 1.1e-3 & 1.0e-3 & \textbf{3.5e-4} & 0.058 & \underline{4.3e-4} & 0.044 & 1.7e-3 \\
SMSE ($\downarrow$)              & 3.8e-3 & 7.0e-4 & \underline{6.8e-4} & 0.036 & 0.043 & \textbf{6.3e-4} & 0.179 & 0.015 \\
Sample MSE ($\downarrow$)         & 0.088 & 0.010 & \underline{9.9e-3} & 0.087 & 0.191 & \textbf{9.3e-3} & 0.365 & 0.094 \\
\midrule
Speed (s/sample) ($\downarrow$)  & 0.38 & 1.09 & \underline{1.25} & 2.11 & 25.60 & 1.98 & 24.43 & 31.43 \\
\midrule
\midrule
\multicolumn{9}{l}{\textbf{PART 2: FULL-PDE GUIDANCE}} \\
\midrule
\multicolumn{9}{l}{\textit{\textbf{Prior: White Noise}}} \\
\midrule
Success Rate ($\uparrow$)        & 100\% & \textbf{100\%} & \underline{100\%} & 100\% & 100\% & 100\% & 100\% & 100\% \\
Raw PDE Residual ($\downarrow$)  & 0.086 & \underline{0.209} & 0.322 & 2.2e+9 & 4.8e+9 & \textbf{0.183} & 1.185 & 0.703 \\
Physical Res. MAE ($\downarrow$) & 3.5e-3 & 1.2e-3 & \textbf{2.0e-5} & 1.3e+5 & 2.9e+5 & 6.5e-3 & \underline{7.2e-5} & 6.6e-3 \\
IC Error ($\downarrow$)          & 0.181 & 0.021 & \underline{1.1e-10} & 1.5e+5 & 5.0e+11 & 7.7e-3 & \textbf{1.5e-20} & 0.168 \\
BC Error ($\downarrow$)          & 6.9e-4 & \textbf{5.6e-4} & 5.6e-3 & 7.6e+18 & 3.0e+11 & \underline{7.8e-4} & 0.224 & 0.085 \\
Mass Error ($\downarrow$)        & 4.5e-3 & \underline{0.034} & 0.085 & 6.5e+14 & 2.5e+9 & \textbf{0.015} & 0.124 & 0.104 \\
CL Error ($\downarrow$)          & 0.025 & 0.056 & 0.106 & 1.0e+5 & 8.8e+2 & \textbf{0.026} & 0.175 & \underline{0.045} \\
\midrule
MMSE ($\downarrow$)              & 1.9e-3 & 8.0e-3 & 8.0e-3 & 8.6e+13 & 3.3e+9 & \textbf{1.2e-3} & 0.041 & \underline{2.2e-3} \\
SMSE ($\downarrow$)              & 4.6e-3 & 5.2e-3 & \underline{5.2e-3} & 6.7e+16 & 5.0e+11 & \textbf{5.0e-3} & 0.093 & 0.065 \\
Sample MSE ($\downarrow$)         & 0.083 & \underline{0.049} & \underline{0.049} & 6.7e+16 & 5.0e+11 & \textbf{0.023} & 0.246 & 0.169 \\
\midrule
Speed (s/sample) ($\downarrow$)  & 0.376 & \textbf{0.891} & \underline{0.928} & 76.720 & 25.055 & 1.860 & 24.343 & 31.443 \\
\midrule
\multicolumn{9}{l}{\textit{\textbf{Prior: Smooth Kernel}}} \\
\midrule
Success Rate ($\uparrow$)        & 100\% & \textbf{100\%} & \textbf{100\%} & \textbf{100\%} & \textbf{100\%} & \textbf{100\%} & \textbf{100\%} & \textbf{100\%} \\
Raw PDE Residual ($\downarrow$)  & 1.274 & 1.335 & \underline{1.324} & 9.8e+8 & 2.609 & \textbf{1.310} & 8.682 & 12.475 \\
Physical Res. MAE ($\downarrow$) & 3.6e-3 & \underline{3.6e-4} & \textbf{8.0e-5} & 5.9e+4 & 4.0e-3 & \underline{2.3e-4} & 5.2e-4 & 1.5e-3 \\
IC Error ($\downarrow$)          & 0.182 & 1.1e-3 & \underline{9.9e-12} & 1.7e+6 & 0.219 & 4.6e-4 & \textbf{1.4e-18} & 6.214 \\
BC Error ($\downarrow$)          & 0.018 & \underline{0.017} & \textbf{0.017} & 2.4e+19 & 0.088 & 0.018 & 1.118 & 3.369 \\
Mass Error ($\downarrow$)        & 0.053 & 0.052 & \textbf{0.047} & 2.1e+15 & 0.082 & \underline{0.051} & 0.198 & 0.153 \\
CL Error ($\downarrow$)          & 0.031 & \textbf{0.036} & 0.052 & 2.0e+5 & \underline{0.041} & \underline{0.041} & 0.091 & 0.107 \\
\midrule
MMSE ($\downarrow$)              & 1.3e-3 & \underline{1.0e-3} & \underline{1.0e-3} & 2.1e+14 & 0.058 & \textbf{4.3e-4} & 0.044 & 3.7e-3 \\
SMSE ($\downarrow$)              & 3.8e-3 & \underline{6.6e-4} & \underline{6.6e-4} & 2.1e+17 & 0.044 & \textbf{6.3e-4} & 0.179 & 1.138 \\
Sample MSE ($\downarrow$)         & 0.088 & 0.010 & \underline{9.8e-3} & 2.1e+17 & 0.191 & \textbf{9.4e-3} & 0.365 & 1.530 \\
\midrule
Speed (s/sample) ($\downarrow$)  & 0.375 & \textbf{0.889} & \underline{0.951} & 77.715 & 25.099 & 1.857 & 24.395 & 31.313 \\
\bottomrule
\end{tabular}%
}
\end{table}

\subsubsection{Discussion}

The empirical results for the Reaction-Diffusion Equation heavily expose the structural limitations of rigid algebraic projections when faced with coupled non-linear PDE constraints over dense temporal grids.

\paragraph{Performance under White Noise Prior} Under standard White Noise and simple Partial (Boundary/IC) constraints, most baseline methods maintain reasonable stability. However, scaling to the Full-PDE task exposes the scalability limits of explicit projection frameworks. Both PCFM and ProFlow encounter severe numerical instabilities along the flow; although they technically complete the generation without throwing NaN exceptions (100\% Success Rate), their Raw PDE Residuals experience severe overflow, reaching $2.2 \times 10^9$ and $4.8 \times 10^9$, respectively. In contrast, DiffPDE and ECI handle the White Noise environment more reliably, maintaining sub-optimal but stable physical residuals without diverging.

\paragraph{Prior Shift and stencil overfitting} The introduction of the spatially correlated Smooth Kernel prior further amplifies the divergence between continuous guidance and post-hoc approximations. The Waterbed Effect is starkly visible in PCFM during Partial tasks: it drives the physical and boundary residuals down to precision limits (marked with $^\dagger$), but this aggressive optimization of the discrete stencil severely degrades the underlying manifold, resulting in poor Sample MSE scores. Furthermore, the prior shift heavily impacts optimization-based methods on Full-PDE tasks. For example, ECI, which performed admirably under White Noise, struggles to interpolate the correlated noise, causing its Sample MSE to degrade to 0.365 under the Smooth Kernel Full-PDE task.

\paragraph{Robustness of the Proposed Method} Our continuous guidance architectures (EGT and Hybrid EGT) stand out as the most robust and computationally efficient frameworks across all configurations. Hybrid EGT consistently excels in the challenging Full-PDE tasks, achieving the best Physical Residual MAE under the White Noise prior ($2.0 \times 10^{-5}$) and maintaining exceptional fidelity under the Smooth Kernel ($8.0 \times 10^{-5}$). Most importantly, our framework enforces these complex non-linear dynamics without resorting to stencil overfitting and without compromising the macroscopic visual coherence of the fields (maintaining top-tier Sample MSE). Beyond raw accuracy, EGT significantly outperforms optimization-based baselines in computational efficiency: it resolves the Full-PDE Smooth Kernel task in just $\sim 0.95$ seconds per sample, bypassing the severe bottlenecks observed in ECI ($\sim 24$ seconds) and PCFM ($\sim 77$ seconds).

\subsection{2D Navier-Stokes Equations Results}
\label{app:results_ns}

Table \ref{tab:results_ns_master} presents the comprehensive evaluation for the 2D Navier-Stokes (NS) Equations. As the only two-dimensional macroscopic fluid dynamics environment in our benchmark, this task requires enforcing complex vorticity structures, non-linear advection, and strict incompressibility over a dense spatial grid. This environment pushes the structural and computational limits of guided generative models.

\begin{table}[htbp]
\centering
\caption{Comprehensive evaluation metrics for the \textbf{2D Navier-Stokes Equations}. The table evaluates zero-shot guidance performance across Partial (Boundary/IC) and Full-PDE tasks using a White Noise prior. $\downarrow$ indicates lower is better, $\uparrow$ indicates higher is better. All physical metrics are averaged over the test set. Instances with a success rate lower than $100\%$ are marked as DNC. $^\dagger$ indicates values that artificially fall below the discrete ground-truth stencil precision limit (stencil overfitting/waterbed effect).}
\label{tab:results_ns_master}
\resizebox{\textwidth}{!}{%
\begin{tabular}{l | c | c c c c c c c }
\toprule
\textbf{Metric} & \textbf{Vanilla} & \textbf{EGT (Ours)} & \textbf{Hybrid EGT (Ours)} & \textbf{PCFM} & \textbf{ProFlow} & \textbf{DiffPDE} & \textbf{ECI} & \textbf{D-Flow} \\
\midrule
\multicolumn{9}{l}{\textbf{PART 1: PARTIAL/BOUNDARY GUIDANCE}} \\
\midrule
\multicolumn{9}{l}{\textit{\textbf{Prior: White Noise}}} \\
\midrule
Success Rate ($\uparrow$)        & 100\% & \textbf{100\%} & \textbf{100\%} & \textbf{100\%} & \textbf{100\%} & \textbf{100\%} & \textbf{100\%} & 14.2\% \\
Raw PDE Residual ($\downarrow$)  & 0.063 & \underline{0.060} & 0.060 & 0.075 & \textbf{0.057} & 0.070 & 0.071 & DNC \\
Physical Res. MAE ($\downarrow$) & 0.291 & 0.010 & 1.4e-4 & \textbf{1.2e-10}$^\dagger$ & 0.197 & 0.124 & \underline{3.8e-5} & DNC \\
IC Error ($\downarrow$)          & 0.136 & 1.7e-4 & 2.4e-8 & \textbf{0.0} & 0.063 & 0.025 & \underline{8.9e-17} & DNC \\
BC Error ($\downarrow$)          & 0.058 & \textbf{0.056} & \underline{0.057} & 0.066 & 0.058 & \underline{0.057} & 0.060 & DNC \\
Mass Error ($\downarrow$)        & 0.011 & 7.6e-3 & 6.2e-3 & \textbf{1.0e-8}$^\dagger$ & \underline{3.1e-3} & 0.039 & 3.2e-3 & DNC \\
CL Error ($\downarrow$)          & 0.012 & 7.6e-3 & 6.1e-3 & \textbf{1.0e-8} & 3.3e-3 & 0.038 & \underline{3.1e-3} & DNC \\
\midrule
MMSE ($\downarrow$)              & 0.018 & \textbf{0.018} & \underline{0.018} & 0.037 & 0.020 & 0.036 & 0.116 & DNC \\
SMSE ($\downarrow$)              & 0.011 & \underline{0.011} & \underline{0.011} & 0.023 & \textbf{9.9e-3} & \underline{0.011} & \underline{0.011} & DNC \\
Sample MSE ($\downarrow$)         & 0.542 & \textbf{0.529} & 0.548 & 0.756 & 0.550 & \underline{0.541} & 0.630 & DNC \\
\midrule
Speed (s/sample) ($\downarrow$)  & 0.690 & \textbf{0.714} & 0.898 & 93.972 & \underline{0.847} & 1.402 & 3.559 & DNC \\
\midrule
\midrule
\multicolumn{9}{l}{\textbf{PART 2: FULL-PDE GUIDANCE}} \\
\midrule
\multicolumn{9}{l}{\textit{\textbf{Prior: White Noise}}} \\
\midrule
Success Rate ($\uparrow$)        & 100\% & \textbf{100\%} & \textbf{100\%} & 0\% & \textbf{100\%} & \textbf{100\%} & \textbf{100\%} & 1.6\% \\
Raw PDE Residual ($\downarrow$)  & 0.062 & 0.080 & 0.124 & DNC & \textbf{0.057} & \underline{0.069} & 0.070 & DNC \\
Physical Res. MAE ($\downarrow$) & 7.5e-3 & \underline{1.6e-3} & \textbf{2.9e-5} & DNC & 5.6e-3 & 4.2e-3 & 1.8e-3 & DNC \\
IC Error ($\downarrow$)          & 0.136 & 1.3e-3 & \underline{4.9e-14} & DNC & 0.064 & 0.025 & \textbf{8.9e-17} & DNC \\
BC Error ($\downarrow$)          & 0.059 & \underline{0.034} & \textbf{1.0e-6} & DNC & 0.057 & 0.057 & 0.059 & DNC \\
Mass Error ($\downarrow$)        & 9.4e-3 & 7.5e-3 & 6.4e-3 & DNC & \textbf{3.1e-3} & 0.035 & \underline{3.2e-3} & DNC \\
CL Error ($\downarrow$)          & 9.9e-3 & 7.6e-3 & 6.2e-3 & DNC & \underline{3.3e-3} & 0.035 & \textbf{3.1e-3} & DNC \\
\midrule
MMSE ($\downarrow$)              & 0.019 & \textbf{0.018} & \textbf{0.018} & DNC & \underline{0.022} & 0.032 & 0.098 & DNC \\
SMSE ($\downarrow$)              & 0.011 & 0.011 & 0.011 & DNC & \textbf{9.8e-3} & \underline{0.010} & \underline{0.010} & DNC \\
Sample MSE ($\downarrow$)         & 0.559 & \textbf{0.547} & \textbf{0.547} & DNC & \underline{0.549} & 0.553 & 0.611 & DNC \\
\midrule
Speed (s/sample) ($\downarrow$)  & 0.858 & \textbf{0.937} & \underline{1.021} & DNC & 1.209 & 1.789 & 4.400 & DNC \\
\bottomrule
\end{tabular}%
}
\end{table}

\subsubsection{Discussion}

The empirical results for the Navier-Stokes equations decisively illustrate the scalability limits and structural fragilities of explicit projection methods when applied to high-dimensional physical systems.

\paragraph{Baseline Performance and Guidance Collapse} Scaling the constraint complexity to a 2D grid exposes severe instabilities in baseline frameworks. D-Flow experiences significant guidance degradation, achieving a mere 14.2\% Success Rate on Partial constraints and dropping to 1.6\% on the Full-PDE task. More critically, PCFM experiences total numerical divergence under Full-PDE constraints (0\% SR, yielding NaN metrics), indicating that its reliance on explicit Jacobian approximations is computationally and mathematically restrictive for complex macroscopic fluid flows. While methods like ProFlow, DiffPDE, and ECI technically complete the Full-PDE generation, they struggle to simultaneously satisfy the physics and preserve the underlying data distribution.

\paragraph{High-Dimensional stencil overfitting} The Waterbed Effect is exceptionally prominent in this 2D setting. Under Partial constraints, PCFM algebraically forces the Initial Condition error to absolute zero (0.0) and the Physical Residual to precision limits ($1.2 \times 10^{-10}$, marked with $^\dagger$). However, this aggressive structural constraint severely distorts the generative trajectory, heavily degrading the Sample MSE to 0.756 (markedly worse than the unguided Vanilla model at 0.542). ECI similarly exhibits stencil overfitting on the IC Error ($8.9 \times 10^{-17}$) while suffering from visual and structural degradation (Sample MSE of 0.611 on the Full-PDE task).

\paragraph{Robustness of the Proposed Method} Our Continuous Dissipative Guidance architectures (EGT and Hybrid EGT) successfully navigate the high-dimensional fluid manifold without succumbing to numerical divergence or stencil overfitting. On the highly constrained Full-PDE task, Hybrid EGT decisively outperforms all baselines, achieving a state-of-the-art Physical Residual MAE of $2.9 \times 10^{-5}$ and driving boundary errors down to $1.0 \times 10^{-6}$. Crucially, our framework accomplishes this while perfectly preserving the structural integrity of the fluid (Sample MSE of 0.547). Beyond accuracy, EGT demonstrates unmatched scalability: it strictly satisfies the complex 2D Navier-Stokes physics in just $\sim 1.02$ seconds per sample, completely bypassing the extreme computational bottlenecks of algebraic methods (e.g., PCFM requiring nearly 94 seconds per sample on Partial tasks).

\subsection{\rev{Macroscopic Physical Diagnostics}}
\label{app:macroscopic_diagnostics}

\rev{To evaluate how pointwise PDE residual minimization affects broader macroscopic statistics, we compute two additional holistic diagnostics on the generated samples from the main evaluation (White Noise, Full-PDE configuration): the 1D Wasserstein Distance between spatial gradient distributions (capturing shock statistics on 1D Burgers) and the Energy Spectrum $L^1$ Error (assessing spectral consistency on 2D Navier--Stokes).}

\begin{table}[htbp]
\centering
\caption{\rev{Macroscopic physical diagnostics on 1D Burgers and 2D Navier--Stokes (White Noise, Full-PDE).}}
\label{tab:macroscopic_diagnostics}
\begin{tabular}{l l l c}
\toprule
\textbf{Benchmark} & \textbf{Model} & \textbf{Diagnostic Metric} & \textbf{Mean $\pm$ Std} $\downarrow$ \\
\midrule
\multirow{5}{*}{Burgers 1D} 
 & Vanilla          & Gradient Wasserstein Dist. & 5.855e-1 $\pm$ 2.577e-1 \\
 & DiffusionPDE     & Gradient Wasserstein Dist. & 4.441e-1 $\pm$ 2.239e-1 \\
 & ECI              & Gradient Wasserstein Dist. & \textbf{3.819e-1 $\pm$ 2.327e-1} \\
 & \textbf{Ours (Pure EGT)}   & Gradient Wasserstein Dist. & 6.870e-1 $\pm$ 3.010e-1 \\
 & \textbf{Ours (Hybrid EGT)} & Gradient Wasserstein Dist. & 8.399e-1 $\pm$ 3.687e-1 \\
\midrule
\multirow{5}{*}{Navier--Stokes 2D} 
 & Vanilla          & Energy Spectrum $L^1$ Error & 9.630e-2 $\pm$ 13.06e-2 \\
 & DiffusionPDE     & Energy Spectrum $L^1$ Error & 3.299e-1 $\pm$ 2.722e-1 \\
 & ECI              & Energy Spectrum $L^1$ Error & \textbf{2.740e-2 $\pm$ 1.870e-2} \\
 & \textbf{Ours (Pure EGT)}   & Energy Spectrum $L^1$ Error & 9.840e-2 $\pm$ 13.09e-2 \\
 & \textbf{Ours (Hybrid EGT)} & Energy Spectrum $L^1$ Error & \underline{8.570e-2 $\pm$ 11.43e-2} \\
\bottomrule
\end{tabular}
\end{table}

\paragraph{\rev{Discussion.}} \rev{Table~\ref{tab:macroscopic_diagnostics} highlights an important distinction between local pointwise consistency and global statistical realism. Because the current EGT formulation explicitly minimizes the pointwise $L^2$ norm of the PDE, IC, and BC residuals, it substantially reduces local physical errors (and improves spectral consistency over DiffusionPDE on 2D Navier--Stokes) but does not directly optimize gradient distributions on 1D Burgers. Because EGT can accommodate any differentiable energy functional, augmenting $E(u)$ with differentiable spectral or gradient-matching penalties represents a natural extension for jointly optimizing local and macroscopic realism.}

\subsection{\rev{Statistical Significance and Seed-to-Seed Variability}}
\label{app:statistical_significance}

\rev{To verify the statistical robustness of our evaluation, we first compare the empirical standard deviations computed across $100$ and $1,000$ generated samples on the 1D Burgers and 2D Navier--Stokes benchmarks (White Noise, Full-PDE). As shown in Table~\ref{tab:variance_convergence}, the standard deviations are virtually identical, confirming that $100$ samples provide reliable variance estimates. We therefore evaluate seed-to-seed variability by generating $100$ samples across $5$ independent random seeds (Table~\ref{tab:seed_variability}, using the linear scheduler for EGT based on our scheduler ablation). EGT exhibits consistently low seed-to-seed variance across both systems, achieving a lower physical MAE ($1.49\text{e-}3$ vs.\ $1.81\text{e-}3$) and more than $2\times$ lower Sample MSE variance ($1.42\text{e-}2$ vs.\ $3.06\text{e-}2$) than ECI on 2D Navier--Stokes.}

\begin{table}[htbp]
\centering
\caption{\rev{Variance convergence analysis comparing standard deviations across $100$ and $1,000$ generated samples (White Noise, Full-PDE).}}
\label{tab:variance_convergence}
\resizebox{\textwidth}{!}{%
\begin{tabular}{l l c c c c}
\toprule
\textbf{Model} & \textbf{Metric} & \textbf{Std NS (100)} & \textbf{Std NS (1000)} & \textbf{Std Burgers (100)} & \textbf{Std Burgers (1000)} \\
\midrule
\textbf{Ours (EGT)} & Sample MSE         & 2.292e-1 & 2.321e-1 & 9.750e-2 & 1.208e-1 \\
\textbf{Ours (EGT)} & Physical MAE       & 3.000e-4 & 3.000e-4 & 5.000e-4 & 6.000e-4 \\
ECI                 & Sample MSE         & 3.746e-1 & 4.020e-1 & 7.380e-2 & 9.500e-2 \\
ECI                 & Physical MAE       & 6.000e-4 & 6.000e-4 & 1.000e-4 & 1.000e-4 \\
DiffusionPDE        & Sample MSE         & 2.448e-1 & 2.297e-1 & 9.700e-2 & 1.147e-1 \\
DiffusionPDE        & Physical MAE       & 6.000e-4 & 8.000e-4 & 4.000e-4 & 4.000e-4 \\
Vanilla             & Sample MSE         & 2.302e-1 & 2.339e-1 & 1.059e-1 & 1.329e-1 \\
Vanilla             & Physical MAE       & 1.000e-3 & 1.400e-3 & 2.100e-3 & 2.500e-3 \\
\bottomrule
\end{tabular}%
}
\end{table}

\begin{table}[htbp]
\centering
\caption{\rev{Seed-to-seed variability across $5$ independent random seeds ($100$ samples per seed, White Noise, Full-PDE).}}
\label{tab:seed_variability}
\resizebox{\textwidth}{!}{%
\begin{tabular}{l c c c c}
\toprule
\textbf{Model} & \textbf{Burgers (Sample MSE)} $\downarrow$ & \textbf{Burgers (Phys MAE)} $\downarrow$ & \textbf{NS (Sample MSE)} $\downarrow$ & \textbf{NS (Phys MAE)} $\downarrow$ \\
\midrule
Vanilla             & 0.163 $\pm$ 9.82e-3 & 4.54e-3 $\pm$ 2.18e-4 & 0.541 $\pm$ 1.43e-2 & 6.78e-3 $\pm$ 9.76e-5 \\
DiffusionPDE        & 0.139 $\pm$ 1.04e-2 & 1.71e-3 $\pm$ 5.89e-5 & 0.547 $\pm$ 1.70e-2 & 3.72e-3 $\pm$ 7.40e-5 \\
ECI                 & \textbf{0.127 $\pm$ 6.12e-3} & \textbf{2.05e-4 $\pm$ 4.98e-6} & 0.620 $\pm$ 3.06e-2 & 1.81e-3 $\pm$ 4.52e-5 \\
\textbf{Ours (EGT)} & 0.146 $\pm$ 9.13e-3 & \underline{6.04e-4 $\pm$ 1.20e-5} & \textbf{0.528 $\pm$ 1.42e-2} & \textbf{1.49e-3 $\pm$ 2.50e-5} \\
\bottomrule
\end{tabular}%
}
\end{table}

\section{Ablation Studies}
\label{app:ablations}

In this appendix, we provide an extensive sensitivity analysis of the core hyperparameters governing Continuous Dissipative Guidance. To evaluate the robustness of our framework across different dimensionalities, we conduct all ablation studies on the 1D Burgers' (Full-PDE) and 2D Navier-Stokes (Full-PDE) datasets. These cases represent the challenging scenarios where the unguided model's physics violation is most prominent. For consistency across all ablation trials, we utilize the standard White Noise prior $p_0$. All tests are evaluated on $256$ samples.

\subsection{Sensitivity to Guidance Scale $\gamma$}
\label{app:ablation_gamma}

\begin{figure}[htbp]
    \centering
    \begin{minipage}{0.48\textwidth}
        \centering
        \includegraphics[width=\textwidth]{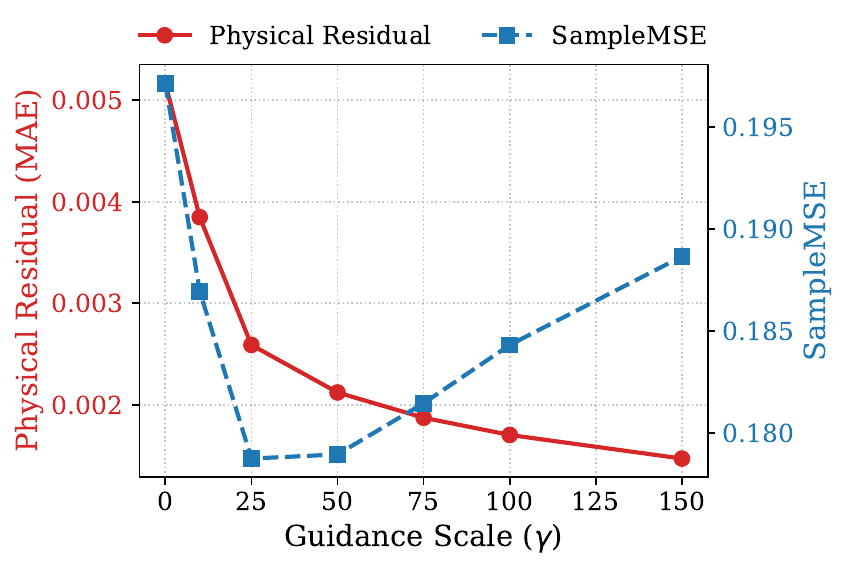}
        \caption*{ (a) Burgers' Dual-Axis }
    \end{minipage}
    \hfill
    \begin{minipage}{0.48\textwidth}
        \centering
        \includegraphics[width=\textwidth]{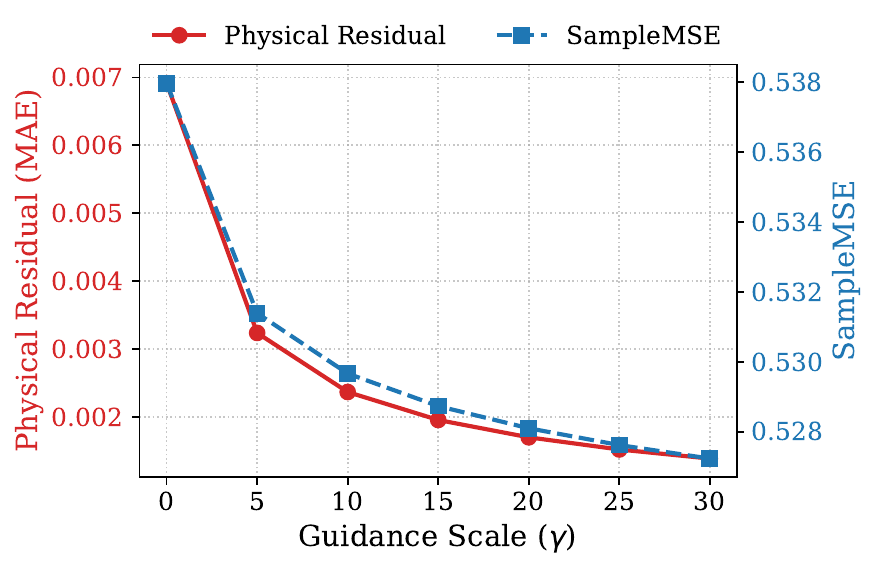}
        \caption*{ (b) NS' Dual-Axis }
    \end{minipage}
    \caption{Impact of the guidance scale $\gamma$. Both dual-axis plots simultaneously track the Physical Residual MAE (physics) and Sample MSE (fidelity) on 1D Burgers' (a) and 2D Navier-Stokes (b).}
    \label{fig:ablation_gamma}
\end{figure}

The guidance scale $\gamma$ controls the magnitude of the dissipative force injected into the generative vector field. Figure \ref{fig:ablation_gamma} illustrates the trade-off between physical adherence and distributional fidelity as a function of $\gamma$.

\paragraph{Discussion.} As shown in the dual-axis plots, increasing $\gamma$ triggers a rapid dissipative behaviour of the physical residual. In 1D Burgers', we observe that the physical MAE reaches a stable floor already at $\gamma \approx 25$, with negligible impact on the Sample MSE. This confirms that our continuous guidance effectively steers trajectories toward the physical manifold without causing the ``manifold collapse'' commonly observed in rigid algebraic projection methods. For 2D Navier-Stokes, the energy landscape is significantly more complex; while EGT successfully navigates this space and improves physical fidelity up to $\gamma = 30$, pushing the guidance scale beyond this threshold disrupts the generative trajectory and results in numerical instability. \rev{Importantly, selecting $\gamma$ in a genuine zero-shot regime does not require access to ground-truth test solutions: since the physical energy $E(u)$ depends exclusively on spatial differential operators evaluated on the generated samples, practitioners can simply perform a validation sweep on a small batch of initial noise, progressively increasing $\gamma$ and selecting the largest value before numerical stiffness or gradient divergence (NaNs) occurs.}

\subsection{Number of Flow Matching Steps $N$}
\label{app:ablation_steps}

\begin{figure}[htbp]
    \centering
    \begin{minipage}{0.48\textwidth}
        \centering
        \includegraphics[width=\textwidth]{plots/ablation_steps_burgers.pdf}
        \caption*{ (a) Burgers' Budget Ablation }
    \end{minipage}
    \hfill
    \begin{minipage}{0.48\textwidth}
        \centering
        \includegraphics[width=\textwidth]{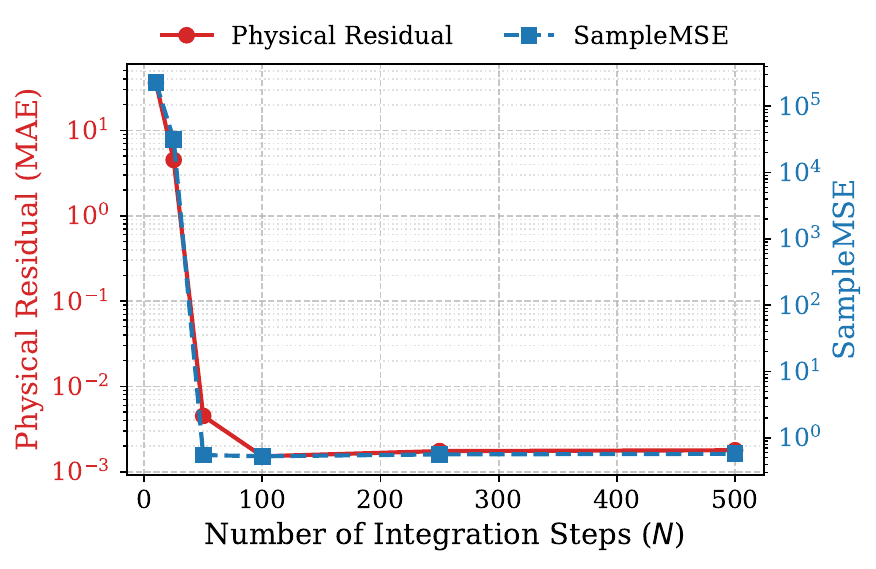}
        \caption*{ (b) NS' Budget Ablation }
    \end{minipage}
    \caption{Evaluation of ODE integration steps $N$. Both environments demonstrate that EGT stabilizes quickly. Note that for 2D Navier-Stokes (b), the y-axis for the Sample MSE is strictly logarithmic to accommodate the severe error variance at highly constrained integration budgets.}
    \label{fig:ablation_steps}
\end{figure}

We investigate the performance and computational efficiency of our framework under different discretization budgets for the ODE solver, ranging from $N=10$ to $N=250$ generative steps (Figure \ref{fig:ablation_steps}).

\paragraph{Discussion.} Unlike standard diffusion or heavy explicit projection models that often require hundreds of steps for stability, our energy-guided FFM maintains rigorous physical consistency even at highly accelerated integration budgets ($N \approx 50$). For instance, in the complex 2D Navier-Stokes setting (where the Sample MSE is plotted on a logarithmic y-axis to capture the extreme variance), lowering $N$ below $25$ leads to severe numerical divergence. However, at just $N=50$, the framework instantly stabilizes, achieving a Sample MSE perfectly aligned with the unguided baseline. While the physical residual further decreases with finer discretization, the relative gain plateaus around $N=100$. This scalability is a significant practical advantage for real-time physics simulators where sub-second inference is required.

\subsection{Effect of Guidance Scheduling}
\label{app:ablation_scheduler}

Finally, we compare different scheduling functions $\gamma(t)$ for the temporal application of the dissipative force. We categorize these into a Constant baseline, ``Late Guidance'' schedules (Linear, Parabolic, Cosine), and a ``Transient'' schedule (Sine).

\begin{figure}[htbp]
    \centering
    \begin{minipage}{0.48\textwidth}
        \centering
        \includegraphics[width=\textwidth]{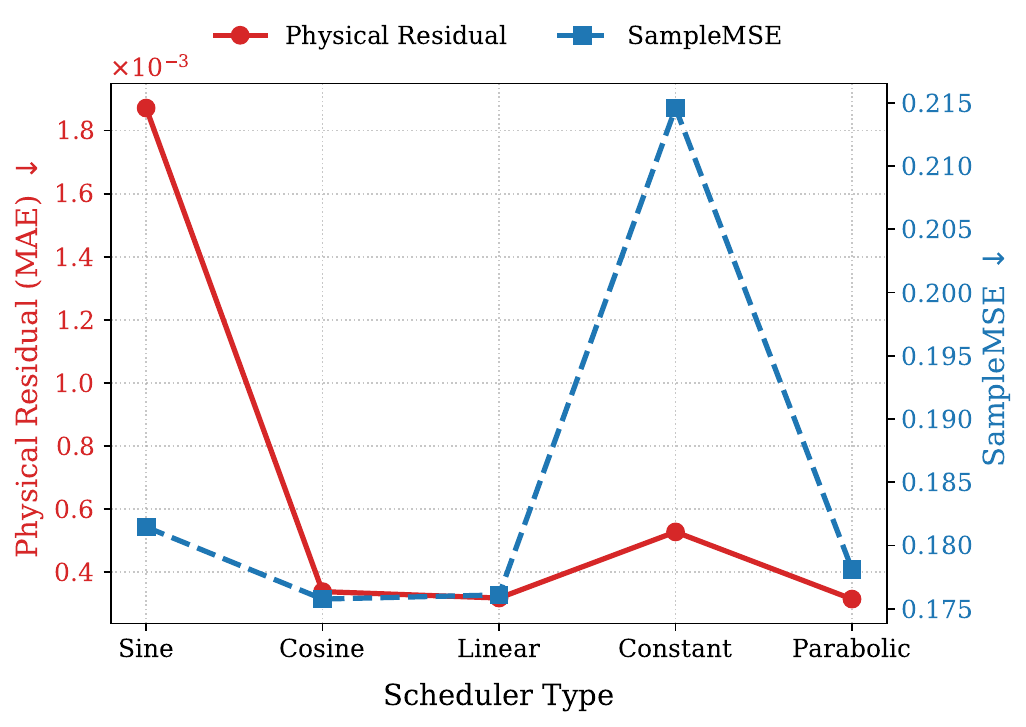}
        \caption*{ (a) Burgers' Scheduler Ablation }
    \end{minipage}
    \hfill
    \begin{minipage}{0.48\textwidth}
        \centering
        \includegraphics[width=\textwidth]{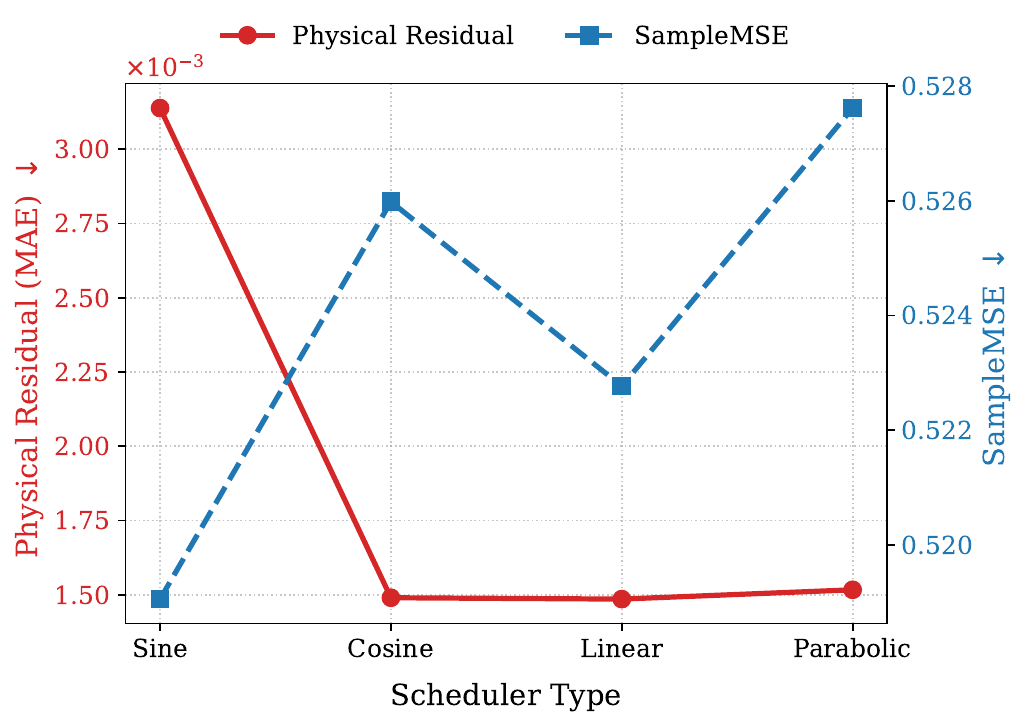}
        \caption*{ (b) NS' Scheduler Ablation }
    \end{minipage}
    \caption{Impact of the temporal guidance scheduler $\gamma(t)$. Both ``Late Guidance'' and ``Transient'' schedules systematically prevent trajectory collapse by allowing the base neural prior to establish macroscopic coherent structures before rigid physical constraints are fully applied.}
    \label{fig:ablation_schedulers}
\end{figure}

\paragraph{Discussion.} Our empirical analysis strongly validates the necessity of ``Late Guidance'' schedules (Linear, Parabolic, Cosine) and the ``Transient'' schedule (Sine). By allowing the unguided neural prior to freely establish the macroscopic structure of the field near pure noise ($t \approx 0$) before smoothly ramping up the physical constraints, these dynamic schedules avoid interfering with the inherently high variance of the initial noise. Most notably, applying a Constant guidance schedule throughout the flow results in a complete divergence ($0\%$ Success Rate) on the 2D Navier-Stokes environment, and a noticeable degradation of Sample MSE in the Burgers' equation. Consequently, smoothly scheduling the dissipative force acts as a necessary annealing process, leading to both optimal Sample MSE and lower final physical residuals.

\subsection{\rev{Pareto Frontier Analysis: Accuracy vs. Stability vs. Speed}}
\label{app:pareto_analysis}

\begin{figure}[htbp]
    \centering
    \begin{subfigure}[b]{0.48\textwidth}
        \centering
        \includegraphics[width=\textwidth]{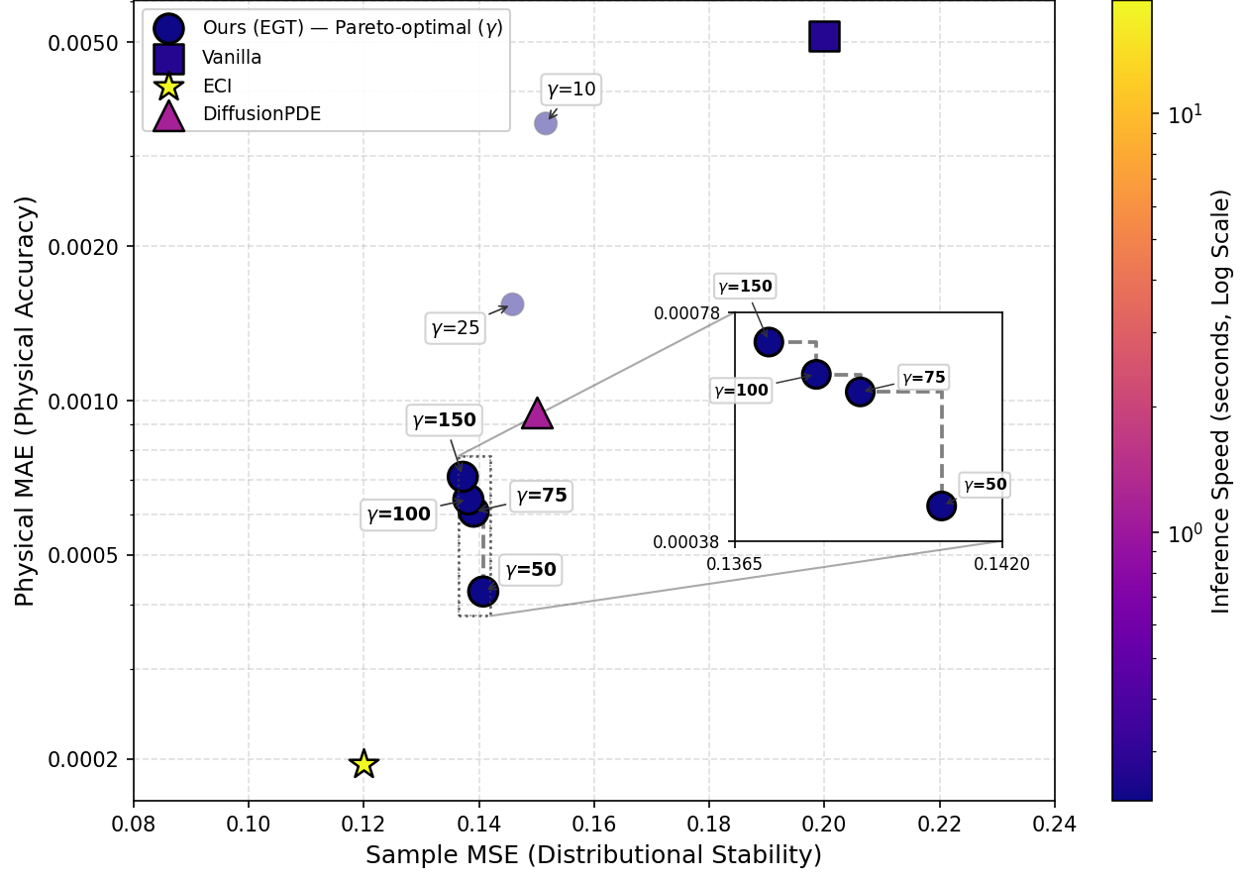}
        \caption{\rev{Burgers 1D ($\gamma$ Ablation)}}
        \label{fig:pareto_burgers_gamma}
    \end{subfigure}
    \hfill
    \begin{subfigure}[b]{0.48\textwidth}
        \centering
        \includegraphics[width=\textwidth]{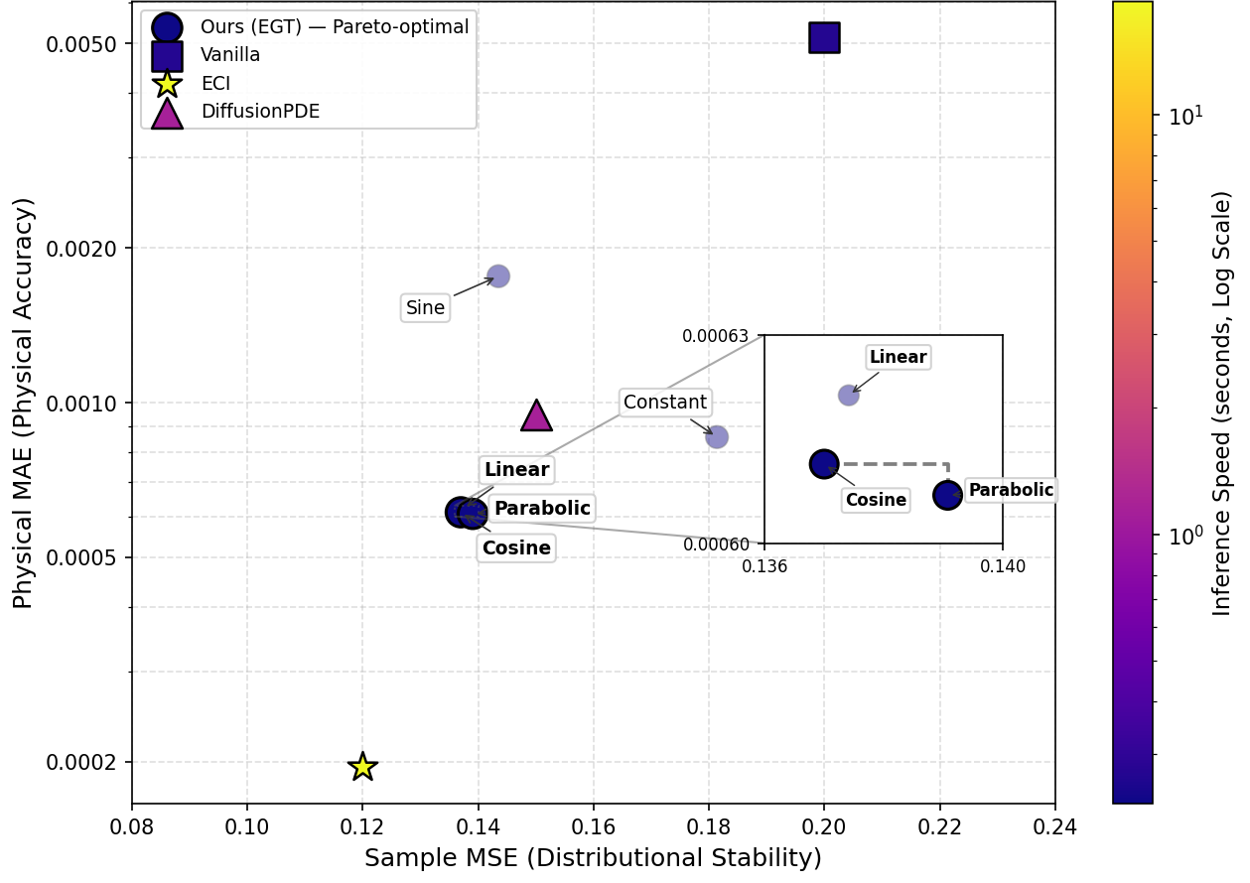}
        \caption{\rev{Burgers 1D (Scheduler Ablation)}}
        \label{fig:pareto_burgers_sched}
    \end{subfigure}
    
    \vspace{0.3cm}
    
    \begin{subfigure}[b]{0.48\textwidth}
        \centering
        \includegraphics[width=\textwidth]{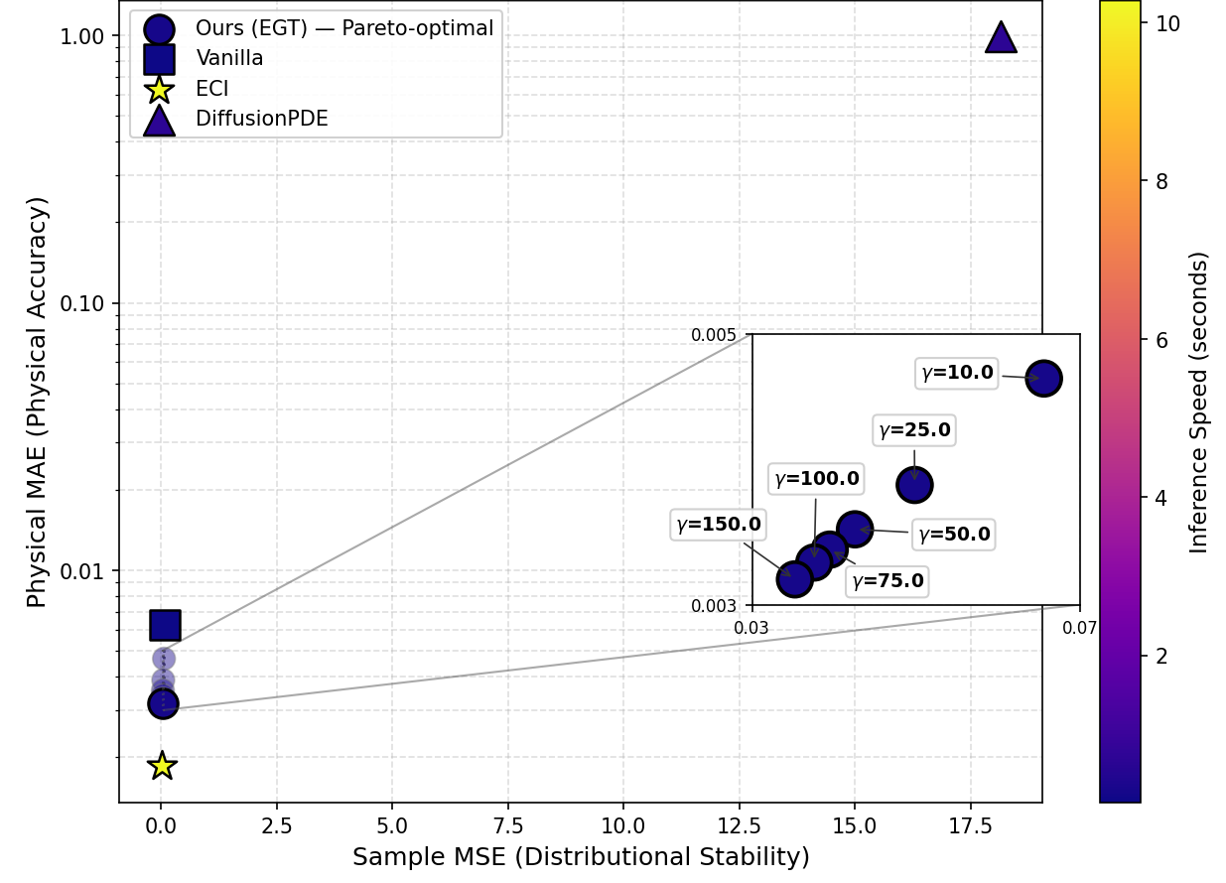}
        \caption{\rev{Heat 1D ($\gamma$ Ablation)}}
        \label{fig:pareto_heat_gamma}
    \end{subfigure}
    \hfill
    \begin{subfigure}[b]{0.48\textwidth}
        \centering
        \includegraphics[width=\textwidth]{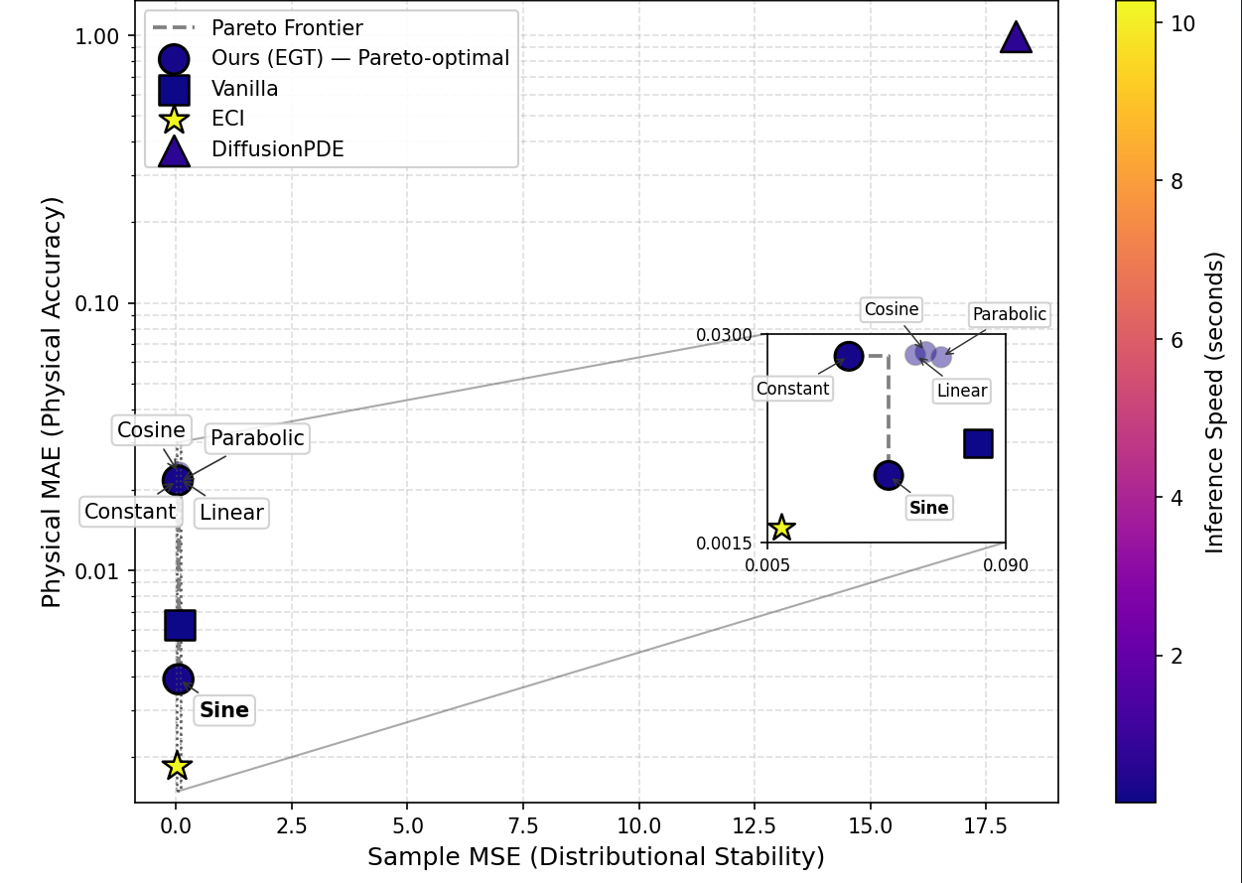}
        \caption{\rev{Heat 1D (Scheduler Ablation)}}
        \label{fig:pareto_heat_sched}
    \end{subfigure}
    \caption{\rev{Pareto frontiers of distributional stability (Sample MSE) versus physical accuracy (Physical MAE), with inference speed (sec/sample) encoded via the colormap. EGT consistently dominates Vanilla and DiffusionPDE while operating $40$--$80\times$ faster than ECI.}}
    \label{fig:pareto_frontiers}
\end{figure}

\rev{To evaluate how EGT navigates the accuracy--stability--speed trade-off compared to projection- and gradient-based baselines, we compute the Pareto frontiers on the 1D Burgers and 1D Heat benchmarks (Full-PDE, $N=100$) across guidance scales ($\gamma$) and schedulers. Figure~\ref{fig:pareto_frontiers} visualizes the four resulting trade-off curves, while Tables~\ref{tab:pareto_burgers} and \ref{tab:pareto_heat} report the exact numerical operating points.}

\begin{table}[htbp]
\centering
\caption{\rev{Guidance strength ($\gamma$) and scheduler Pareto ablation on the \textbf{1D Burgers} system. EGT provides a controllable trade-off between physical accuracy and distributional fidelity while maintaining a constant sub-second runtime ($0.23$ s/sample).}}
\label{tab:pareto_burgers}
\resizebox{\textwidth}{!}{%
\begin{tabular}{l c c c | l c c c}
\toprule
\textbf{Method ($\gamma$ Ablation)} & \textbf{Speed (s)} $\downarrow$ & \textbf{Phys MAE} $\downarrow$ & \textbf{Sample MSE} $\downarrow$ & \textbf{Method (Scheduler Ablation)} & \textbf{Speed (s)} $\downarrow$ & \textbf{Phys MAE} $\downarrow$ & \textbf{Sample MSE} $\downarrow$ \\
\midrule
Vanilla              & 0.13 & 5.53e-3 & 2.17e-1 & Vanilla          & 0.13 & 5.53e-3 & 2.17e-1 \\
ECI                  & 9.30 & 1.99e-4 & 1.66e-1 & ECI              & 9.30 & 1.99e-4 & 1.66e-1 \\
DiffusionPDE         & 0.63 & 1.27e-3 & 1.88e-1 & DiffusionPDE     & 0.63 & 1.27e-3 & 1.88e-1 \\
\midrule
EGT ($\gamma=10$)    & 0.23 & 3.47e-3 & 1.52e-1 & EGT (Sine)       & 0.23 & 1.76e-3 & 1.44e-1 \\
EGT ($\gamma=25$)    & 0.23 & 1.54e-3 & 1.46e-1 & EGT (Cosine)     & 0.23 & 6.11e-4 & 1.37e-1 \\
EGT ($\gamma=50$)    & 0.23 & 4.24e-4 & 1.41e-1 & EGT (Linear)     & 0.23 & 6.21e-4 & 1.37e-1 \\
EGT ($\gamma=75$)    & 0.23 & 6.07e-4 & 1.39e-1 & EGT (Constant)   & 0.23 & 8.56e-4 & 1.81e-1 \\
EGT ($\gamma=100$)   & 0.23 & 6.41e-4 & 1.38e-1 & EGT (Parabolic)  & 0.23 & 6.07e-4 & 1.39e-1 \\
EGT ($\gamma=150$)   & 0.23 & 7.10e-4 & 1.37e-1 &                  &      &         &         \\
\bottomrule
\end{tabular}%
}
\end{table}

\begin{table}[htbp]
\centering
\caption{\rev{Guidance strength ($\gamma$) and scheduler Pareto ablation on the \textbf{1D Heat} system.}}
\label{tab:pareto_heat}
\resizebox{\textwidth}{!}{%
\begin{tabular}{l c c c | l c c c}
\toprule
\textbf{Method ($\gamma$ Ablation)} & \textbf{Speed (s)} $\downarrow$ & \textbf{Phys MAE} $\downarrow$ & \textbf{Sample MSE} $\downarrow$ & \textbf{Method (Scheduler Ablation)} & \textbf{Speed (s)} $\downarrow$ & \textbf{Phys MAE} $\downarrow$ & \textbf{Sample MSE} $\downarrow$ \\
\midrule
Vanilla              & 0.13 & 7.38e-3 & 1.12e-1 & Vanilla          & 0.13 & 7.38e-3 & 1.12e-1 \\
ECI                  & 8.80 & 1.95e-3 & 2.80e-2 & ECI              & 8.80 & 1.95e-3 & 2.80e-2 \\
DiffusionPDE         & 0.63 & 6.90e-1 & 9.40e+1 & DiffusionPDE     & 0.63 & 6.90e-1 & 9.40e+1 \\
\midrule
EGT ($\gamma=10$)    & 0.26 & 4.67e-3 & 6.56e-2 & EGT (Sine)       & 0.27 & 3.91e-3 & 4.82e-2 \\
EGT ($\gamma=25$)    & 0.26 & 3.88e-3 & 4.98e-2 & EGT (Cosine)     & 0.27 & 2.30e-2 & 6.14e-2 \\
EGT ($\gamma=50$)    & 0.26 & 3.56e-3 & 4.25e-2 & EGT (Linear)     & 0.27 & 2.21e-2 & 5.78e-2 \\
EGT ($\gamma=75$)    & 0.26 & 3.41e-3 & 3.95e-2 & EGT (Constant)   & 0.27 & 2.17e-2 & 3.40e-2 \\
EGT ($\gamma=100$)   & 0.26 & 3.31e-3 & 3.76e-2 & EGT (Parabolic)  & 0.27 & 2.14e-2 & 6.69e-2 \\
EGT ($\gamma=150$)   & 0.26 & 3.19e-3 & 3.52e-2 &                  &      &         &         \\
\bottomrule
\end{tabular}%
}
\end{table}

\paragraph{\rev{Discussion.}} \rev{The Pareto analysis highlights two key operational insights. First, the guidance scale $\gamma$ smoothly controls the operating point of EGT in a problem-dependent manner: on Burgers, increasing $\gamma$ continuously improves Sample MSE while the physical residual reaches its minimum around $\gamma=50$; on Heat, both metrics monotonically improve across the evaluated range. Second, the optimal temporal schedule is problem-dependent: cosine, linear, and parabolic schedules yield the strongest overall balance on Burgers, whereas the transient sine schedule achieves by far the lowest physical residual on Heat. Across all configurations, EGT reduces inference cost by one to two orders of magnitude compared to explicit projections (ECI) while avoiding any computational penalty when tuning hyperparameters.}

\subsection{\rev{Contribution of Continuous Guidance vs. Post-Hoc Refinement}}
\label{app:ablation_refinement}

\rev{To isolate the distinct contribution of our continuous transport mechanism (Pure EGT) from the optional terminal energy descent (Hybrid EGT), we apply the exact same post-hoc optimization procedure (identical learning rate and number of refinement steps) to the terminal states generated by Vanilla FFM, DiffusionPDE, and ECI under the White Noise Full-PDE setting ($\gamma=75$ for Burgers, $\gamma=25$ for Navier--Stokes).}

\begin{table}[htbp]
\centering
\caption{\rev{Ablation study applying identical post-hoc terminal refinement across baselines on 1D Burgers and 2D Navier--Stokes (White Noise, Full-PDE).}}
\label{tab:ablation_refinement}
\resizebox{\textwidth}{!}{%
\begin{tabular}{l c c c c}
\toprule
\textbf{Model} & \textbf{Burgers (Sample MSE)} $\downarrow$ & \textbf{Burgers (Phys MAE)} $\downarrow$ & \textbf{NS (Sample MSE)} $\downarrow$ & \textbf{NS (Phys MAE)} $\downarrow$ \\
\midrule
Vanilla (Pure FFM)     & 2.04e-1 $\pm$ 1.33e-1 & 5.10e-3 $\pm$ 2.50e-3 & 5.60e-1 $\pm$ 2.34e-1 & 7.50e-3 $\pm$ 1.40e-3 \\
Vanilla (Refined)      & 2.01e-1 $\pm$ 1.31e-1 & 1.54e-4 $\pm$ 1.00e-4 & 5.40e-1 $\pm$ 2.34e-1 & 3.96e-3 $\pm$ 1.42e-3 \\
DiffusionPDE (Refined) & 1.53e-1 $\pm$ 1.15e-1 & 1.16e-4 $\pm$ 2.22e-5 & 5.41e-1 $\pm$ 2.30e-1 & 4.42e-4 $\pm$ 5.49e-4 \\
ECI (Refined)          & 1.16e-1 $\pm$ 9.50e-2 & 1.18e-4 $\pm$ 1.80e-5 & 6.20e-1 $\pm$ 4.02e-1 & 3.16e-5 $\pm$ 1.17e-5 \\
\midrule
\textbf{Ours (Pure EGT)}   & 1.89e-1 $\pm$ 1.21e-1 & 1.90e-3 $\pm$ 6.00e-4 & 5.47e-1 $\pm$ 2.32e-1 & 1.60e-3 $\pm$ 3.00e-4 \\
\textbf{Ours (Hybrid EGT)} & 1.89e-1 $\pm$ 1.18e-1 & 1.50e-4 $\pm$ 2.90e-5 & 5.47e-1 $\pm$ 2.32e-1 & 2.90e-5 $\pm$ 7.00e-6 \\
\bottomrule
\end{tabular}%
}
\end{table}

\paragraph{\rev{Discussion.}} \rev{Table~\ref{tab:ablation_refinement} demonstrates that post-hoc refinement alone cannot replace continuous guidance during the flow. While terminal optimization reduces residuals across all methods on 1D Burgers, the non-linear 2D Navier--Stokes system exposes a clear gap: refining unconditional samples (Vanilla Refined) plateaus at a physical MAE of $3.96\text{e-}3$, whereas Pure EGT already achieves $1.60\text{e-}3$ before any refinement. Applying the identical terminal optimization to EGT (Hybrid EGT) drives the error down to $2.90\text{e-}5$. Although ECI + Refinement reaches a comparable residual ($3.16\text{e-}5$), it relies on repeated projections that are nearly $5\times$ slower and degrade Sample MSE ($0.620$ vs.\ $0.547$).}

\newpage
\section*{NeurIPS Paper Checklist}

\begin{enumerate}

\item {\bf Claims}
    \item[] Question: Do the main claims made in the abstract and introduction accurately reflect the paper's contributions and scope?
    \item[] Answer: \answerYes{} 
    \item[] Justification: The abstract and introduction accurately summarize the paper's primary contributions: the Energy-Guided Transport framework, its theoretical foundations, and its empirical advantages over projection-based baselines.

\item {\bf Limitations}
    \item[] Question: Does the paper discuss the limitations of the work performed by the authors?
    \item[] Answer: \answerYes{} 
    \item[] Justification: Limitations regarding the sensitivity to the guidance scale $\gamma$ in high-dimensional settings and the reliance on discrete stencils for energy evaluation are discussed in the Conclusion and Appendix J.

\item {\bf Theory assumptions and proofs}
    \item[] Question: For each theoretical result, does the paper provide the full set of assumptions and a complete (and correct) proof?
    \item[] Answer: \answerYes{} 
    \item[] Justification: All mathematical assumptions required for the derivation of the dissipative force and the stability analysis are explicitly stated in Appendix A.

    \item {\bf Experimental result reproducibility}
    \item[] Question: Does the paper fully disclose all the information needed to reproduce the main experimental results of the paper to the extent that it affects the main claims and/or conclusions of the paper (regardless of whether the code and data are provided or not)?
    \item[] Answer: \answerYes{} 
    \item[] Justification: We fully disclose all necessary information to reproduce our experiments. The model architectures, continuous guidance hyperparameters, ODE integration budgets, and evaluation metrics are meticulously detailed in Appendices E, G, and J. Additionally, we provide a GitHub repository containing the complete source code to seamlessly reproduce all baseline and proposed framework results.

\item {\bf Open access to data and code}
    \item[] Question: Does the paper provide open access to the data and code, with sufficient instructions to faithfully reproduce the main experimental results, as described in supplemental material?
    \item[] Answer: \answerYes{} 
    \item[] Justification: We provide a link to an GitHub repository containing the implementation, dataset generation scripts, and instructions for reproduction in the introduction.

\item {\bf Experimental setting/details}
    \item[] Question: Does the paper specify all the training and test details (e.g., data splits, hyperparameters, how they were chosen, type of optimizer) necessary to understand the results?
    \item[] Answer: \answerYes{} 
    \item[] Justification: Detailed training regimes, FNO architectures, and optimization hyperparameters are documented in Appendices E and G.

\item {\bf Experiment statistical significance}
    \item[] Question: Does the paper report error bars suitably and correctly defined or other appropriate information about the statistical significance of the experiments?
    \item[] Answer: \answerYes{} 
    \item[] Justification: We report Sample MSE, MMSE, and SMSE across all benchmarks, and provide a dedicated statistical significance and multi-seed variability analysis (across 5 independent random seeds) in Appendix J.7.

\item {\bf Experiments compute resources}
    \item[] Question: For each experiment, does the paper provide sufficient information on the computer resources (type of compute workers, memory, time of execution) needed to reproduce the experiments?
    \item[] Answer: \answerYes{} 
    \item[] Justification: Hardware specifications (NVIDIA RTX 8000, A6000, L40) and average inference times per sample are provided in Appendices E and J.
    
\item {\bf Code of ethics}
    \item[] Question: Does the research conducted in the paper conform, in every respect, with the NeurIPS Code of Ethics \url{https://neurips.cc/public/EthicsGuidelines}?
    \item[] Answer: \answerYes{} 
    \item[] Justification: This research focuses on fundamental computational physics and does not have foreseeable direct negative societal impacts.

\item {\bf Broader impacts}
    \item[] Question: Does the paper discuss both potential positive societal impacts and negative societal impacts of the work performed?
    \item[] Answer: \answerNA{} 
    \item[] Justification: Our study uses synthetic physical datasets (Heat, Burgers, Navier-Stokes) and does not involve human subjects or data derived from them.
    
\item {\bf Safeguards}
    \item[] Question: Does the paper describe safeguards that have been put in place for responsible release of data or models that have a high risk for misuse (e.g., pre-trained language models, image generators, or scraped datasets)?
    \item[] Answer: \answerNA{} 
    \item[] Justification: This research focuses on solving standard continuous physical systems (e.g., Navier-Stokes, Burgers' equations) using synthetic data, which poses no high risks for societal misuse or dual-use.

\item {\bf Licenses for existing assets}
    \item[] Question: Are the creators or original owners of assets (e.g., code, data, models), used in the paper, properly credited and are the license and terms of use explicitly mentioned and properly respected?
    \item[] Answer: \answerYes{} 
    \item[] Justification: All existing assets (e.g., baseline model architectures, PDE simulation environments) used or adapted in our experiments are properly cited in the main text and Appendices, respecting their original open-source licenses.

\item {\bf New assets}
    \item[] Question: Are new assets introduced in the paper well documented and is the documentation provided alongside the assets?
    \item[] Answer: \answerYes{} 
    \item[] Justification: We introduce a new codebase for the Energy-Guided Transport framework. It is thoroughly documented and provided via an GitHub repository linked in the submission, complete with instructions for data generation and reproducibility.

\item {\bf Crowdsourcing and research with human subjects}
    \item[] Question: For crowdsourcing experiments and research with human subjects, does the paper include the full text of instructions given to participants and screenshots, if applicable, as well as details about compensation (if any)? 
    \item[] Answer: \answerNA{} 
    \item[] Justification: This research does not involve crowdsourcing or human subjects.

\item {\bf Institutional review board (IRB) approvals or equivalent for research with human subjects}
    \item[] Question: Does the paper describe potential risks incurred by study participants, whether such risks were disclosed to the subjects, and whether Institutional Review Board (IRB) approvals (or an equivalent approval/review based on the requirements of your country or institution) were obtained?
    \item[] Answer: \answerNA{} 
    \item[] Justification: This research does not involve human subjects, making IRB approval non-applicable.

\item {\bf Declaration of LLM usage}
    \item[] Question: Does the paper describe the usage of LLMs if it is an important, original, or non-standard component of the core methods in this research? Note that if the LLM is used only for writing, editing, or formatting purposes and does \emph{not} impact the core methodology, scientific rigor, or originality of the research, declaration is not required.
    \item[] Answer: \answerNA{} 
    \item[] Justification: LLMs were used for writing assistance, structural editing, and formatting purposes.

\end{enumerate}

\end{document}